%% file: main.tex
\documentclass[11pt]{article}

\usepackage[margin=1in]{geometry}
\usepackage[T1]{fontenc}
\usepackage[utf8]{inputenc}
\usepackage{lmodern}
\usepackage{microtype}
\usepackage{amsmath,amssymb}
\usepackage{graphicx}
\usepackage{booktabs}
\usepackage{makecell}
\usepackage{caption}
\usepackage[natbibapa]{apacite}
\usepackage{xcolor}
\usepackage{hyperref}
\hypersetup{colorlinks=true, linkcolor=blue!50!black, citecolor=blue!50!black, urlcolor=blue!50!black}

\graphicspath{{figures/}}
\title{Deep LLM layers expose passage-level context in EEG during silent reading}

\author{Ingo Marquardt\thanks{Corresponding author: \texttt{ingo@nubrain.com}}, Anthilia Alchanat, Priyanka Jain\\[0.5em] nubrain}

\date{October 2026}

\begin{document}

\shortcites{abdulghani2024enhancing,card2024accurate,cooney2019classification,correia2014brain,csaky2025towards,dash2020decoding,defossez2023decoding,grattafiori2024llama,jiang2024large,kunz2025inner,ling2019visual,simistira2022rethinking,martin2016word,metzger2023high,moses2021neuroprosthesis,mugler2014direct,radford2021learning,sato2024scaling,wandelt2024representation,marquardt2026decoding,schrimpf2021neural,goldstein2022shared,caucheteux2023evidence,tenney2019bert,willett2023high,voita2019bottom,wehbe2014aligning}

\maketitle

\begin{abstract}
In a previous report we showed that a decoder can recover open-vocabulary word-level information from 19-channel EEG recorded during silent reading, by aligning short EEG windows with hidden-state embeddings of the presented word taken from a large language model (LLM). That analysis contrasted two of the model's 33 layers: the context-free input embedding and one mid-depth layer. Here we ask how EEG-decodable structure is distributed over the full depth of the language model. We trained one decoder per layer of Llama-3.1-8B, with and without a causal transformer over the sequence of EEG windows. We distinguish word-level decodability from a context-tracking component. Word-level decodability peaks in early layers and declines steadily towards the last layer. In contrast, the context-tracking component of the retrieval gain rises with depth, both with and without a causal transformer. Unexpectedly, a decoder that sees a single 0.5-s window at a time tracks the text passage too. Two new controls dissect this component. A target-geometry control, a synthetic predictor made of the true target plus noise calibrated to each decoder's word-level accuracy, shows that context tracking is not a by-product of word decoding and target geometry. A control that permutes predictions across runs from either the same or from different sessions shows that most of the observed context-tracking is not caused by an EEG recording session signature. Instead, context tracking mostly follows the text. A single EEG window therefore carries passage-level information that a contextual target can absorb. The results locate word identity in the early layers of the language model, discourse state in the deeper layers, and make the choice of target layer a design dial for EEG language decoders.
\end{abstract}

\section{Introduction}
\label{sec:introduction}

Language models and human readers process text in the same order: one token, or one word, at a time. A causal language model updates a layered internal state with every token, and layer-wise analyses of such models show how that state changes with depth. A token's vector becomes more context-specific and less predictable from the token's identity with every layer \citep{ethayarajh2019contextual}. Lexical and local syntactic information is read out from the lower layers and context-dependent structure from the upper layers \citep{tenney2019bert}. In a causal model that reads text left-to-right, information about the current token fades with depth while information about the next token accumulates \citep{voita2019bottom}. A human reader updates an internal state with every word too. When brain activity during language comprehension is aligned with the internal states of a language model, the alignment depends on which layer the states are taken from, and the depth profile of that alignment has become a standard tool for characterising the representation the brain signal is closest to \citep{jain2018incorporating, toneva2019interpreting, schrimpf2021neural, caucheteux2022brains, caucheteux2023evidence}. Most of that work uses linear encoding models, fitted to fMRI or MEG and in some cases to intracranial recordings \citep{schrimpf2021neural, goldstein2022shared}. The closest precedent to our paradigm aligned MEG recorded during word-by-word reading with the word embedding and the context state of a recurrent language model \citep{wehbe2014aligning}. But whereas many previous reports predict brain recordings from text embeddings, we attempt to predict text embeddings from EEG data. Specifically, we train a contrastive decoder to retrieve the presented word from 19-channel dry-electrode EEG.

In a previous report \citep[henceforth report 1]{marquardt2026decoding} we established the paradigm and the decoder, and showed that word-level information is recoverable from EEG during silent reading of naturalistic text, that it extends to rare and mid-frequency words, that it scales log-linearly with data volume, and that it is not reducible to narrative topic tracking or to a positional prior. That analysis used two target layers of Llama-3.1-8B \citep{grattafiori2024llama}: layer 0, the context-free input embedding, and layer 20, chosen on the assumption that a mid-to-deep layer would carry rich semantic representations without being dominated by next-token prediction. The present report systematically tests that assumption. We trained one decoder for every layer, 0 to 32, with and without a sequence model over the EEG, on five train/validation splits, and evaluated every fit with the full set of controls of report 1.

We ask two questions. First, how is EEG-decodable structure distributed over the depth of the language model, along each axis report 1 used to characterise decoding: word-level versus passage-level retrieval, the position probes, and the word-frequency bins? Second, a question raised by our initial results. A decoder that sees one 0.5-s EEG window at a time has no access to neighbouring windows, yet if the target carries context, it tracks the passage being read. What does a single window share with its passage? We introduce two controls to answer it, a lag-dependent version of report 1's run-mate swap with a calibrated target-geometry control, and a comparison of narrative neighbours recorded in the same versus different sessions.

\section{Methods}
\label{sec:methods}

\subsection{Data, decoder and metrics}
\label{sec:recap}

Paradigm, acquisition, preprocessing, decoder and evaluation are those of report 1 \citep{marquardt2026decoding}. Here we summarise the essentials and refer to report 1 for details.

\textbf{Data.} One participant read continuous narrative text (detective stories, sections from a fantasy novel and a science-fiction novel, and two newspaper articles) presented one word at a time by rapid serial visual presentation, with typography randomised on every trial, while EEG was recorded from 19 dry electrodes at 600\,Hz. The complete dataset comprises 409 recording runs of about 600 words each, recorded in 52 sessions on 42 days spanning 146 days (a session is one sitting with the cap on), 249{,}753 word presentations, about 51\,h of reading, of which 232{,}028 (93\%) pass the amplitude inclusion criterion. Report 1 analysed the first 393 runs; the present sweep uses all 409, so absolute values differ slightly from report 1.

\textbf{Decoder.} A convolutional encoder maps a 0.5-s onset-locked EEG window (19 channels, 300 samples) to a 256-dimensional feature vector. Optionally, a four-layer causal transformer over the sequence of feature vectors within a recording run (the ``sequence model'') lets the prediction for word $i$ depend on the EEG of words $j \le i$; a learned positional embedding is its only non-neural, trial-varying input. Encoder (or transformer) outputs and linearly projected LLM targets meet in a shared, L2-normalised 256-dimensional space, trained with the symmetric CLIP objective \citep{radford2021learning} over batches of eight complete runs (about 4{,}400 in-batch negatives). Fits ran for 100 epochs with AdamW; checkpoints were selected on the within-run gain.

\textbf{Targets.} Each word is represented by the hidden state of the last sub-word token of its punctuation-stripped form at a chosen layer of Llama-3.1-8B, after a causal forward pass over the run's full text. Layer 0 is the input embedding, which depends only on the word's identity; layers 1 to 32 are the outputs of the 32 decoder blocks, each incorporating all preceding words through causal attention; layer 32 is the final hidden state. The targets are these 4{,}096-dimensional hidden states. The decoder's bias-free linear projection maps them into the shared, L2-normalised 256-dimensional space, in which all similarities of this report are computed. The controls of Sections~\ref{sec:lag} and~\ref{sec:session} likewise operate on the projected targets and predictions.

\textbf{Metrics.} All metrics are word-grouped top-10 retrieval within pools of 512 validation trials (probability mass summed across trials sharing a core word before ranking), averaged over five pool draws, and reported as the gain over an empirical permutation baseline (10 permutations), in percentage points (pp), so that chance is 0\,pp. We rely on the following metrics to distinguish between different aspects of decoding and to account for possible confounds:

\begin{enumerate}
  \item The \textbf{within-run gain}, the primary metric, with pools of consecutive trials from one run and a within-run permutation null, so that topic tracking cannot contribute.
  \item The \textbf{overall gain}, with random cross-run pools
  \item The \textbf{context-tracking gain}, the overall gain recomputed after replacing every trial's EEG prediction with that of a random other trial of the same run, which measures the passage-level component of the overall gain.
  \item The \textbf{context-independent gain} is the overall minus the context-tracking gain.
  \item Two \textbf{position probes} re-run a transformer fit with the EEG replaced by the learnable mask token, for every trial (position-only) or for the current trial only (mask-current), and decompose the within-run gain into a non-neural position-only term, a preceding-EEG term and a current-trial term.
  \item The \textbf{position-corrected} within-run gain (within-run minus position-only) is the quotable word-level figure for transformer fits, and coincides with the within-run gain for fits without a sequence model, where both probes are zero by construction. Gains are also computed per word-frequency tertile (rare, mid, frequent), each against its own baseline.
\end{enumerate}

\subsection{The layer sweep}
\label{sec:sweep}

One decoder was trained for every target layer $\ell = 0, \ldots, 32$ and for each architecture (without and with the sequence model), with every other hyperparameter fixed at a configuration that performed well in report 1's sweeps: EEG windows locked to word onset, 0.5-s window length, no temporal-shift augmentation, amplitude scaling only, no masking of within-run negatives, trainable temperature, batch size 8, learning rate $10^{-4}$, 100 epochs. The $33 \times 2$ grid was repeated for five randomisation seeds, 330 fits in total. The seed fixes the content-aware train/validation split (about 20\% of the corpus by word count, 85--86 validation runs per seed, no 10-gram overlap between the two sides); model initialisation and batch order are not seeded and differ between every pair of fits. The five seeds are therefore independent fits on different held-out text, and ``mean $\pm$ SD over seeds'' throughout this report is a conservative error bar that includes both training stochasticity and the choice of held-out passages.

\subsection{Lag-dependent run-mate swap and the target-geometry control}
\label{sec:lag}

In order to quantify to what extent our EEG model decodes textual context (as opposed to individual word identity), we previously calculated a context-tracking gain by replacing each trial's prediction with that of a \emph{randomly} sampled trial from the same run (report 1). For contextual targets that gain is positive even without a sequence model. That is not necessarily expected; without a sequence model, the EEG model only sees a single EEG segment (corresponding to a single word) at a time, so the contextual information has to come from this single EEG segment. We propose three potential mechanisms:

\begin{enumerate}
  \item \emph{Word decoding plus target geometry}: To calculate the context-tracking gain, a trial's prediction is swapped with another trial's prediction from the same run. The true target vectors (from the LLM) of words from the same text passage are expected to be more similar to each other than to targets from other text passages, and the EEG model might learn this pattern. Hence, a decoded run-mate still ranks the true target above candidates from other runs; no context in the EEG is required.
  \item \emph{Context in the single window}: The evoked response to a word depends on the discourse state \citep{kutas2011thirty}, and the decoder maps it to the run's contextual component. This mechanism would be genuine decoding of discourse context from the neural response.
  \item \emph{A recording-session signature}: Slow non-language EEG variance (electrode impedance, cap placement, vigilance) that correlates with progress through the corpus because the narrative continues across runs and sessions, and that generalises to a held-out run because validation runs are interleaved in time with training runs.
\end{enumerate}

We repeated the swap with the run-mate at a fixed signed lag of $k$ included trials (roughly $k$ words) earlier or later, $k \in \{1, 2, 4, \ldots, 256\}$, reflecting at run boundaries so that $|k|$ is exact wherever the run is long enough (Figure~\ref{fig:design}b). Mechanisms 1 and 2 should decay with the lag, because the contextual similarity of two words falls off with their distance in the text, whereas a session signature is constant within a run. To bound mechanism 1 we built a \textbf{target-geometry control}: a synthetic predictor that has word-level information and nothing else. Its prediction for a trial is the trial's true projected target plus isotropic Gaussian noise in the shared 256-dimensional space, re-normalised, with the noise standard deviation calibrated per checkpoint by a 14-step bisection search so that the control's within-run gain equals the checkpoint's (Supplement, Section~\ref{sec:s0}). The control is then swapped at the same lags. It contains no information from the EEG, so any cross-run gain it shows arises from the geometry of the targets alone, and only the excess of the decoder over the control requires context in the EEG. It is distinct from report 1's context-tracking control (the random-run-mate swap), which is applied to the decoder's own predictions.

\subsection{Narrative neighbours across session boundaries}
\label{sec:session}

\begin{figure}[tp]
  \centering
  \includegraphics[alt={Schematic of the layer sweep design, the lag-dependent run-mate swap and the narrative-neighbour comparison across sessions},width=\textwidth]{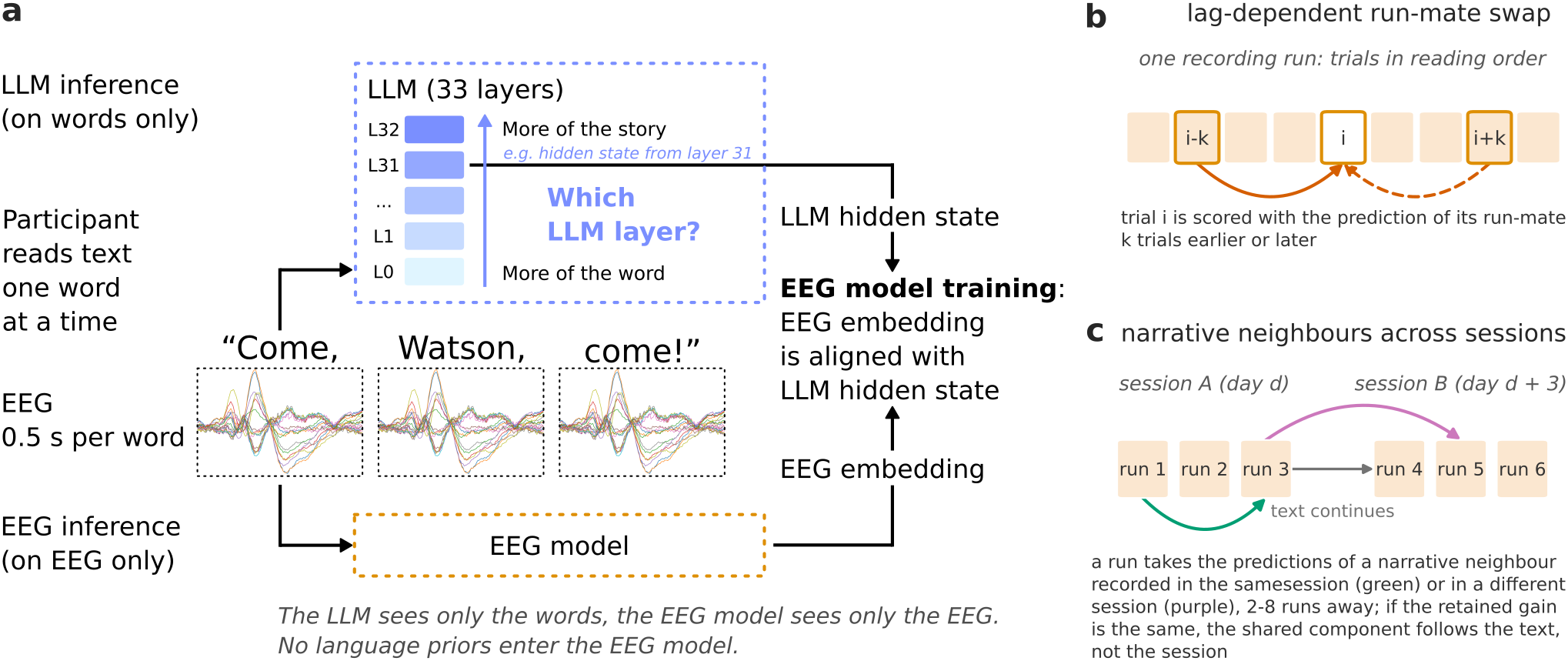}
  \caption{\textbf{Design.} \textbf{(a)} Two readers. A human participant reads text one word at a time while EEG is recorded. An LLM processes the same text one token at a time. Each token updates the LLM's hidden states. We extract a hidden state from each layer of the LLM. Early layers represent token / word identity, whereas deeper layers are more contextual. Our EEG decoder learns to align a 0.5-s EEG window with the LLM's hidden state for that word. We separately train one EEG decoder per layer (0--32), without and with a transformer over the sequence of EEG windows. \textbf{(b)} Lag-dependent run-mate swap. We have previously observed that our EEG decoder tracks the narrative context of text, and that this context tracking component is distinct from word-level decoding \citep{marquardt2026decoding}. Here, we conduct further tests on the context tracking. The lag-dependent run-mate swap probes how far the information shared between EEG trials reaches. Each trial's own prediction is replaced by the prediction of a run-mate k trials earlier or later before computing the gain in retrieval accuracy over a permutation baseline. \textbf{(c)} Narrative neighbours across sessions. The context-tracking gain may be confounded by an EEG session signature that is unrelated to the text. To test this, we permute predictions within sessions and across sessions.}
  \label{fig:design}
\end{figure}

The lag swap cannot separate confound mechanism 3 (a potential recording-session EEG signature) from mechanism 2 (a successfully decoded run-level neural discourse state), because both are constant within a run. But they can be distinguished across runs. The participant read the corpus continuously, so consecutive runs continue the text within a session and across the boundary between sessions\footnote{A session is one sitting with the EEG cap on, typically separated by one or more days.} Two runs that are neighbours in the text can therefore belong to the same session or to different sessions. For every validation run and every shift $d \in \{\pm 2, \ldots, \pm 8\}$,\footnote{Shift $\pm 1$ yields only one to four cross-session pairs of runs per split (each scored once in each direction). Both runs in a pair must be validation runs, because only held-out runs have uncontaminated predictions. A cross-session pair at shift $\pm 1$ therefore needs two consecutive runs that straddle one of the 51 session boundaries and both fall in the validation split; with about a fifth of the runs held out, that happens only a few times per split. At larger shifts the partner crosses a session boundary far more often.} the run's trials took the predictions of the validation run $d$ places away in the narrative, either position-matched (trial $m$ takes trial $m$ of the partner, preserving any component that depends on position within the run) or position-scrambled (a random trial of the partner), and were scored in the random cross-run pools as before. The retained fraction of the run's own cross-run gain was averaged, trial-weighted, over shifts, separately for same-session and cross-session neighbours (Figure~\ref{fig:design}c). If the shared component follows the text, same-session and cross-session neighbours retain the same fraction; if it follows the session, the fraction retained by cross-session neighbours collapses.

As a complementary view, we asked what makes the decoder's predictions for two runs alike. For each validation run we averaged its predictions into one mean prediction vector and its projected targets into one mean target vector. For every pair of validation runs, the cosine similarity of the two mean predictions was regressed, by ordinary least squares with all variables standardised, on three candidate sources: the cosine similarity of the two mean targets (the text), a same-session indicator (a session signature) and the log of one plus the gap in days between the two recordings (slow drift). A session signature would appear as a positive same-session coefficient. Pairs share runs and are not independent, so the coefficients are descriptive. Because only 2--3\% of the pairs come from the same session, the standardised same-session coefficient is small by construction; the Supplement therefore also reports the term on the indicator's own 0/1 scale, compares same-session pairs directly with same-day pairs from different sessions, and tests it against a permutation null (Section~\ref{sec:s-regression}).

\subsection{Restricted candidate set}
\label{sec:small-pool}

The retrieval metrics above rank the true word among the words that occur in a pool of 512 validation trials, about 250--300 distinct words of an open vocabulary, and quote the gain over a permutation baseline. Closed-vocabulary speech neuroprostheses report instead the word error rate of sentences decoded under a fixed vocabulary, typically of 50 words \citep{moses2021neuroprosthesis, willett2023high, card2024accurate}, and the accuracy of classifying one isolated word of that set \citep{moses2021neuroprosthesis, willett2023high}. To make our numbers legible in the latter form, we re-scored the predictions of every fit on trials whose word belongs to a fixed list of 50 words. The model was not retrained; the restriction is applied at inference only. Two lists were defined once on the full corpus, so that they are identical for every split (Supplement, Section~\ref{sec:s-small-pool}, Table~\ref{tab:word-lists}): the 50 most frequent content words, and the 50 most frequent words of any kind, almost all of them function words. Pools held exactly one validation trial of every list word, so that a pool is a 50-way identification task with uniform chance (2\% top-1, 20\% top-10). The candidates of such a pool come from different passages, so a contextual target can be identified through the discourse state rather than through the word. Hence, two controls were scored on the same 50-word pools. In the first, every prediction is replaced by that of the run-mate one trial earlier, which carries the local discourse state and the position but not the current word. In the second, every prediction is replaced by that of a random run-mate (the context-tracking control). Finally, the within-run gain was recomputed on the list words alone, with pools made of the list-word trials of one run, where candidates from other passages are absent.

\subsection{Reading the curves}
\label{sec:statistics}

Every value is read at the epoch that maximised the within-run gain, on the validation set that also drove that selection, so absolute levels are optimistic. Cross-layer orderings are reliable (report 1, Section 2.8). Adjacent layers of a language model carry highly similar representations, so a real effect is expected to be a smooth function of the layer index. Where we report a peak, we report the peak of the seed-mean curve, each seed's own peak, and the mean pairwise Spearman correlation between the per-seed curves as a measure of how well the seeds agree on the shape of the profile.

\section{Results}
\label{sec:results}

Every cell of the $33 \times 2$ grid is complete for all five seeds. Values are means over seeds with the across-seed SD, in pp; all tables are in the Supplement.

\subsection{Word-level decodability peaks in the early layers and declines with depth}
\label{sec:results-word}

\begin{figure}[tp]
  \centering
  \includegraphics[alt={Within-run and position-corrected within-run gain against the target-embedding layer for both decoders},width=\textwidth]{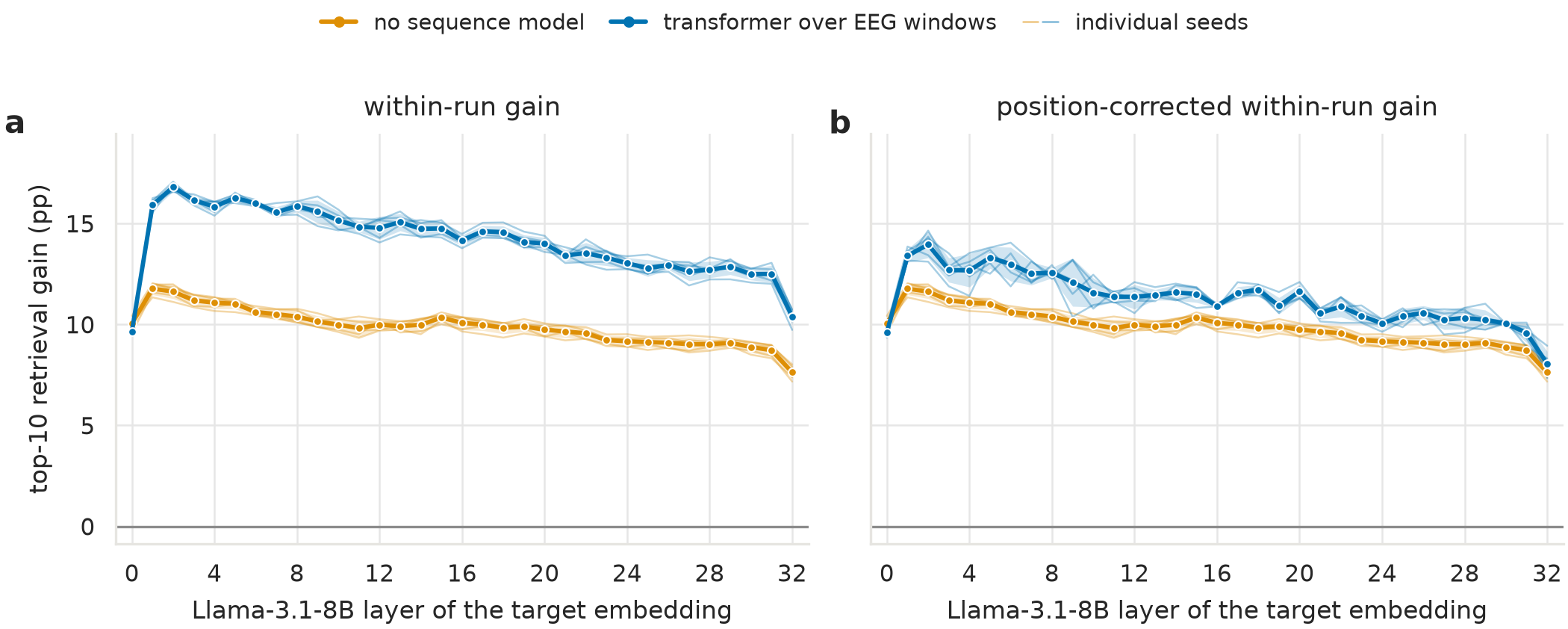}
  \caption{\textbf{Word-level decodability across layers.} \textbf{(a)} Within-run retrieval gain (candidates from the same passage; the primary metric) against the layer of Llama-3.1-8B the target embedding was taken from, for the decoder without a sequence model (orange) and with the transformer over EEG windows (blue). Lines and bands are the mean and SD over the five train/validation splits, and faint lines are the individual splits. \textbf{(b)} The same after subtracting the transformer's position-only probe (without a sequence model (a) and (b) are identical because there is no positional embedding). Chance is 0\,pp. The two decoders are indistinguishable at layer 0, where the target carries no context; word-level decodability peaks one or two blocks into the model and declines steadily with depth, and the shape is the same in every split. (b) shows that the profile survives removal of the transformer's positional shortcut, indicating that the profile is the result of EEG-to-LLM alignment, and not a modelling confound.}
  \label{fig:word}
\end{figure}

Figure~\ref{fig:word} shows the primary metric against the target layer. At layer 0 the target carries no context and the two decoders are indistinguishable: $10.1 \pm 0.3$\,pp without and $9.7 \pm 0.1$\,pp with the sequence model. A single layer of context changes this. Without a sequence model the within-run gain rises to $11.8 \pm 0.3$\,pp at layer 1 and then declines steadily, to $9.7$\,pp at layer 20 and $7.6 \pm 0.4$\,pp at the last layer. With the sequence model it rises to $16.8 \pm 0.2$\,pp at layer 2, stays above 15\,pp through layer 10, and declines to $14.0$\,pp at layer 20 and $10.4 \pm 0.4$\,pp at layer 32. After subtracting the positional shortcut (Figure~\ref{fig:word}b), the transformer's curve peaks at $14.0 \pm 0.7$\,pp at layer 2 and ends at $8.1 \pm 0.6$\,pp, with the same shape.

The shape replicates. Four of the five seeds peak at layer 1 and one at layer 2 without the sequence model, all five peak at layer 2 with it, and the per-seed curves correlate at $\rho = 0.95$ and $0.97$ pairwise (Supplementary Table~\ref{tab:peaks}); the across-seed SD is a few tenths of a point. From layer 1 onward the decline is close to monotonic: over layers 1--32 the Spearman correlation between layer index and within-run gain is $-0.97$ without and $-0.98$ with the sequence model ($-0.94$ and $-0.81$ over all 33 layers, where layer 0 precedes the peak; Supplementary Table~\ref{tab:peaks}). Layer 20, report 1's choice, sits in the declining part of both curves: for the transformer it gives 14.0\,pp against 16.8\,pp at the peak.

\subsection{Deeper layers buy the passage, not the word}
\label{sec:results-context}

\begin{figure}[tp]
  \centering
  \includegraphics[alt={Overall gain, context-tracking gain, context-independent gain and context-tracking share against layer for both decoders},width=\textwidth]{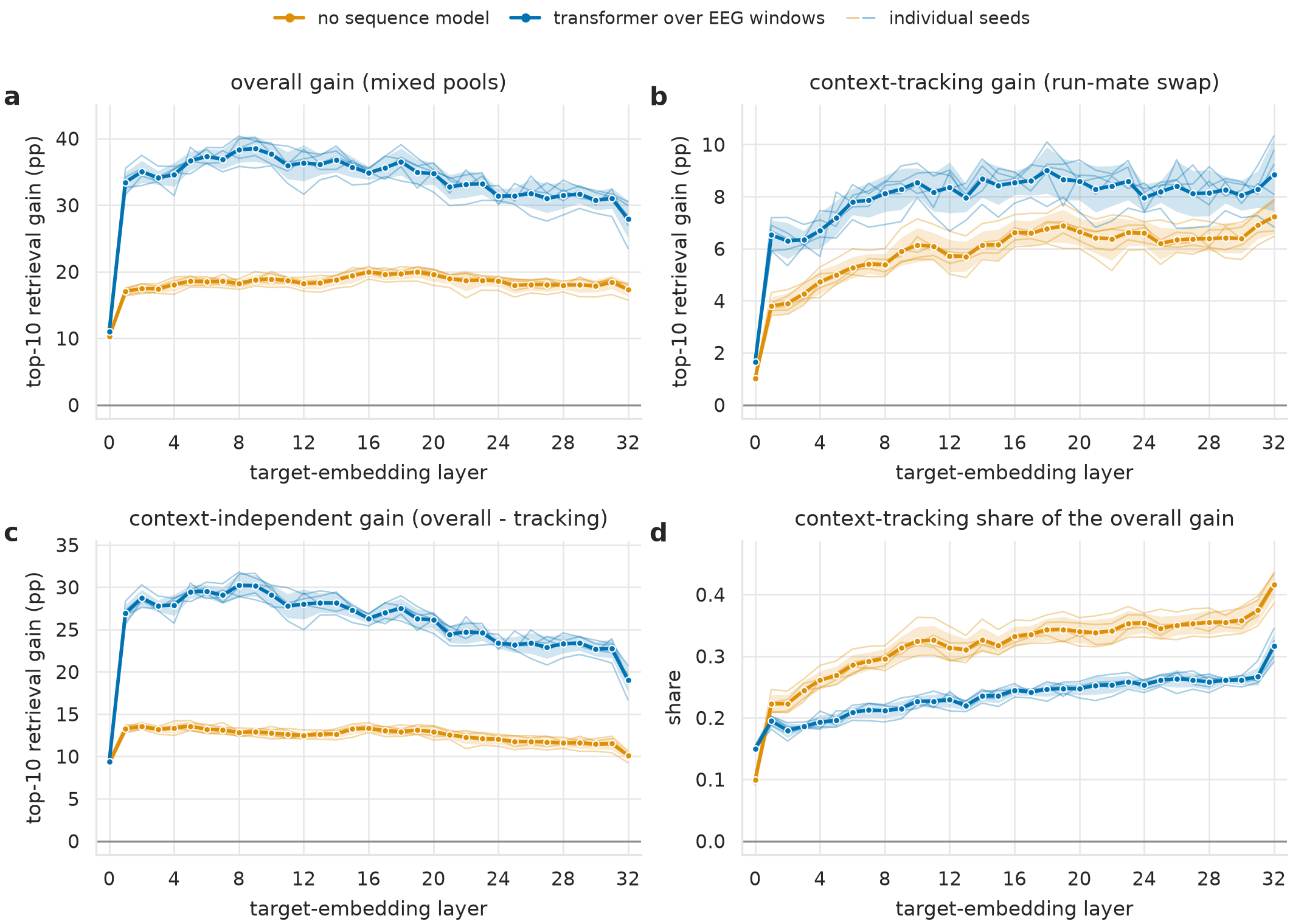}
  \caption{\textbf{Overall gain and its context-tracking share across layers.} \textbf{(a)} Overall gain in random cross-run pools (where the correct word has to be selected among candidates from many different runs / text passages). \textbf{(b)} Its context-tracking component, measured by replacing every trial's prediction with a randomly sampled prediction from the same run. \textbf{(c)} The context-independent remainder (overall minus context-tracking). \textbf{(d)} The context-tracking share of the overall gain. Conventions as in Figure~\ref{fig:word}. Panel (a) shows how much the decoder gains when candidates come from many passages, which is far more with the sequence model. (b) shows that the part of that gain earned by recognising the passage rises with depth for both decoders, i.e. also for the one without a sequence model that only sees a single EEG window (discussed in Section~\ref{sec:results-window}). (c) shows that the remaining context-independent gain (after subtracting the context-tracking component) follows the word-level metric's slow decline towards the deep layers. (d) shows that the passage becomes a larger share of what the decoder reads with every layer.}
  \label{fig:context}
\end{figure}

The overall gain in mixed pools (Figure~\ref{fig:context}a) leaps from $11.0$\,pp at layer 0 to $33.4 \pm 1.5$\,pp at layer 1 with the sequence model, peaks at $38.5 \pm 1.9$\,pp at layer 9, and falls to $27.9 \pm 2.8$\,pp at the last layer; without a sequence model it rises from $10.3$ to about 17--20\,pp and stays there. The composition of this gain shifts with depth. Its context-tracking component (Figure~\ref{fig:context}b) grows with depth: without a sequence model almost monotonically, from $1.0$ to $7.2 \pm 0.6$\,pp (Spearman $\rho$ with layer index $0.84$; every seed peaks at the last layer), and with the sequence model from $1.6$ to about 8--9\,pp by the middle layers, where it stays ($8.9 \pm 1.4$\,pp at layer 32; $\rho = 0.52$), while the context-independent remainder (Figure~\ref{fig:context}c) follows the slow decline of the within-run gain. The share of the overall gain that is passage tracking (Figure~\ref{fig:context}d) rises from 19\% at layer 1 to 32\% at layer 32 with the sequence model and from 10\% at layer 0 to 42\% at layer 32 without. Deeper targets therefore make the decoder better at recognising which passage is being read, not which word.

The orange curve of Figure~\ref{fig:context}b is the observation that motivated the second half of this report. The decoder without a sequence model sees one 0.5-s window and nothing else, yet from layer 4 onwards it tracks the passage at 5--7\,pp. Section~\ref{sec:results-window} asks what a single window can share with its passage.

\subsection{Rare words are decoded best against mid-depth targets}
\label{sec:results-frequency}

\begin{figure}[tp]
  \centering
  \includegraphics[alt={Within-run gain per word-frequency bin against layer for both decoders, and the rare-to-frequent ratio},width=\textwidth]{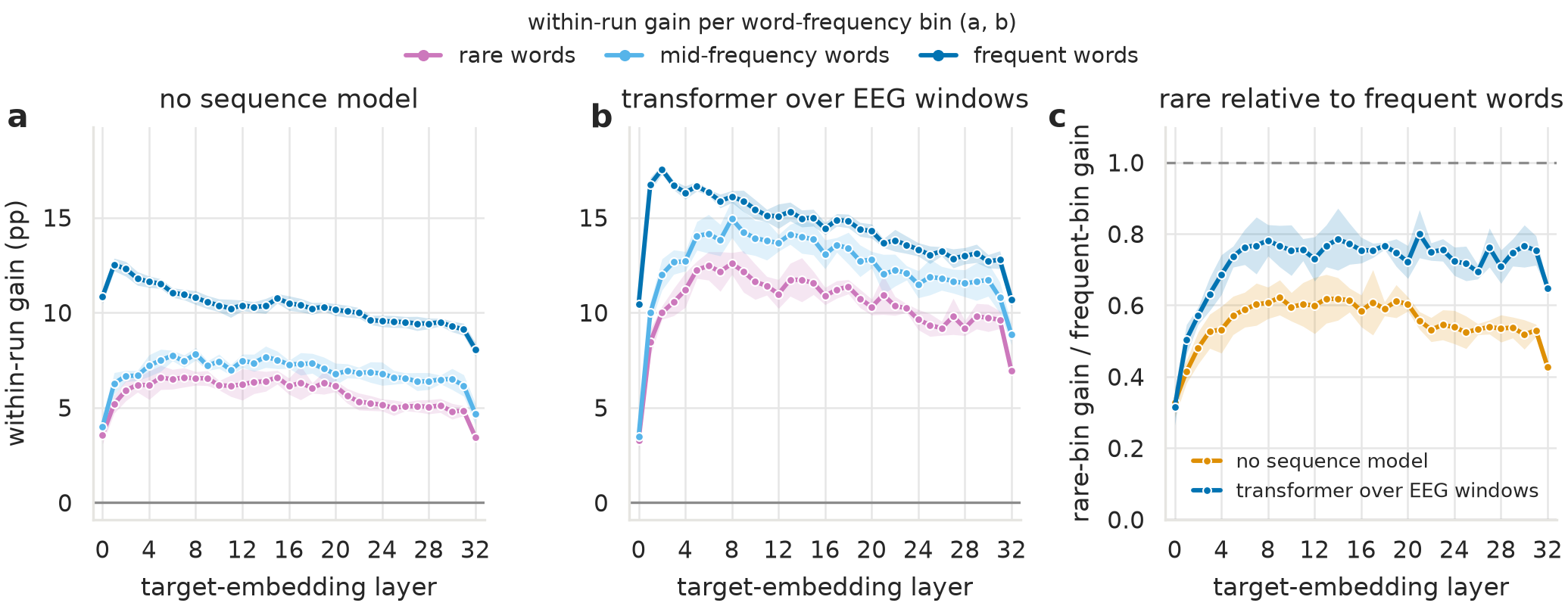}
  \caption{\textbf{Word-frequency bins across layers.} Within-run gain for rare, mid-frequency and frequent words (tertiles by validation trial count, each against its own baseline) against layer, \textbf{(a)} without and \textbf{(b)} with the sequence model. \textbf{(c)} Rare-bin gain divided by frequent-bin gain; 1.0 would mean rare words decode as well as frequent ones. Means and SD over splits. Panels (a, b) show that frequent words drive the early peak, whereas rare and mid-frequency words keep improving further into the middle of the network; (c) makes the comparison directly: relative to frequent words, rare words are decoded worst against the context-free layer 0 and best against mid-depth targets.}
  \label{fig:bins}
\end{figure}

Frequent words drive the early peak (Figure~\ref{fig:bins}). Their within-run gain peaks at layers 1--2 in both decoders ($12.5 \pm 0.3$ and $17.5 \pm 0.2$\,pp) and declines from there. Rare and mid-frequency words behave differently: their curves keep rising past the frequent-word peak and plateau across the middle of the network. With the sequence model the rare-word gain rises from $3.3$\,pp at layer 0 to $12.6 \pm 0.6$\,pp at layer 8 and stays above 10\,pp through layer 20 (10.3\,pp) before falling to $6.9$\,pp at layer 32; without a sequence model it rises from $3.6$ to $6.6 \pm 0.4$\,pp at layer 15. The rare-to-frequent ratio, $0.32$--$0.33$ at layer 0 for both decoders, rises to 0.7--0.8 across the middle layers with the sequence model (peak $0.80 \pm 0.07$ at layer 21) and to about 0.6 without (peak $0.62 \pm 0.05$ at layer 9), and eases off in the last few layers (Figure~\ref{fig:bins}c). For the words a communication device needs most, the context-free target is the worst choice.

\subsection{Position probes and what the sequence model adds}
\label{sec:results-probes}

\begin{figure}[tp]
  \centering
  \includegraphics[alt={Nested position-probe decomposition of the transformer's within-run gain per layer, and the transformer minus no-sequence-model difference per metric},width=\textwidth]{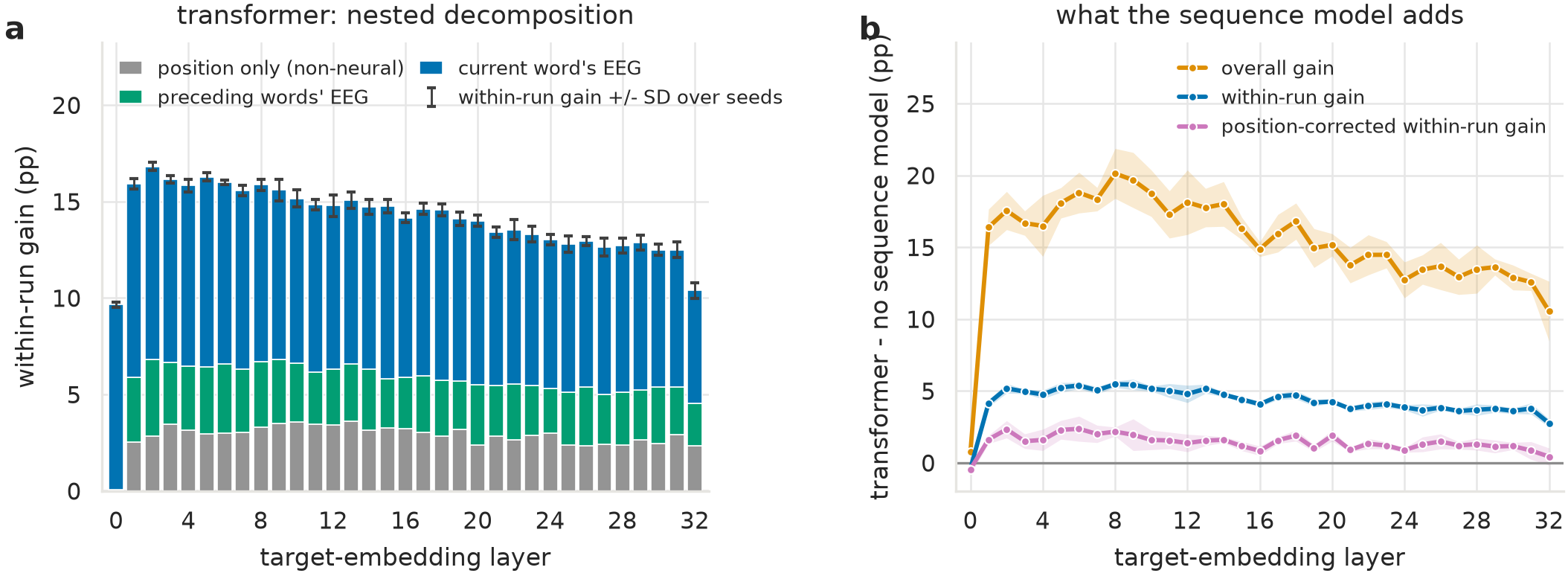}
  \caption{\textbf{Position probes and the transformer's advantage across layers.} \textbf{(a)} The transformer's within-run gain decomposed into the position-only (non-neural), preceding-EEG and current-trial terms, per layer (means over splits; error bars are the SD of the within-run gain). \textbf{(b)} Transformer minus no-sequence-model, paired within split, for the overall gain, the within-run gain and the position-corrected within-run gain. Panel (a) shows that the non-neural position-only term is flat across depth, so the depth profile of Figure~\ref{fig:word} is explained by the EEG terms; (b) shows that the transformer's advantage is large on the overall gain, where it is passage tracking, small on the position-corrected word-level metric, and slightly negative at layer 0, where the target carries no context for it to exploit.}
  \label{fig:probes}
\end{figure}

The transformer's within-run gain decomposes into the three nested probe terms at every layer (Figure~\ref{fig:probes}a). The non-neural position-only term is $0.0 \pm 0.2$\,pp at layer 0 and between 2.3 and 3.6\,pp everywhere else, without a trend across depth ($2.9 \pm 0.6$\,pp at layer 2, $2.4 \pm 0.3$\,pp at layer 20), the depth profile of Figure~\ref{fig:word} is explained by the neural terms. The preceding-EEG term is largest at the early layers ($4.0 \pm 0.4$\,pp at layer 2) and 2--3\,pp in the deep layers; the current-trial term falls from $10.0 \pm 0.3$\,pp at layer 2 to $5.9 \pm 0.4$\,pp at layer 32.

What the sequence model adds depends on the metric (Figure~\ref{fig:probes}b; differences paired within seed). On the overall gain its advantage is large and depth-dependent, from $17.6 \pm 1.3$\,pp at layer 2 through $20.1 \pm 1.7$\,pp at layer 8 to $10.6 \pm 2.1$\,pp at layer 32. The sequence model's larger overall gain is mostly passage tracking. On the within-run gain the sequence model's advantage is $4$--$5.5$\,pp across most of the depth, and once the positional shortcut is removed it shrinks to $2.3 \pm 0.6$\,pp at layer 2, $1.9 \pm 0.4$\,pp at layer 20 and $0.4 \pm 0.6$\,pp at layer 32. At layer 0, where the target carries no context, the sequence model is slightly harmful ($-0.4 \pm 0.3$\,pp), reproducing report 1's interaction between target type and architecture along the whole depth axis.

\subsection{What a single window shares with its passage}
\label{sec:results-window}

\begin{figure}[tp]
  \centering
  \includegraphics[alt={Swap gain against lag for selected layers and both decoders with the target-geometry control, and swap gain for near, far and random run-mates against layer},width=\textwidth]{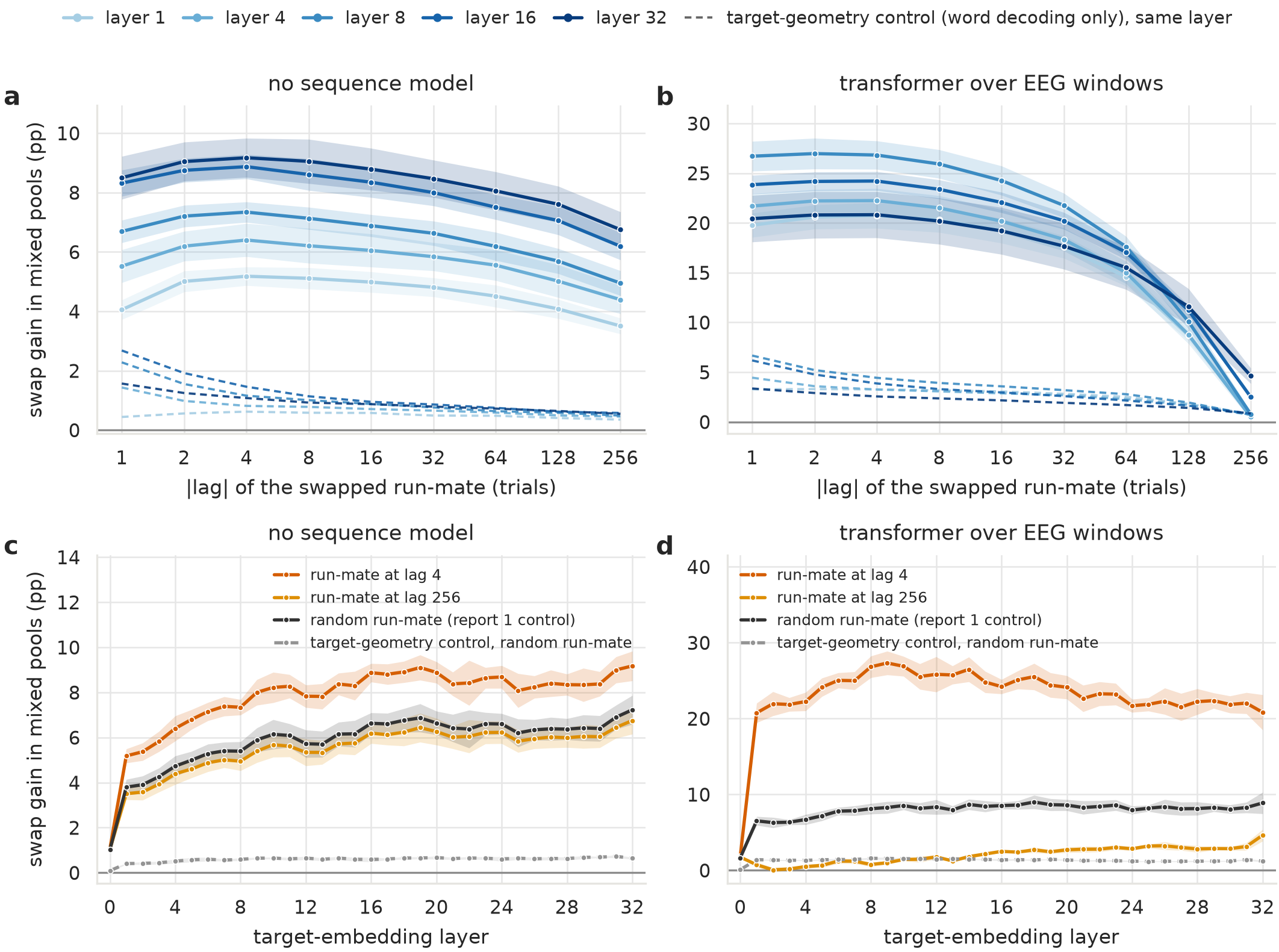}
  \caption{\textbf{Lag-dependent run-mate swap and the target-geometry control.} \textbf{(a, b)} Cross-run gain after replacing each trial's prediction with that of the run-mate $|k|$ trials away (both directions averaged), for five layers, \textbf{(a)} without and \textbf{(b)} with the sequence model; dashed lines are the target-geometry control at the same layers. The profile shows how far along the run the information shared between trials reaches: a flat profile is a component that the whole run shares, a decaying one a local component, and the target-geometry control marks what word decoding with no context in the EEG would produce at each lag. Both without (a) and with (b) a sequence model, the profiles lie far above the control, so the decoder's predictions share something that the geometry of the targets does not explain. Without a sequence model the profile is comparatively flat (note the different y-axis scales in (a) and (b)), so the shared component is run-level. With the sequence model the profile falls steeply with the lag of the swapped trial, so its context is local. \textbf{(c, d)} Against layer: the swap gain for a run-mate 4 and 256 trials away, the random-run-mate control of report 1, and the target-geometry control's random swap. These panels show that the ordering of (a, b) holds at every layer, and that the run-level component grows with depth while the control stays at 1\,pp or less without a sequence model and below 2\,pp with it. Means and SD over splits.}
  \label{fig:lag}
\end{figure}

\textbf{Word decoding alone does not produce the single window's context-tracking gain.} Section~\ref{sec:results-context} left open whether the context-tracking gain of the decoder without a sequence model could be a by-product of word decoding and target geometry (mechanism 1 of Section~\ref{sec:lag}). Supplementary Table~\ref{tab:lag} and Figure~\ref{fig:lag} compare each checkpoint with its target-geometry control, a pure word decoder calibrated to the same within-run gain. Without a sequence model the control's random-swap gain is 0.4--0.7\,pp at every layer above 0, against 3.8--7.2\,pp for the real decoder. Roughly nine tenths of the context-tracking gain therefore comes from something in the predictions themselves that the geometry of the targets does not explain, a component that the predictions for the trials of one run share. In the following, we will address how far along the run that shared component extends. At layer 0 the control gives 0.1\,pp: target geometry contributes nothing there, as expected for context-free targets, whose projected vectors are barely more similar within a run than between runs (mean cosine similarity 0.123 against 0.116 in the decoder's projected space). The decoder nevertheless gives 1.0\,pp, because its predictions cluster by run far more than the targets do (0.09 within against 0.03 between runs). Hence, a run-level component is present in the predictions even when the target has no context, and it earns a small gain through the weak vocabulary clustering of the targets.

\textbf{Without a sequence model, the shared component is mostly constant within a run.} The swap gain decays only gently with the lag (Figure~\ref{fig:lag}a): at layer 8 it is $7.3 \pm 0.3$\,pp for a run-mate four trials away and $4.9 \pm 0.4$\,pp for one 256 trials away, against $5.4 \pm 0.4$\,pp for a random run-mate and $0.6 \pm 0.1$\,pp for the target-geometry control. We read the value for the farthest run-mate as a run-level part that all trials of the run share, and the excess of the near over the far run-mate, $7.3 - 4.9 \approx 2.4$\,pp at layer 8 and 1.7--2.7\,pp across layers 1--32, as a local part. The control shows the same kind of decay at only 0.3--0.9\,pp, so most of the local part, too, goes beyond target geometry. The full set of profiles is in Supplementary Figure~\ref{fig:s3}.

\textbf{With the sequence model, the shared component is local.} The transformer's swap gain is large for close run-mates ($26.8 \pm 1.4$\,pp at layer 8 and lag 4, against an overall gain of $38.3 \pm 1.9$\,pp) and falls steeply beyond a lag of about 64 trials, to $0.8 \pm 0.3$\,pp at lag 256 (Figure~\ref{fig:lag}b, Supplementary Figure~\ref{fig:s4}). The sequence model's context has a characteristic length of a few dozen words. The random swap, which averages over all lags, understates how local it is.

\begin{figure}[tp]
  \centering
  \includegraphics[alt={Retained fraction of the cross-run gain after taking a narrative neighbour's predictions, same versus different session, matched versus scrambled, and regression coefficients of run-centroid similarity},width=\textwidth]{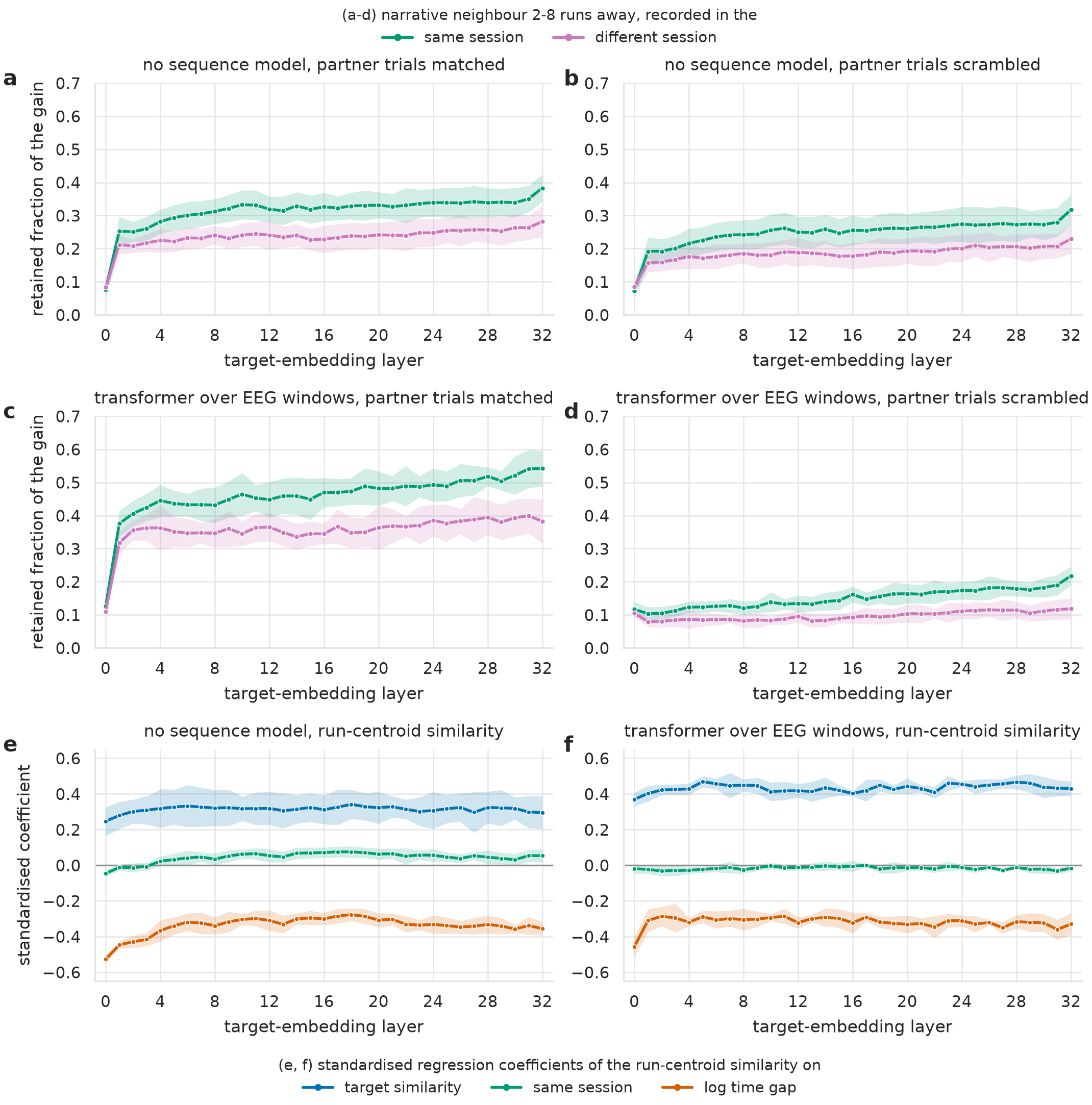}
  \caption{\textbf{Probing potential confounds that could explain the observed context-tracking gain.} The context-tracking might be caused by a recording-session signature (mechanism 3 in \ref{sec:lag}) or by the similarity of target embeddings for adjacant words (mechanism 1 in \ref{sec:lag}). \textbf{(a--d)} Fraction of a validation run's cross-run gain retained when its trials take the predictions of the validation run 2--8 places away in the narrative, for neighbours recorded in the same (green) or in a different (purple) session, with the partner's trials position-matched (a, c) or scrambled (b, d), without (a, b) and with (c, d) the sequence model. Panels (a, c) test whether what neighbouring runs share follows the text or the recording session: a session signature would make the different-session curve collapse to zero, whereas it stays within about a quarter of the same-session curve at every layer. Panels (b, d) show how much of the shared component is tied to position within the run, by scrambling the partner's trial order: little without a sequence model, most of it with the transformer, whose positional embedding the probes of Figure~\ref{fig:probes} already measure. \textbf{(e, f)} Standardised coefficients of the cosine similarity between two validation runs' mean predictions regressed on the similarity of their mean targets, a same-session indicator and the log gap in recording time, over all pairs of validation runs. These panels weigh the three candidate sources of run-to-run similarity against each other: the text carries most of the weight, the same-session indicator little on this standardised scale (Supplementary Table~\ref{tab:session}), and the time gap a negative weight that is present even at layer 0, where the targets are context-free, and is therefore a slow drift along the recording axis rather than passage tracking through the targets. Means and SD over splits.}
  \label{fig:session}
\end{figure}

\textbf{Most of the run-level component follows the text, not the session.} The lag swap cannot tell a decoded discourse state from a recording-session signature, because both are constant within a run; narrative neighbours recorded in different sessions can (Section~\ref{sec:session}). When a validation run takes the predictions of a narrative neighbour two to eight runs away, the fraction of its cross-run gain that is retained is $0.32 \pm 0.03$ for neighbours recorded in the same session and $0.24 \pm 0.02$ for neighbours recorded in a different session without a sequence model, and $0.47 \pm 0.04$ versus $0.36 \pm 0.02$ with it (mean $\pm$ SD over layers 1--32 of the split means; Supplementary Table~\ref{tab:session}, Figure~\ref{fig:session}a,c). A recording-session signature would collapse at the boundary; instead, neighbours recorded in a different session retain about three quarters of what neighbours recorded in the same session retain, with either decoder (0.75 without and 0.77 with the sequence model). So about a quarter of the context tracking component may be session-linked, and the rest follows the text. Scrambling the partner run's trial positions removes about a fifth of the retained fraction without a sequence model ($0.32 \to 0.25$) and about two thirds of it with the transformer ($0.47 \to 0.15$), whose positional embedding carries a within-run position component that the probes of Section~\ref{sec:results-probes} already measure (Figure~\ref{fig:session}b,d).

The regression over pairs of validation runs gives the same picture (Figure~\ref{fig:session}e,f). It asks what makes the decoder's mean predictions for two runs alike, and weighs three candidate sources against each other: the similarity of the two runs' mean targets (the text), whether the runs were recorded in the same session, and how far apart in time they were recorded. Three results follow; see Supplement for details (Section~\ref{sec:s-regression}, Supplementary Table~\ref{tab:session}). First, the text carries most of the weight: its standardised coefficient is $0.32 \pm 0.01$ without and $0.44 \pm 0.02$ with the sequence model (mean $\pm$ SD over layers 1--32 of the split means). Second, the session contributes little. Without a sequence model, two runs from the same session are slightly more alike than their text and their recording-time gap predict, by about 0.09 in cosine similarity (on its scale from $-1$ to 1); the effect is absent at layers 0--2, grows with depth, and exceeds a permutation null in about half of the checkpoints, mostly at layers 9--32. With the sequence model there is no such effect. Third, the time gap carries a negative weight of about $-0.3$ for both decoders: predictions for runs recorded months apart are less alike than their text similarity predicts. This drift is present even at layer 0, where the targets are context-free, so it is not passage tracking through the targets. Because the participant read the corpus in recording order, recording time and position in the corpus are nearly the same axis, and the size of the drift term, though not its sign, depends on how the text is controlled for (Supplement, Section~\ref{sec:s-regression}). The drift is a slow change of the prediction geometry along the recording axis, whether of the recordings themselves or of a state that follows the reader's progress through the corpus, and it is not the carrier of the context-tracking gain, most of which survives the session boundary. In sum, the decoder without a sequence model carries a small session-linked component at the deeper layers and the transformer does not, and in both cases it is small next to the text and the drift.

\subsection{A restricted candidate set}
\label{sec:results-small-pool}

For comparison with closed-vocabulary speech neuroprostheses, the Supplement (Section~\ref{sec:s-small-pool}, Figure~\ref{fig:s5}, Table~\ref{tab:small-pool}) re-scores every fit on a fixed list of 50 words, as a 50-way identification task with uniform chance of 2\% (top-1) and 20\% (top-10), together with two controls that measure how much of the closed-set accuracy comes from the passage rather than from the word. Against the context-free layer-0 target, where only the word can be read, a single 0.5-s window identifies one of 50 content words at $5.1 \pm 0.7$\% top-1 and $33.8 \pm 1.2$\% top-10 without a sequence model, and at $4.8 \pm 0.3$ and $34.0 \pm 1.3$\% with it, and both controls sit near chance. Against contextual targets the closed-set accuracy jumps, to $12.8 \pm 0.5$\% top-1 at layer 1 without a sequence model and to $33.0 \pm 2.0$\% at layer 8 with it, but so does the control that replaces each prediction with that of the run-mate one trial earlier, which carries the local discourse state but not the current word: at layer 8 it recovers $27.9 \pm 1.3$ of the transformer's 33.0\%. The word-level part of the closed-set accuracy, the excess over that control, is largest at layers 1--2 ($5.8 \pm 1.1$\,pp without and $7.7 \pm 0.7$\,pp with the sequence model, top-1) and declines to about 2.5\,pp at the last layer: the profile of Figure~\ref{fig:word}, in closed-set form. The 50 most frequent words give the same picture.

\subsection{Sanity checks}
\label{sec:results-checks}

Every fit converged within the 100-epoch budget (Supplementary Figure~\ref{fig:s1}); checkpoints were selected at epoch 58--92 on average per layer, and no fit approached the temperature clamp (Supplementary Figure~\ref{fig:s2}). The lag-swap reconstruction reproduced the logged overall and context-tracking gains of every checkpoint to within 0.07\,pp.

\section{Discussion}
\label{sec:discussion}

\subsection{Summary}
\label{sec:summary}

Across the full depth of Llama-3.1-8B, the EEG-decodable structure of silent reading has a simple distribution. Word identity is easiest to read out against the early layers, a few blocks in, and fades steadily with depth. What deeper layers add is predicting the text passage. The context-tracking component of the retrieval gain and its share of the total rise with depth for both decoders. Rare and mid-frequency words are decoded best against mid-depth targets, where the frequent words have already begun to decline. The profile replicates across five train/validation splits, holds after subtracting the positional shortcut, and is the same in shape for a decoder that integrates over the sequence of EEG windows and for one that sees one window at a time.

The second result we would like to highlight concerns that single window. A decoder that has no access to neighbouring windows tracks the passage once the target carries context, and the two controls introduced here show what that is and is not. The passage tracking is not a by-product of word decoding and target geometry: a pure word decoder calibrated to the same word-level accuracy produces a tenth of it. It is largely constant within a run, with a small local part. And most of it follows the text rather than the recording session: narrative neighbours recorded in a different session retain about three quarters of what neighbours recorded in the same session retain. The regression over run pairs attributes the missing quarter to a slow drift along the recording axis and, for the decoder without a sequence model only, in part to a small session-linked component at the deeper layers; for the transformer it finds no session-linked component at all. The half-second EEG response to a word therefore carries information about the passage being read, which a contextual target can absorb.

\subsection{Early layers hold the word, deep layers hold the discourse}
\label{sec:discussion-depth}

The depth profile is what one would expect if the EEG response to a word carries the word's identity together with a weaker trace of its context. That the context-tracking gain approaches the word-level gain in percentage points at the deeper layers does not contradict this: telling apart the roughly 85 validation runs that feed a cross-run pool is a far easier discrimination than telling apart the 250--300 words of one passage, so a weak contextual signal buys a large gain. Layer-wise dissections of language models find that the lowest hidden layers are the least context-specific and the top layer of a causal model the most \citep{ethayarajh2019contextual}, that lexical information is read out early and context-dependent structure late \citep{tenney2019bert}, and that under a next-token objective information about the current token fades with depth while information about the next token builds up \citep{voita2019bottom}. On that reading, a layer-1 or layer-2 hidden state is still dominated by the input token, mixed with a small amount of the preceding text. A mid-depth state carries a substantial share of the preceding passage. The final state is specialised for next-token prediction. The word-level metric, which by construction cannot be earned from passage identity, tracks the word's identity and therefore peaks where the hidden state is still mostly the input token. The context-tracking metric rises as the hidden state absorbs more of the preceding text. The profiles differ from those typically reported for fMRI and MEG encoding models, where alignment follows an inverted U across depth and peaks in the middle layers, at least for contexts longer than about 15 words \citep{toneva2019interpreting, caucheteux2022brains}, but the quantities differ too: an encoding model predicts a slow haemodynamic signal, or an MEG response averaged over time, from the target, whereas our primary metric asks how well a half-second EEG window discriminates among the words of one passage, a question about the word rather than about the brain state. Consistent with this, the word-by-word ECoG dataset analysed by \citet{schrimpf2021neural} was better predicted by earlier layers of GPT-2 XL than their fMRI datasets were, a difference the authors suggested reflects the prediction of individual word representations rather than whole-sentence representations.

Three implications follow for the design of EEG language decoders. First, the target layer is a dial between ``which word'' and ``which passage''. Early-to-mid layers are the place for word identity, deeper layers for discourse state, and a deployed decoder probably wants both, in the same way that invasive speech neuroprostheses combine the neural evidence for each word with a language-model prior over the words already decoded in the sentence \citep{moses2021neuroprosthesis, metzger2023high, card2024accurate}. Training against an early-layer target for the word and a deeper-layer target for the context is an obvious next step. Second, the context-free layer 0, the cleanest choice for a control analysis, is the worst choice for rare content words, which prefer context-rich targets. Third, the sequence model's advantage is mostly passage tracking; its contribution to word-level decoding, once the positional shortcut is removed, is 1--2\,pp and largest at the early layers.

\subsection{Context in a single window}
\label{sec:discussion-window}

That the evoked response to a word depends on its context is well established at the level of single EEG components \citep{kutas2011thirty}; the present result is that a decoder can read enough of that dependence from a 0.5-s window to recognise the passage in a held-out run, and that this is largely not an artefact of a recording-session signature. Two loopholes remain. First, the target-geometry control assumes isotropic errors around the true target, whereas a real decoder's errors are structured, so it is one instance of the word-decoding mechanism rather than a reproduction of it. Errors that fall towards semantically or contextually similar words rather than in random directions would give a pure word decoder a larger cross-run gain than isotropic errors do, and the decoder's error structure cannot be measured independently of the run-level component we are trying to isolate, so the control is an estimate of the mechanism's contribution, not a bound on it. Second, the session-boundary test limits slow recording artefacts to a minority of the component, and even that shortfall is not necessarily explained by an EEG session signature, since a discourse state may simply have cooled over a break of hours or days. The regression over run pairs puts a number on the part that is tied to the session itself: about 0.1 in cosine similarity between run centroids (on its scale from $-1$ to 1) for the decoder without a sequence model at the deeper layers, and nothing detectable for the transformer. What the test does not exclude is a potential feature of the stimulus stream that is linked to the text, visible in a single window, and non-semantic. Randomised typography attenuates the obvious visual candidate, and position scrambling removes progress through the run, but a residual feature of that kind cannot be excluded within a reading paradigm. The passive-listening data planned for the next stage of this project, in which no such visual feature exists, is the natural test.

\subsection{Closed-set accuracy, and what it measures}
\label{sec:discussion-small-pool}

Imposing the vocabulary at the language-model stage, downstream of a vocabulary-agnostic phoneme decoder, is how the invasive literature obtains its 50-word numbers \citep{willett2023high, card2024accurate}, and the Supplement restricts the candidate set at inference in the same spirit (Section~\ref{sec:s-small-pool}). The restricted numbers are modest where they are clean. Against the context-free target, a single EEG window identifies one of 50 words at a few times chance in top-1. They become large as soon as the target carries context, and the controls show why. In a cross-run pool the competitors come from other passages, so the discourse state identifies the true candidate, and the earlier-neighbour swap, which keeps that state and removes the word, recovers most of the closed-set accuracy of the deeper layers. The depth profile of Section~\ref{sec:results} thus reappears in closed-set form, with early layers for the word and deeper layers for the passage. It also carries a warning for closed-set evaluations that align brain data with contextual language-model targets: without a control that preserves the local context and removes the word, a 50-way accuracy conflates the two.

\subsection{Limitations}
\label{sec:limitations}

All results come from one densely sampled participant. Five seeds give error bars but not a large sample; the smoothness of the curves across adjacent layers is the stronger evidence for the profile's shape. All values are computed on the validation set that also drove checkpoint selection, so levels are upper bounds and orderings are the reliable output. The validation runs are also interleaved in time with the training runs, and the time-gap term of Section~\ref{sec:session} shows that the prediction geometry drifts over the months of recording, so the levels are optimistic for use on a later point in time. The layer index is confounded with the geometry of the hidden states as the decoder sees them through a linear projection (their anisotropy changes with depth, and the final layer sits after the model's last normalisation), so a layer effect is an effect of the representation as presented to the decoder, not a direct statement about the language model's semantics. The dataset is slightly larger than report 1's (409 versus 393 runs), which is why the values at layers 0 and 20 differ slightly from those reported there.

\subsection{Conclusion}
\label{sec:conclusion}

Word-level information in EEG during silent reading aligns best with the early layers of a language model and passage-level information with its deeper layers, rare words prefer the middle. A single half-second window already carries a trace of the passage that mostly follows the text rather than the recording session. Even a decoder without a sequence model that sees only a 0.5-s EEG window performs some context tracking. Adding a sequence model to the decoder, so that it can see preceding EEG segments, substantially increases the overall gain, but most of that advantage is due to the sequence model's passage tracking ability. That is not to say that context tracking is undesirable; on the contrary, explicitly leveraging both word-level decoding and context tracking might prove useful for (non-invasive) language decoding.

\bibliographystyle{apacite}
\bibliography{references}

\clearpage
\appendix
\setcounter{section}{-1}
\renewcommand{\thesection}{S\arabic{section}}
\setcounter{table}{0}
\renewcommand{\thetable}{S\arabic{table}}
\setcounter{figure}{0}
\renewcommand{\thefigure}{S\arabic{figure}}

\begin{center}
  {\LARGE\bfseries Supplementary Material}
\end{center}

\section{Implementation notes}
\label{sec:s0}

The lag-dependent run-mate swap and the target-geometry control of Section~\ref{sec:lag} answer two different questions. The swap asks how far along a run the information shared between the decoder's predictions reaches: each trial's prediction is replaced by that of a run-mate a fixed number of trials earlier or later, and the cross-run gain that survives the swap is read against that lag, so that a component shared by the whole run (a discourse state or a recording-session signature) shows as a flat profile, whereas a local component decays with the lag. The target-geometry control asks how much of that swap gain word decoding alone would produce: it is a synthetic predictor made of each trial's true target plus noise, with the noise calibrated so that its word-level accuracy matches the decoder's, and it contains no information from the EEG, so any swap gain it shows comes from the similarity between the targets of neighbouring words, and only the decoder's excess over it requires context in the EEG. The notes below give implementation details of both, and of the narrative-neighbour analysis of Section~\ref{sec:session}.

\paragraph{Lag swap at run boundaries} A trial whose partner at lag $k$ would fall outside the run takes the partner in the other direction, so that $|k|$ is exact wherever the run is longer than $2k$; validation runs contain about 550 included trials, so at $k = 256$ roughly half of the trials are reflected and none falls back to a shorter lag. Directions were also analysed separately, and the figures average the two. Without a sequence model the earlier and later profiles differ by at most 1.1\,pp at any lag and layer. With the sequence model the later partner is the better one from a lag of about 4 trials on, by up to 6.8\,pp at lags 32--64, because its prediction has already integrated the current trial's EEG window through the transformer.

\paragraph{Short lags} The swap profiles are not monotone at the shortest lags. Without a sequence model the gain for the immediate neighbour is lower than for a run-mate four trials away at every layer, by 0.7\,pp on average, and with the sequence model at 32 of the 33 layers, by 0.3\,pp; the target-geometry control, which shares the targets' geometry but was never trained, is highest at lag 1 at nearly every layer above 0. We attribute the dip to the training objective: run-mates are in-batch negatives, a trial's immediate neighbours have the most similar targets and are therefore its hardest negatives, and the decoder is thus trained to discriminate a trial from its neighbours in particular. At layer 0 even the control is below zero at lag 1, because adjacent words are less similar in the input embedding than two words drawn at random from the same run (consecutive words tend to differ in word class).

\paragraph{Calibration of the target-geometry control} The control's prediction for a trial is the trial's true projected target plus isotropic Gaussian noise with standard deviation $\sigma$ per dimension, re-normalised to unit length; the same noise seed is used throughout, so that the control's within-run gain is a deterministic function of $\sigma$. That function decreases monotonically with $\sigma$, from perfect retrieval at $\sigma = 0$ to chance at large $\sigma$, so a unique $\sigma$ matches any attainable gain, and $\sigma$ was found by bisection on the interval $[0, 8]$. At each of 14 steps the control was built at the midpoint of the current interval and scored on the within-run gain with the same pool draws, permutations and learned temperature as the checkpoint; if its gain exceeded the checkpoint's, the noise was too small and the upper half of the interval was kept, otherwise the lower half. After 14 halvings the interval is $8 / 2^{14} \approx 5 \times 10^{-4}$ wide, and the calibrated within-run gain matches the checkpoint's to within 0.04\,pp in every case (Supplementary Table~\ref{tab:layer-lag}). The calibrated $\sigma$ is then used for every swap of the control.

\paragraph{Run index} The 409 runs were ordered by recording time, which is the narrative order. There were 52 sessions on 42 days, about 8 runs per session, 146 days between the first and last; the longest break between two sessions was 23 days. Each split's validation set contained 85--86 runs spread over 38--40 sessions, yielding 66--104 same-session and 60--120 cross-session neighbour pairs at shifts 2--8 per split (each unordered pair counted once in each direction).

\paragraph{Retained fraction} The retained fraction measures how much of a validation run's cross-run gain survives when its trials take the predictions of a narrative neighbour. It is built from the per-trial quantities behind the overall gain of the main text. For every trial, the \emph{top-10 hit} is 1 if the trial's true word is among the ten highest-ranked candidate words of the trial's random cross-run pool (word-grouped, as in the main text) and 0 otherwise; averaged over trials, this is the top-10 accuracy. The \emph{permutation-baseline hit rate} of a trial is the fraction of the 10 within-pool permutations of the predictions in which the true word still lands in the top 10, that is, the chance level of that trial. The top-10 hit minus the permutation-baseline hit rate, averaged over a set of trials, is the gain over the permutation baseline in which all metrics of this report are expressed. For a run $r$ with partner $p$, the \emph{swapped gain} is this difference averaged over $r$'s trials after every prediction of $r$ has been replaced by a prediction of $p$ (position-matched: trial $m$ of $r$ takes trial $m$ of $p$, starting again from $p$'s first trial when $p$ is the shorter run; scrambled: a random trial of $p$), and the \emph{unswapped gain} is the same difference averaged over the same trials with $r$'s own predictions. Both are scored in the random cross-run pools of the overall gain, with the same pool draws and the same permutations, so that the two differ only in the predictions. The retained fraction is the swapped gain divided by the unswapped gain: a value of 1 means that the partner's predictions earn the run as much gain as its own, a value of 0 that they earn nothing. Within a checkpoint, the swapped and the unswapped gains were each averaged over all run pairs of the same kind (shifts 2--8 pooled, separately for same-session and different-session partners and for position-matched and scrambled trials), with each pair weighted by the number of trials in $r$, and the ratio of these two averages is the checkpoint's retained fraction; the figures and tables report its mean and SD over seeds.

\section{Run-to-run similarity of the predictions}
\label{sec:s-regression}

Section~\ref{sec:session} summarises a regression that attributes the similarity between two runs' predictions to the text, the session and the recording-time gap.

\paragraph{Quantities} For each validation run $r$ of a checkpoint, the mean of its trials' predictions in the shared 256-dimensional space, re-normalised to unit length, is the run's prediction centroid $\bar{p}_r$, and the mean of its trials' projected targets, likewise normalised, is its target centroid $\bar{t}_r$. For every pair of validation runs $(a, b)$ the prediction similarity is $\cos(\bar{p}_a, \bar{p}_b)$ and the target similarity is $\cos(\bar{t}_a, \bar{t}_b)$. With 85--86 validation runs per split there are 3{,}570--3{,}655 pairs, of which 79--97 (depending on the split) are pairs of runs recorded in the same session.

\paragraph{Why centroids} Averaging over the trials of a run removes the word-level content of the predictions, which changes from trial to trial, and leaves what the predictions of a run have in common: the run-level component that the lag swap identified. The target centroid plays the same role for the text: two runs whose target centroids are similar are about similar things in the language model's space.

\paragraph{Model} The prediction similarity is regressed by ordinary least squares, with an intercept, on three predictors: the target similarity, an indicator that is 1 when the two runs were recorded in the same session and 0 otherwise, and $\log(1 + \Delta)$, where $\Delta$ is the number of days between the two recordings (a fraction of a day for same-session pairs, up to 146). The logarithm compresses the strongly skewed distribution of gaps. Response and predictors are standardised to zero mean and unit variance, so each coefficient is the change in prediction similarity, in standard deviations, per standard deviation of the predictor with the other two held fixed. The model is fitted separately for every checkpoint (layer, decoder and split); Figure~\ref{fig:session}e,f shows the mean and SD over splits per layer, and Supplementary Table~\ref{tab:session} the mean and SD over layers 1--32.

\paragraph{Reading the coefficients} The target-similarity coefficient measures how much of the run-to-run similarity of the predictions follows the text. It is expected to be positive for any decoder that recovers word identity at all, because the mean of noisy word predictions tracks the run's mean target; its role here is to let the other two terms be read net of the text. A recording-session signature would make two runs from the same session more alike than their text and time gap predict, and would appear as a positive same-session coefficient. A slow drift of the recordings, or of the decoder's mapping of them, would make runs recorded far apart less alike and would appear as a negative time-gap coefficient. The same-session indicator and the time gap are related, since a same-session pair has a gap of a few hours at most, but they are not collinear, because the cross-session pairs span the full range of gaps. The same-session coefficient is the deviation of the same-session pairs from the log-gap line extrapolated from the cross-session pairs, and therefore isolates a boundary step over and above the smooth drift, to the extent that the logarithmic form describes the drift at short gaps. The drift term is largest at layer 0 (about $-0.5$), where the targets carry no context, and about $-0.3$ at the deeper layers.

\paragraph{The same-session term on its own scale} The standardised coefficient of a binary predictor is its per-pair effect times the predictor's SD divided by the response's SD. The same-session indicator marks 79--97 of 3{,}570--3{,}655 pairs and has an SD of 0.15, so its standardised coefficient is six to seven times smaller than the extra similarity of a same-session pair in SD of the response, and the common scale of Figure~\ref{fig:session}e,f understates it next to the two continuous predictors. Supplementary Table~\ref{tab:session} therefore adds three readings. First, the same model with the indicator left on its 0/1 scale, whose coefficient is the extra similarity of a same-session pair in SD of the response. Second, a direct contrast that assumes no drift curve: same-session pairs minus different-session pairs recorded on the same day (29--65 per split; the cap was taken off and put back between them), after regressing the response on the target similarity alone. Third, a block-permutation null for the standardised same-session term, which asks whether a coefficient of the observed size could arise from any division of the time-ordered runs into contiguous blocks, that is, from temporal contiguity alone rather than from the actual session boundaries. The 52 session sizes are kept and their order is shuffled over the time-ordered runs, so that a permuted session is still a contiguous block of runs, with 1{,}000 draws per checkpoint and a two-sided $p$. Permuted and real partitions overlap, so the test is conservative; the null's mean is not zero (about $+0.02$ without and $-0.02$ with the sequence model), which shows that part of a raw same-session coefficient is generic contiguity rather than a boundary step. Without a sequence model the three readings agree: $0.31 \pm 0.15$ SD on the 0/1 scale (SD over layers; about 0.20 over splits), rising with depth from near zero at layers 0--2 to about 0.4 SD at layers 16--32; a same-day contrast of $0.60 \pm 0.10$ SD, larger than the model-based reading because the logarithmic drift absorbs part of the step; and $p < 0.05$ in 52\% of the checkpoints, 23\% at layers 1--8 and 62\% at layers 9--32, in about four fifths of the layers in three of the five splits and in at most a sixth in the other two. With the sequence model the three readings are $-0.11 \pm 0.06$ SD on the 0/1 scale, a same-day contrast of $0.19 \pm 0.08$ SD, and $p < 0.05$ in 6\% of the checkpoints, the chance rate. The decoder without a sequence model therefore carries a small session-linked component at the deeper layers, of the order of 0.1 in cosine between run centroids, and the transformer does not.

\paragraph{Caveats} Each run enters 84--85 pairs, so the pairs are not independent observations, and no confidence intervals or tests are reported; the across-split SD is the only error bar. Target similarity and time gap are themselves correlated (Spearman $\rho$ between $-0.4$ and $-0.7$, depending on layer and decoder), because the text of runs recorded far apart differs more, and within a passage the log time gap and the log narrative distance in runs correlate at 0.88, because the corpus was read in recording order. The division of weight between the text and the time gap is therefore only as good as the linear model, and the size of the time-gap term depends on what else is in it. To test how much of it is carried by narrative proximity rather than by recording time, two further text controls were added: a same-passage indicator, which is 1 when both runs lie in the same continuous stretch of text (no change of source text between them), and the log of the two runs' distance in the narrative, counted in runs within that stretch. The time-gap coefficient is $-0.33$ and $-0.31$ in the three-predictor model, $-0.25 \pm 0.05$ and $-0.28 \pm 0.03$ with the two text controls added, and $-0.10 \pm 0.04$ and $-0.12 \pm 0.05$ when the model is fitted to within-passage pairs only, with the narrative distance alongside (Supplementary Table~\ref{tab:session}). Its sign is robust, its size is not, and the regression cannot separate a drift of the recordings from a slowly varying component that follows the reader's progress through the corpus; the layer-0 result excludes only passage tracking through the targets. The same-day contrast and the permutation null of the same-session term do not depend on this division.

\section{Restricted candidate set}
\label{sec:s-small-pool}

\begin{figure}[htb]
  \centering
  \includegraphics[alt={Closed-set accuracy on fixed 50-word lists against layer, with the two swap controls and the restricted within-run gain},width=\textwidth]{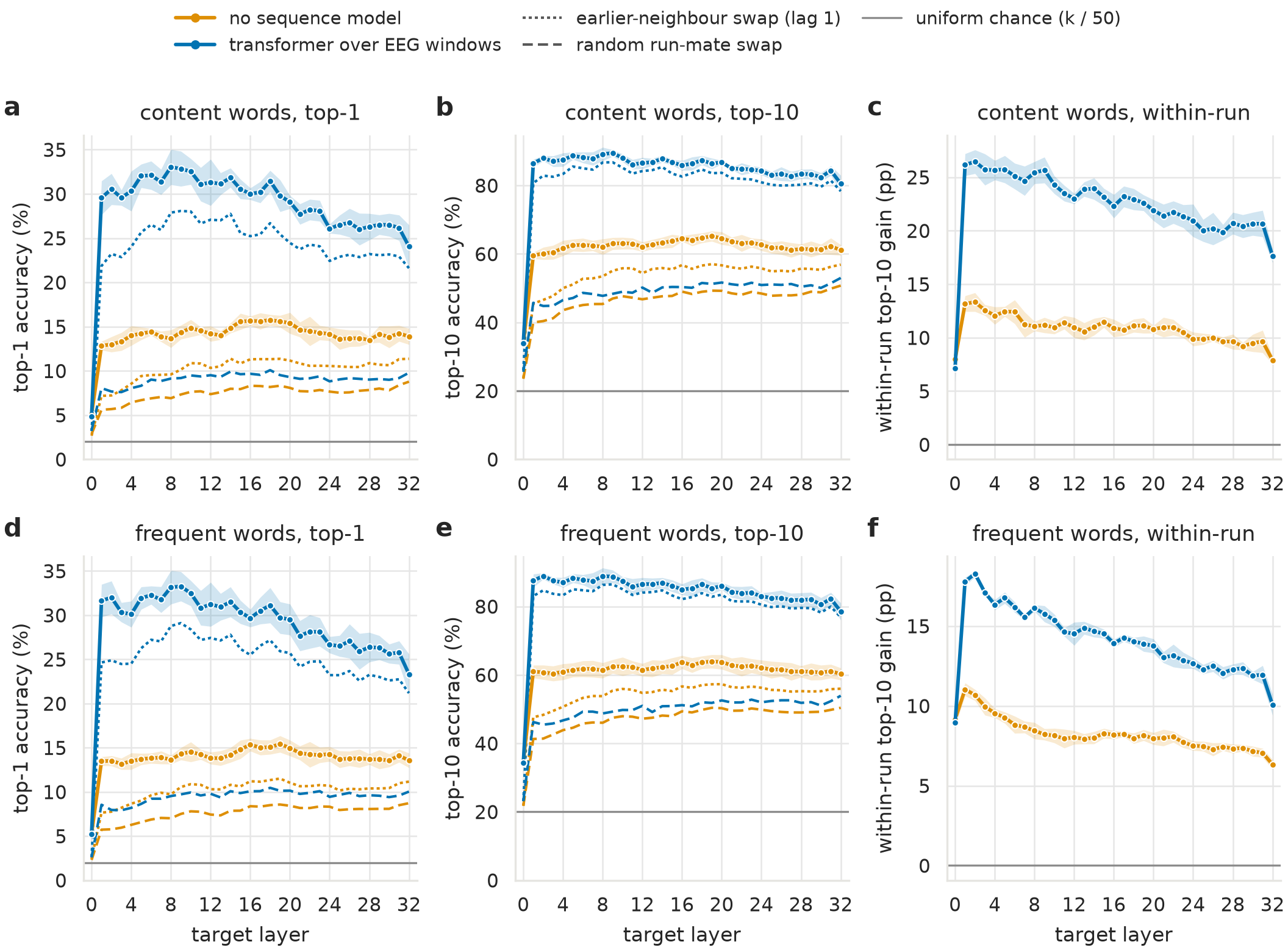}
  \caption{\textbf{Restricted candidate set across layers.} The validation predictions of the main text, scored on pools that hold one trial of each of 50 fixed words (Section~\ref{sec:s-small-pool}): the 50 most frequent content words (\textbf{a--c}) and the 50 most frequent words of any kind (\textbf{d--f}). \textbf{(a, d)} Top-1 and \textbf{(b, e)} top-10 accuracy of the 50-way identification against the target layer, for both decoders (solid, mean and SD over splits), with uniform chance (2\% and 20\%) as the horizontal line. The dotted and dashed lines are the two controls scored on the same pools: every prediction replaced by that of the run-mate one trial earlier (local discourse state and position, no current word) or of a random run-mate. At layer 0 the targets carry no context and both controls sit within a few points of chance (Table~\ref{tab:small-pool}), so the accuracy there is word identification; the small excess of the controls over chance comes from the run-level component of the predictions, which is present even at layer 0 (Section~\ref{sec:s-small-pool}). With depth the earlier-neighbour control rises with the accuracy itself: in a cross-run pool a contextual target is found through the discourse state, and the word-level part of the closed-set accuracy is the distance between the solid and the dotted line. \textbf{(c, f)} Report 1's within-run gain restricted to the list words (pools are the list trials of one run), where candidates from other passages are absent; chance is 0\,pp. For the transformer these values are not position-corrected and are upper bounds on word-level decoding (Section~\ref{sec:s-small-pool}).}
  \label{fig:s5}
\end{figure}

The retrieval metrics of the main text rank the true word among the 250--300 distinct words of an open-vocabulary pool, which cannot be compared directly with the accuracies reported for closed-vocabulary speech neuroprostheses, where one word of a fixed set of typically 50 is identified. The purpose of this analysis is to re-score the same predictions, without retraining, as a 50-way identification of a word from a fixed list, so that our numbers can be read in that form, and to add two controls that measure how much of the closed-set accuracy comes from the passage rather than from the word, because the candidates of such a pool come from different passages and a contextual target can therefore be identified through the discourse state alone. Section~\ref{sec:small-pool} summarises the design; this section gives the rules, the scoring and the results (Figure~\ref{fig:s5}, Tables~\ref{tab:small-pool} and~\ref{tab:word-lists}).

\paragraph{Word lists} Both lists were built from the full corpus of the participant, i.e.\ every presentation of every run, with words reduced to the punctuation-stripped, lower-cased form used for word grouping throughout. The \emph{content} list holds the 50 most frequent words that are alphabetic, not in the English stop-word list of scikit-learn, and not proper names, a word being counted as a proper name when it was capitalised in more than 90\% of its presentations (which removes, for example, the characters' names and the title ``Mr''). The \emph{frequent} list holds the 50 most frequent words of any kind. Table~\ref{tab:word-lists} lists both with their corpus counts. The content words account for about 5\% of all presentations and the frequent words for about 44\%.

\paragraph{Pools and scoring} For a given fit, the trials of the validation split whose word belongs to the list were selected, and pools were drawn that hold exactly one trial of every list word present in the split. Within a draw, the trials of each list word were put in random order and dealt out to the pools in turn: the first pool took the first trial of every word, the second pool the second trial, and so on. The number of pools per draw equals the largest trial count among the list words, and a word with fewer trials than that starts again from its first trial once its trials are used up. Every trial therefore serves as a query at least once per draw, every pool holds exactly one trial of every word, and every word contributes the same number of queries, so the accuracy is a macro-average over words. Five draws were scored and averaged. Within a pool, the prediction of every trial was compared with the projected target of every trial, with the learned temperature, and the rank of the trial's own target was read; ties count against the model. Because every candidate is a different word, no word grouping is needed and uniform chance is $k / V$ for top-$k$, with $V$ the number of list words present in the split. The empirical permutation baseline of the main text (predictions permuted within the pool, 10 permutations) was computed as well and reproduces $k / V$. The two controls replace every trial's prediction, over the full validation set and before the restriction to the list, by that of the run-mate one trial earlier (reflected at run boundaries as in Section~\ref{sec:lag}) or of a random run-mate (one draw per pool seed), and score the result on the same pools. The within-run and random-pool variants call the main text's pooled evaluation on the list trials alone (within-run: the list trials of one run per pool, within-run permutation null; random: cross-run pools of up to 512 list trials, word-grouped).

\paragraph{Results} All 330 fits were scored, and the recomputed full-set within-run gain agreed with the logged value to within 0.01\,pp in every case. Every list word was present in every validation split ($V = 50$ throughout), with 2{,}464--2{,}698 content-word trials and 21{,}150--21{,}585 frequent-word trials per split, and a median of 29 content-word and 246 frequent-word trials per run. Figure~\ref{fig:s5} and Table~\ref{tab:small-pool} give the results, Table~\ref{tab:word-lists} the lists.

Against the context-free layer-0 target, the 50-way identification of a content word from one 0.5-s window reaches $5.1 \pm 0.7$\% top-1 and $33.8 \pm 1.2$\% top-10 without a sequence model, and $4.8 \pm 0.3$\% and $34.0 \pm 1.3$\% with it (chance 2\% and 20\%), and both controls lie within 1.5 points of chance in top-1 (earlier-neighbour swap $2.7 \pm 0.2$ and $3.2 \pm 0.4$\%, random run-mate $2.9 \pm 0.1$ and $3.3 \pm 0.2$\%). In top-10 they sit further above chance, at $23.8 \pm 1.5$ and $26.1 \pm 0.9$\% for the earlier-neighbour swap and $23.5 \pm 0.4$ and $25.6 \pm 0.4$\% for the random run-mate (content list; 21.7--23.2\% for the frequent list), as Figure~\ref{fig:s5}b,\,e shows. This excess is not a property of the layer-0 targets, which are the same vector for every presentation of a word, but of the predictions. The decoder's predictions carry a run-level component even at layer 0 (Section~\ref{sec:results-window}: cosine similarity 0.09 within against 0.03 between runs), which is tilted towards the words that are frequent in the run; a trial's own word is by construction one of the words of its run, and is on average more frequent there than in a random run, so a run-mate's prediction ranks it slightly above chance. Three features of the data fit this account. The random and the earlier-neighbour control are equal within error at layer 0, so the effect is not specific to the neighbour (the neighbour's EEG window cannot contain the current word: words were shown for at least 500\,ms, report 1, and the window is 0.5\,s long). The content list is further above chance than the frequent list, because content words are passage-specific whereas function words occur in every passage. And the transformer's controls lie above the other decoder's, because its prediction for a run-mate has integrated the preceding EEG and carries more of the run state. The bias lifts a word into the top 10 of 50 more easily than to rank 1, which is why it shows more in top-10 than in top-1. The frequent list gives $5.3 \pm 0.1$\% top-1 and about 34.5\% top-10 for both decoders. At layer 0, then, either decoder identifies one of 50 words at about two and a half times chance in top-1 and 1.7 times chance in top-10, and the sequence model has nothing to add when the target carries no context.

A single layer of context changes the closed-set numbers far more than it changes the word-level metric of the main text. Without a sequence model, content-word top-1 rises from 5.1\% at layer 0 to $12.8 \pm 0.5$\% at layer 1 and stays between 13 and 16\% through the depth (peak $15.7 \pm 0.6$\% at layer 18); with the sequence model it rises to $29.6 \pm 1.9$\% at layer 1, peaks at $33.0 \pm 2.0$\% at layer 8 and declines to $24.1 \pm 2.4$\% at layer 32, with top-10 above 80\% at every contextual layer. The controls show where this comes from. The earlier-neighbour swap, which keeps the local discourse state and the position and removes the current word, rises with the accuracy itself: for the transformer it recovers $21.9 \pm 1.5$ of the 29.6\% at layer 1 and $27.9 \pm 1.3$ of the 33.0\% at layer 8; without a sequence model it recovers $7.2 \pm 0.5$ of the 12.8\% at layer 1 and $11.2 \pm 0.8$ of the 15.4\% at layer 20. The random run-mate swap, which keeps only the run-level component, stays at 6--10\% for both decoders, so what the earlier neighbour recovers and a random run-mate does not is local context, consistent with the lag profiles of Section~\ref{sec:results-window}. The word-level part, the excess of the accuracy over the earlier-neighbour control, has the depth profile of Figure~\ref{fig:word}: in top-1 it peaks at layers 1--2 ($5.8 \pm 1.1$\,pp without and $7.7 \pm 0.7$\,pp with the sequence model for the content list, $5.8 \pm 0.3$ and $7.1 \pm 0.3$\,pp for the frequent list) and declines to $2.5 \pm 0.2$ and $2.4 \pm 1.0$\,pp at the last layer. In top-10 the excess is largest at layers 0--1 without a sequence model ($10.0 \pm 0.8$ to $14.1 \pm 1.4$\,pp) and is only 2--3\,pp for the transformer from layer 8 on, whose top-10 accuracy is near its ceiling there.

The restricted within-run gain (Figure~\ref{fig:s5}c,\,f) peaks at layers 1--2 as well ($13.3 \pm 0.9$\,pp without and $26.5 \pm 1.1$\,pp with the sequence model for the content list) and declines with depth. For the transformer this variant is not position-corrected, and the list trials of a run are spread over the whole run, so the positional embedding separates them more easily than it separates the consecutive trials of the main text's pools; the transformer's values in panels c and f are therefore upper bounds on word-level decoding, and the earlier-neighbour control on the balanced pools is the cleaner control for that decoder. Per word, the best-identified content words at the word-level peak are those with characteristic local contexts, the dialogue attributions ``thought'', ``told'' and ``said'' and the forms of address ``sir'' and ``yes'' (top-1 at layer 1 without a sequence model: about 25, 24, 17, 18 and 18\%, against a median of 12\% over the 50 words), and the least identified are short frequent nouns and verbs such as ``way'', ``hand'' and ``say'' (6--7\%).

\paragraph{Reading} Three points follow. First, the numbers that are comparable to closed-vocabulary classification of attempted speech are the layer-0 ones, and they are modest: one of 50 words from a single scalp-EEG window at a few times chance. Second, closed-set accuracy against a contextual target is not word identification unless a control removes the local context; without one, the transformer's 50-way accuracy at layer 8 would be quoted at 33\% where about 5 points are the word. Third, the restriction to a candidate set is a property of the evaluation, not of the model: the same checkpoints, selected on the open-vocabulary within-run gain, produced every number here.

\section{Number of trainable parameters}
\label{sec:s-params}

Supplementary Table~\ref{tab:params} gives the number of trainable parameters of the two decoders, by component. Both use the architecture of report 1 \citep{marquardt2026decoding} with 19 channels; the counts were obtained by instantiating the model with the sweep configuration and summing the sizes of all parameter tensors that receive gradients, including the affine parameters of the normalisation layers. They are the same for every target layer, because the layer changes which hidden state of the language model serves as the target, not its dimension. The decoder without a sequence model has 3.03 million trainable parameters, the decoder with the transformer 6.55 million.

Three points help to read the table. First, the EEG encoder holds 1.92 million parameters, most of them in the fully-connected head, which maps the 1{,}536 concatenated pathway features (384 attention-pooled global features and $192 \times 6$ pooled local features) into 512 units; the convolutional blocks of the global and local pathways together hold 0.73 million. Second, the target projection, the bias-free linear map from the 4{,}096-dimensional hidden state into the 256-dimensional shared space, is trained jointly with the rest of the model and accounts for 1.05 million parameters, about a third of the total without a sequence model; it acts on the targets, not on the EEG. Third, the sequence model adds 3.52 million parameters: four pre-norm transformer encoder layers of 0.79 million each, and the learned positional embedding of $1{,}400 \times 256$ entries (0.36 million), whose contribution the position probes of Section~\ref{sec:results-probes} measure. The counts are not logged as a metric; the training script prints them at start-up, and the table was produced by instantiating the model directly.

\clearpage
\section{Supplementary figures}
\label{sec:s-figures}

\begin{figure}[htb]
  \centering
  \includegraphics[alt={Layer by epoch heat maps of the within-run gain for both decoders},width=\textwidth]{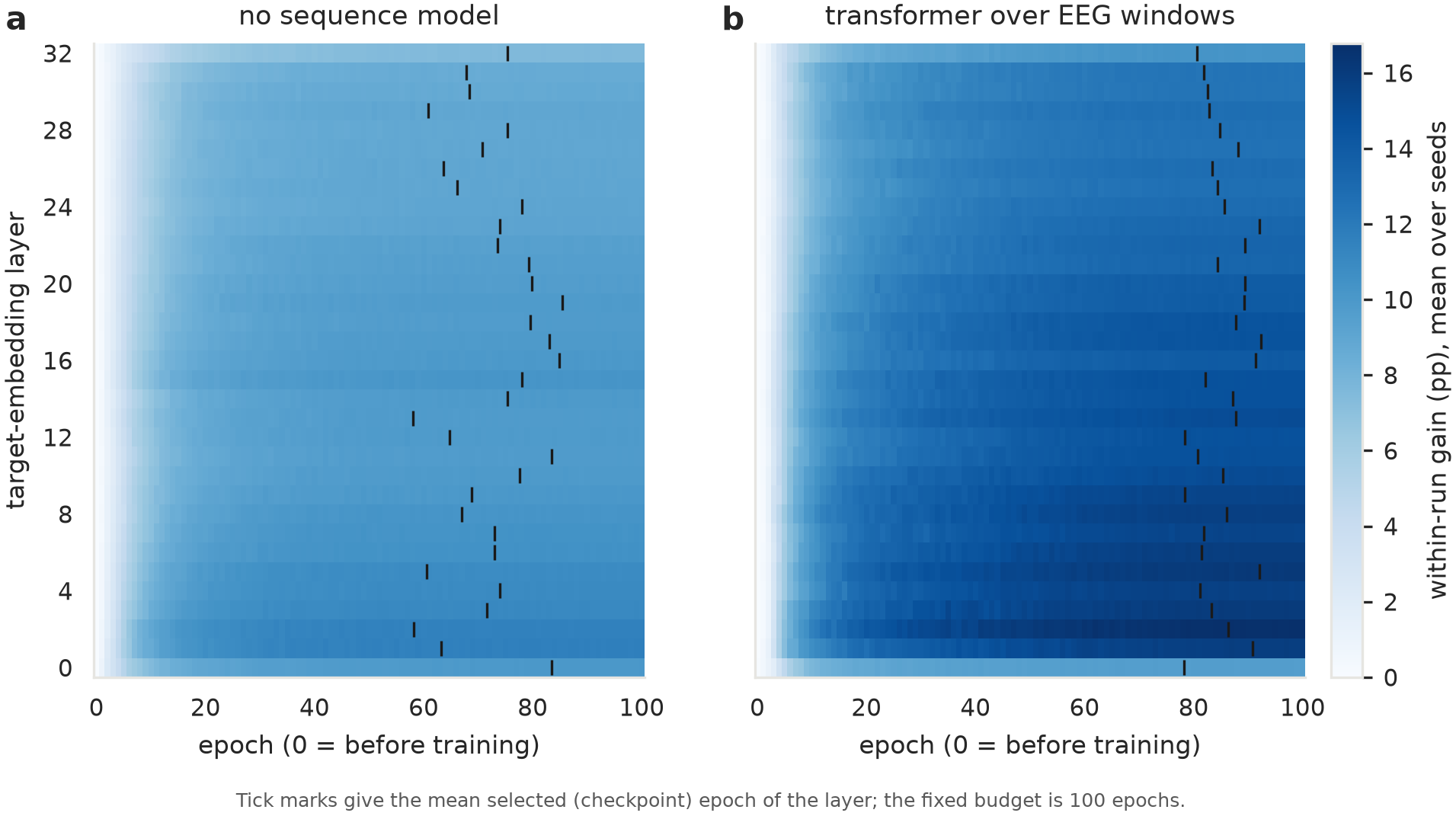}
  \caption{\textbf{Training dynamics per layer.} Within-run gain against epoch for every layer, averaged over splits, \textbf{(a)} without and \textbf{(b)} with the sequence model. Tick marks give the mean selected epoch; the budget was 100 epochs. The figure shows that every fit converged within the budget, with the selected epochs well before the ceiling at every layer.}
  \label{fig:s1}
\end{figure}

\begin{figure}[htb]
  \centering
  \includegraphics[alt={Selected epoch, validation loss, temperature and normalised accuracies against layer},width=\textwidth]{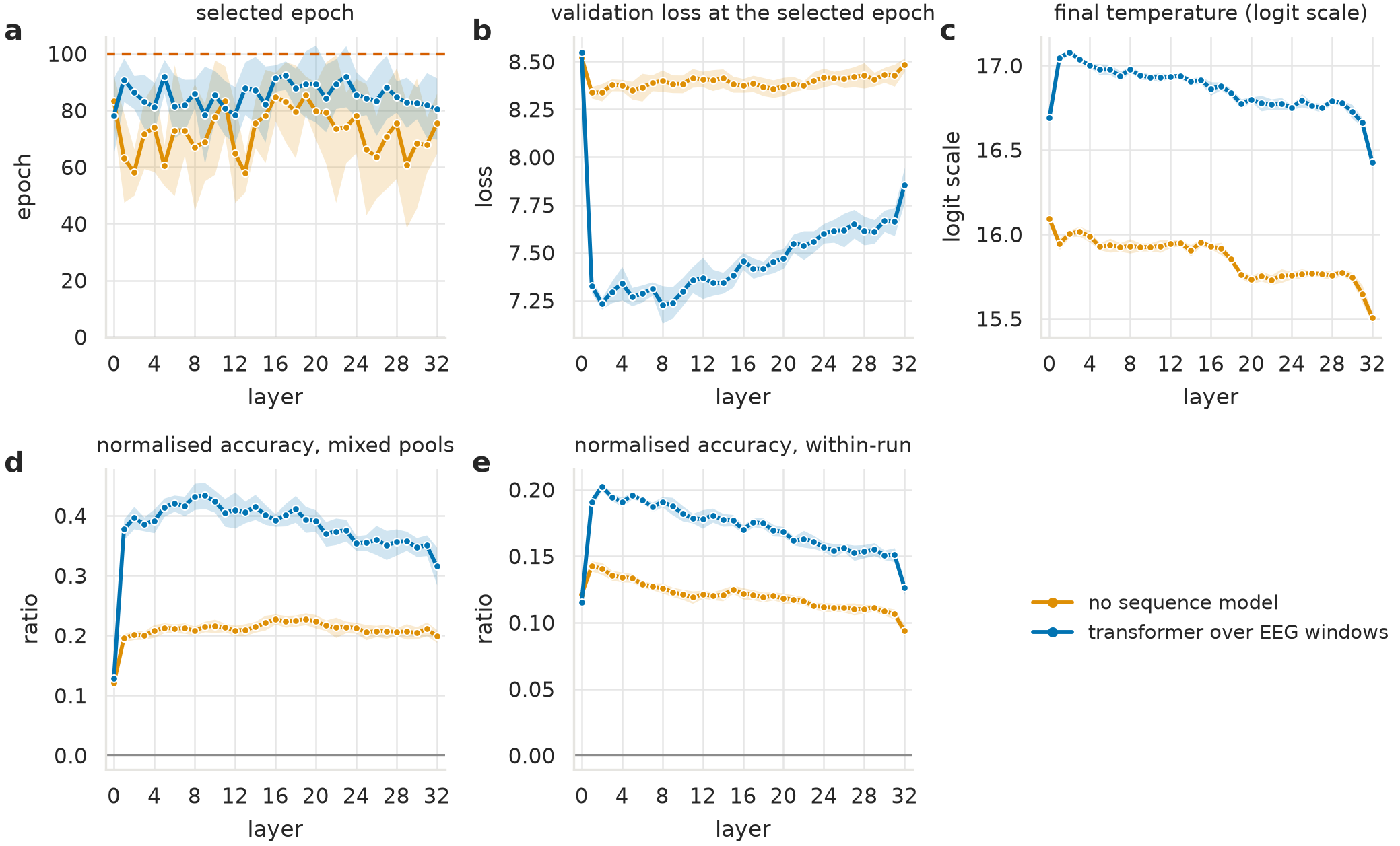}
  \caption{\textbf{Diagnostics per layer.} \textbf{(a)} Selected checkpoint epoch against the 100-epoch ceiling. \textbf{(b)} Validation loss at the selected epoch; not comparable across layers, because the contrastive loss depends on the geometry of the projected targets. \textbf{(c)} Final learned temperature (clamp 100). \textbf{(d, e)} Headroom-normalised accuracy in mixed and within-run pools. The figure shows that checkpoint selection was not truncated by the epoch ceiling and that no fit approached the temperature clamp; (d, e) restate the gains of Figures~\ref{fig:word} and~\ref{fig:context} as accuracy relative to headroom. The headroom is the maximum gain a decoder could reach. Even a perfect EEG model that correctly identifies each word on each trial would not have a gain of 1.0 because the permutation baseline is subtracted.}
  \label{fig:s2}
\end{figure}

\begin{figure}[htb]
  \centering
  \includegraphics[alt={Swap gain against lag for every layer, decoder without sequence model},width=\textwidth]{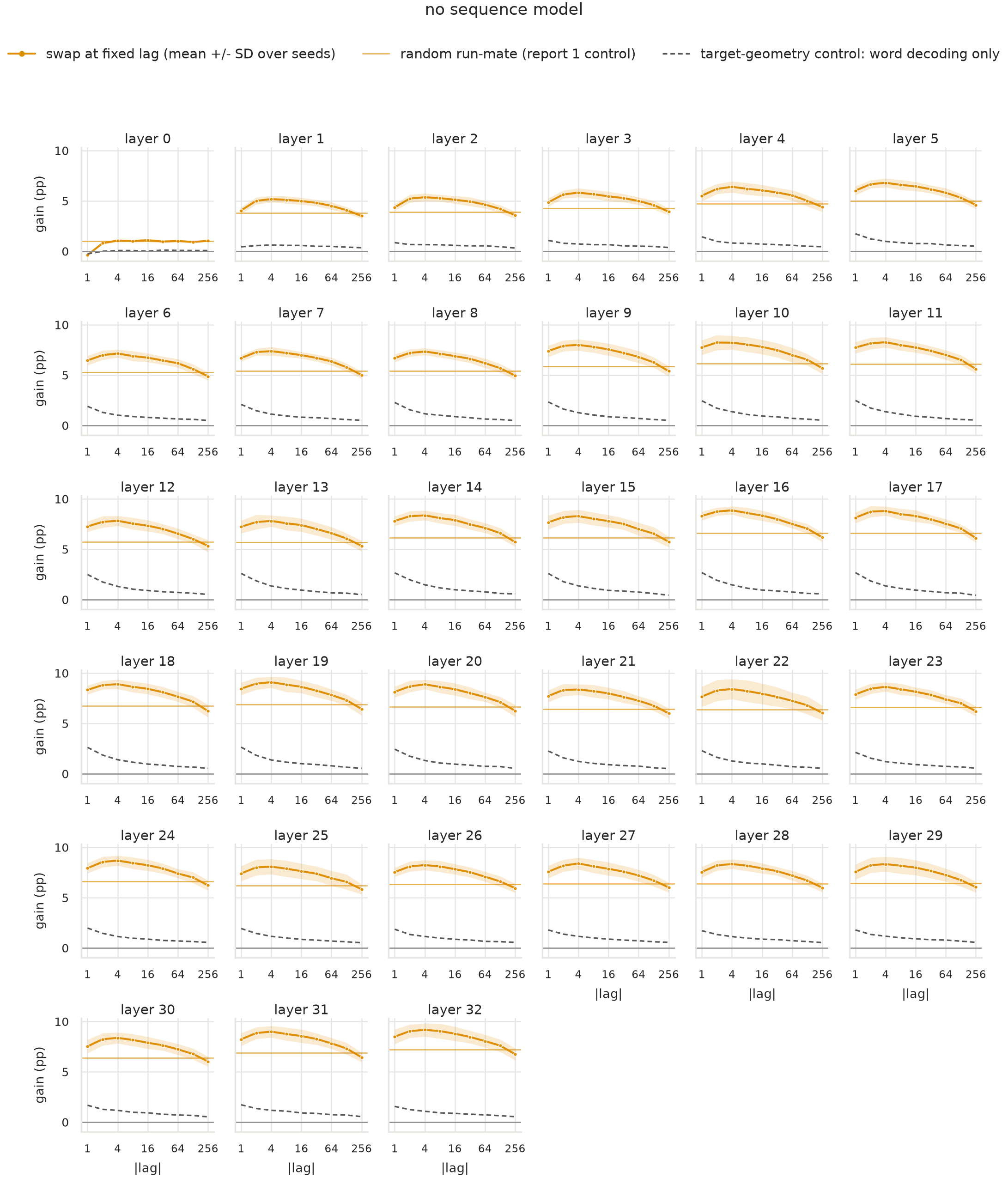}
  \caption{\textbf{Lag profiles for every layer, decoder without sequence model.} Swap gain against the lag of the swapped run-mate (mean and SD over splits), the random-run-mate control (horizontal line) and the target-geometry control (dashed). The figure shows that the comparatively flat profile without sequence model, far above the target-geometry control, holds at every layer above 0.}
  \label{fig:s3}
\end{figure}

\begin{figure}[htb]
  \centering
  \includegraphics[alt={Swap gain against lag for every layer, transformer decoder},width=\textwidth]{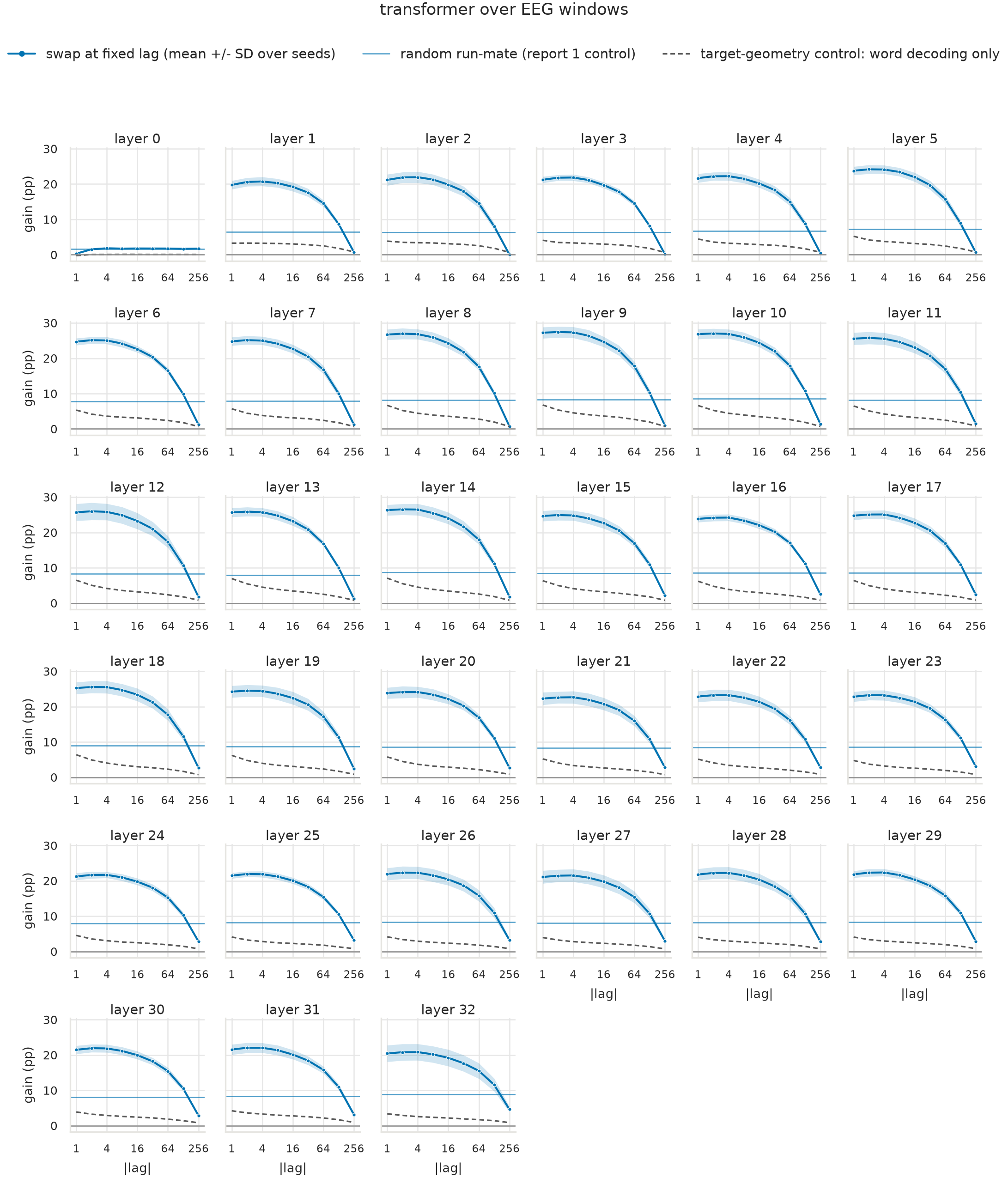}
  \caption{\textbf{Lag profiles for every layer, transformer decoder.} As Supplementary Figure~\ref{fig:s3}, for the decoder with the sequence model. The steep decay beyond a lag of a few dozen trials holds at every layer, so the sequence model's context is local throughout the depth.}
  \label{fig:s4}
\end{figure}

\clearpage
\section{Supplementary tables}
\label{sec:s-tables}

Tables~\ref{tab:peaks}--\ref{tab:session} summarise the results quoted in the main text: the peaks of the depth profiles, the lag-dependent swap at selected layers, and the session analysis pooled over layers. Tables~\ref{tab:layer-gains}--\ref{tab:diagnostics} give the complete per-layer values. Tables~\ref{tab:small-pool} and~\ref{tab:word-lists} belong to the restricted-candidate-set analysis of Section~\ref{sec:s-small-pool}. Table~\ref{tab:params} gives the number of trainable parameters of the two decoders by component (Section~\ref{sec:s-params}). Throughout the tables, noTF denotes the decoder without a sequence model and TF the decoder with the transformer over the sequence of EEG windows.

\input{tables/table_S1}
\input{tables/table_S2}
\input{tables/table_S3}
\input{tables/table_S4}
\input{tables/table_S5}
\input{tables/table_S6}
\input{tables/table_S7}
\input{tables/table_S8}
\input{tables/table_S9}
\input{tables/table_S10}
\input{tables/table_S11}
\input{tables/table_S12}

\end{document}

%% file: tables/table_S1.tex
\begin{table}[htb]
  \centering
  \caption{Where each metric peaks along the depth axis. Peak layer and value of the seed-mean curve (mean $\pm$ SD over the seeds at the peak layer), each seed's own peak layer, the mean pairwise Spearman correlation between the per-seed curves (agreement on the shape of the depth profile), the values at layers 0, 20 and 32, and the Spearman correlation between layer index and value (a monotonic-trend descriptor). Gains in percentage points (pp).}
  \label{tab:peaks}
  \footnotesize
  \setlength{\tabcolsep}{4pt}
  \resizebox{\textwidth}{!}{%
  \begin{tabular}{llccccrrrc}
    \toprule
    Metric & Decoder & Peak & Value at peak & Peak per seed & Seed agreement & L0 & L20 & L32 & $\rho$(layer) \\
    \midrule
    within-run gain & noTF & L1 & 11.8 $\pm$ 0.3 & 1/1/2/1/1 & 0.95 & 10.1 & 9.7 & 7.6 & -0.94 \\
     & TF & L2 & 16.8 $\pm$ 0.2 & 2/2/2/2/2 & 0.97 & 9.7 & 14.0 & 10.4 & -0.81 \\
    position-corrected within-run gain & noTF & L1 & 11.8 $\pm$ 0.3 & 1/1/2/1/1 & 0.95 & 10.1 & 9.7 & 7.6 & -0.94 \\
     & TF & L2 & 14.0 $\pm$ 0.7 & 2/1/2/5/2 & 0.88 & 9.6 & 11.6 & 8.1 & -0.79 \\
    overall gain & noTF & L19 & 20.0 $\pm$ 0.9 & 19/16/16/19/16 & 0.78 & 10.3 & 19.6 & 17.3 & 0.06 \\
     & TF & L9 & 38.5 $\pm$ 1.9 & 9/9/8/5/6 & 0.85 & 11.0 & 34.7 & 27.9 & -0.59 \\
    context-tracking gain & noTF & L32 & 7.2 $\pm$ 0.6 & 32/32/32/32/32 & 0.91 & 1.0 & 6.6 & 7.2 & 0.84 \\
     & TF & L18 & 9.0 $\pm$ 0.9 & 14/32/18/20/19 & 0.54 & 1.6 & 8.6 & 8.9 & 0.52 \\
    context-independent gain & noTF & L5 & 13.6 $\pm$ 0.5 & 5/4/5/2/2 & 0.90 & 9.2 & 12.9 & 10.1 & -0.69 \\
     & TF & L8 & 30.2 $\pm$ 1.4 & 9/9/8/5/6 & 0.92 & 9.4 & 26.1 & 19.0 & -0.71 \\
    within-run gain, rare bin & noTF & L15 & 6.6 $\pm$ 0.4 & 6/16/9/9/12 & 0.70 & 3.6 & 6.1 & 3.4 & -0.58 \\
     & TF & L8 & 12.6 $\pm$ 0.6 & 9/7/5/6/7 & 0.78 & 3.3 & 10.3 & 6.9 & -0.47 \\
    within-run gain, mid bin & noTF & L8 & 7.8 $\pm$ 0.3 & 5/8/14/8/6 & 0.67 & 4.0 & 6.8 & 4.7 & -0.43 \\
     & TF & L8 & 14.9 $\pm$ 1.0 & 8/8/8/7/8 & 0.83 & 3.5 & 12.8 & 8.9 & -0.46 \\
    within-run gain, frequent bin & noTF & L1 & 12.5 $\pm$ 0.3 & 1/1/1/1/1 & 0.96 & 10.8 & 10.1 & 8.1 & -0.96 \\
     & TF & L2 & 17.5 $\pm$ 0.2 & 2/2/2/2/2 & 0.97 & 10.4 & 14.3 & 10.7 & -0.82 \\
    rare / frequent ratio & noTF & L9 & 0.62 $\pm$ 0.05 & 19/16/11/9/12 & 0.53 & 0.33 & 0.60 & 0.43 & -0.13 \\
     & TF & L21 & 0.80 $\pm$ 0.07 & 21/21/17/13/14 & 0.43 & 0.32 & 0.72 & 0.65 & 0.11 \\
    \bottomrule
  \end{tabular}}
\end{table}

%% file: tables/table_S2.tex
\begin{table}[htb]
  \centering
  \caption{Lag-dependent run-mate swap at selected layers. Every column is a cross-run gain in the random (mixed) pools of the overall gain, mean $\pm$ SD over seeds, in pp; noTF and TF denote the decoder without and with the sequence model. \emph{Overall}: the gain of the decoder's own predictions, with no swap. \emph{Random run-mate}: the gain after replacing each trial's prediction with that of a random trial of the same run (report 1's context-tracking control). \emph{Lag 4} and \emph{Lag 256}: the gain after replacing each trial's prediction with that of the run-mate 4 or 256 trials away (earlier and later partner averaged). \emph{Geometry control, random} and \emph{Geometry control, lag 4}: the random and the lag-4 swap applied to the target-geometry control, a synthetic predictor made of the true target plus noise calibrated to the decoder's within-run gain, which has word-level information but none from the EEG; its swap gain is what the geometry of the targets alone produces, and the decoder's excess over it requires context in the EEG.}
  \label{tab:lag}
  \footnotesize
  \setlength{\tabcolsep}{4pt}
  \resizebox{\textwidth}{!}{%
  \begin{tabular}{llcccccc}
    \toprule
    Decoder & Layer & Overall & Random run-mate & Geometry control, random & Lag 4 & Lag 256 & Geometry control, lag 4 \\
    \midrule
    noTF & 0 & 10.3 $\pm$ 0.3 & 1.0 $\pm$ 0.1 & 0.1 $\pm$ 0.1 & 1.1 $\pm$ 0.1 & 1.0 $\pm$ 0.1 & 0.1 $\pm$ 0.0 \\
     & 1 & 17.0 $\pm$ 0.6 & 3.8 $\pm$ 0.3 & 0.4 $\pm$ 0.0 & 5.2 $\pm$ 0.3 & 3.5 $\pm$ 0.3 & 0.6 $\pm$ 0.1 \\
     & 2 & 17.5 $\pm$ 0.7 & 3.9 $\pm$ 0.4 & 0.4 $\pm$ 0.1 & 5.4 $\pm$ 0.4 & 3.6 $\pm$ 0.4 & 0.7 $\pm$ 0.1 \\
     & 4 & 18.1 $\pm$ 1.0 & 4.7 $\pm$ 0.5 & 0.5 $\pm$ 0.1 & 6.4 $\pm$ 0.6 & 4.4 $\pm$ 0.5 & 0.8 $\pm$ 0.1 \\
     & 8 & 18.2 $\pm$ 0.6 & 5.4 $\pm$ 0.4 & 0.6 $\pm$ 0.1 & 7.3 $\pm$ 0.3 & 4.9 $\pm$ 0.4 & 1.2 $\pm$ 0.1 \\
     & 12 & 18.2 $\pm$ 0.8 & 5.7 $\pm$ 0.6 & 0.7 $\pm$ 0.1 & 7.8 $\pm$ 0.5 & 5.3 $\pm$ 0.6 & 1.3 $\pm$ 0.0 \\
     & 16 & 20.0 $\pm$ 0.7 & 6.6 $\pm$ 0.5 & 0.6 $\pm$ 0.1 & 8.9 $\pm$ 0.4 & 6.2 $\pm$ 0.5 & 1.5 $\pm$ 0.1 \\
     & 20 & 19.6 $\pm$ 1.0 & 6.6 $\pm$ 0.5 & 0.7 $\pm$ 0.1 & 8.9 $\pm$ 0.5 & 6.3 $\pm$ 0.6 & 1.3 $\pm$ 0.1 \\
     & 24 & 18.6 $\pm$ 0.8 & 6.6 $\pm$ 0.5 & 0.6 $\pm$ 0.1 & 8.7 $\pm$ 0.5 & 6.2 $\pm$ 0.5 & 1.1 $\pm$ 0.1 \\
     & 28 & 18.0 $\pm$ 0.6 & 6.4 $\pm$ 0.5 & 0.6 $\pm$ 0.1 & 8.3 $\pm$ 0.5 & 6.0 $\pm$ 0.4 & 1.1 $\pm$ 0.1 \\
     & 32 & 17.3 $\pm$ 1.0 & 7.2 $\pm$ 0.6 & 0.6 $\pm$ 0.0 & 9.2 $\pm$ 0.7 & 6.8 $\pm$ 0.6 & 1.1 $\pm$ 0.1 \\
    TF & 0 & 11.0 $\pm$ 0.1 & 1.6 $\pm$ 0.0 & 0.1 $\pm$ 0.0 & 1.8 $\pm$ 0.1 & 1.7 $\pm$ 0.0 & 0.1 $\pm$ 0.1 \\
     & 1 & 33.4 $\pm$ 1.5 & 6.5 $\pm$ 0.6 & 1.4 $\pm$ 0.0 & 20.7 $\pm$ 1.3 & 0.7 $\pm$ 0.3 & 3.3 $\pm$ 0.1 \\
     & 2 & 35.0 $\pm$ 1.7 & 6.3 $\pm$ 0.7 & 1.3 $\pm$ 0.1 & 22.0 $\pm$ 1.6 & 0.1 $\pm$ 0.3 & 3.4 $\pm$ 0.1 \\
     & 4 & 34.6 $\pm$ 1.8 & 6.7 $\pm$ 0.7 & 1.3 $\pm$ 0.1 & 22.2 $\pm$ 1.2 & 0.5 $\pm$ 0.3 & 3.3 $\pm$ 0.0 \\
     & 8 & 38.3 $\pm$ 1.9 & 8.1 $\pm$ 0.7 & 1.6 $\pm$ 0.1 & 26.8 $\pm$ 1.4 & 0.8 $\pm$ 0.3 & 4.4 $\pm$ 0.2 \\
     & 12 & 36.3 $\pm$ 2.7 & 8.3 $\pm$ 1.0 & 1.5 $\pm$ 0.1 & 25.8 $\pm$ 2.3 & 1.8 $\pm$ 0.3 & 4.1 $\pm$ 0.3 \\
     & 16 & 34.8 $\pm$ 1.0 & 8.5 $\pm$ 0.5 & 1.4 $\pm$ 0.1 & 24.2 $\pm$ 1.0 & 2.5 $\pm$ 0.1 & 3.9 $\pm$ 0.2 \\
     & 20 & 34.7 $\pm$ 1.6 & 8.6 $\pm$ 0.7 & 1.4 $\pm$ 0.1 & 24.1 $\pm$ 1.5 & 2.7 $\pm$ 0.4 & 3.7 $\pm$ 0.2 \\
     & 24 & 31.3 $\pm$ 1.1 & 7.9 $\pm$ 0.5 & 1.2 $\pm$ 0.1 & 21.7 $\pm$ 0.9 & 2.8 $\pm$ 0.3 & 3.0 $\pm$ 0.1 \\
     & 28 & 31.4 $\pm$ 2.0 & 8.1 $\pm$ 0.9 & 1.2 $\pm$ 0.1 & 22.2 $\pm$ 1.7 & 2.8 $\pm$ 0.4 & 3.0 $\pm$ 0.2 \\
     & 32 & 27.9 $\pm$ 2.8 & 8.9 $\pm$ 1.4 & 1.2 $\pm$ 0.1 & 20.8 $\pm$ 2.3 & 4.6 $\pm$ 0.8 & 2.6 $\pm$ 0.1 \\
    \bottomrule
  \end{tabular}}
\end{table}

%% file: tables/table_S3.tex
\begin{table}[htb]
  \centering
  \caption{Narrative neighbours across session boundaries. Top: fraction of a validation run's cross-run gain that is retained when its trials take the predictions of the validation run 2--8 places away in the narrative, with the partner's trials matched by position or scrambled, for neighbours recorded in the same or in a different session. Bottom: the regression of the run-centroid prediction similarity on target similarity, the same-session indicator and the log recording-time gap, over all pairs of validation runs: the standardised coefficients; the same-session term on its own 0/1 scale and as a direct contrast between same-session pairs and same-day pairs from different sessions after regressing on target similarity alone; the time-gap term with a same-passage dummy and the log narrative distance added, and within passages with the narrative distance alongside; and a block-permutation null for the same-session term (session sizes kept, their order shuffled over the time-ordered runs, 1{,}000 draws per checkpoint). Both parts: mean $\pm$ SD over layers 1--32 of the seed-mean values.}
  \label{tab:session}
  \footnotesize
  \setlength{\tabcolsep}{4pt}
  \begin{tabular}{llcc}
    \toprule
    Decoder & Partner trials & Same session & Different session \\
    \midrule
    noTF & matched & 0.32 $\pm$ 0.03 & 0.24 $\pm$ 0.02 \\
     & scrambled & 0.25 $\pm$ 0.03 & 0.19 $\pm$ 0.02 \\
    TF & matched & 0.47 $\pm$ 0.04 & 0.36 $\pm$ 0.02 \\
     & scrambled & 0.15 $\pm$ 0.03 & 0.10 $\pm$ 0.01 \\
    \bottomrule
  \end{tabular}
  \par\medskip
  \begin{tabular}{p{8.6cm}cc}
    \toprule
     & noTF & TF \\
    \midrule
    Target similarity (standardised) & 0.32 $\pm$ 0.01 & 0.44 $\pm$ 0.02 \\
    Same session (standardised) & 0.05 $\pm$ 0.02 & -0.02 $\pm$ 0.01 \\
    Log time gap (standardised) & -0.33 $\pm$ 0.04 & -0.31 $\pm$ 0.02 \\
    Same session, indicator on its 0/1 scale (SD of the response) & 0.31 $\pm$ 0.15 & -0.11 $\pm$ 0.06 \\
    Same-session minus same-day different-session pairs, text held fixed (SD of the response) & 0.60 $\pm$ 0.10 & 0.19 $\pm$ 0.08 \\
    Log time gap, same-passage dummy and narrative distance added (standardised) & -0.25 $\pm$ 0.05 & -0.28 $\pm$ 0.03 \\
    Log time gap, within-passage pairs, narrative distance alongside (standardised) & -0.10 $\pm$ 0.04 & -0.12 $\pm$ 0.05 \\
    Permutation null of the same-session term: median $p$ (share of checkpoints with $p < 0.05$) & 0.04 (0.52) & 0.58 (0.06) \\
    \bottomrule
  \end{tabular}
\end{table}

%% file: tables/table_S4.tex
\begin{table}[htb]
  \centering
  \caption{Per-layer retrieval-gain decomposition, mean $\pm$ SD over seeds (pp).}
  \label{tab:layer-gains}
  \footnotesize
  \setlength{\tabcolsep}{4pt}
  \resizebox{\textwidth}{!}{%
  \begin{tabular}{lcccccccccc}
    \toprule
    Layer & \makecell[c]{within-run\\noTF} & \makecell[c]{within-run\\TF} & \makecell[c]{position-corr.\\noTF} & \makecell[c]{position-corr.\\TF} & \makecell[c]{overall\\noTF} & \makecell[c]{overall\\TF} & \makecell[c]{context-tracking\\noTF} & \makecell[c]{context-tracking\\TF} & \makecell[c]{context-indep.\\noTF} & \makecell[c]{context-indep.\\TF} \\
    \midrule
    0 & 10.1 $\pm$ 0.3 & 9.7 $\pm$ 0.1 & 10.1 $\pm$ 0.3 & 9.6 $\pm$ 0.2 & 10.3 $\pm$ 0.3 & 11.0 $\pm$ 0.1 & 1.0 $\pm$ 0.1 & 1.6 $\pm$ 0.0 & 9.2 $\pm$ 0.3 & 9.4 $\pm$ 0.1 \\
    1 & 11.8 $\pm$ 0.3 & 15.9 $\pm$ 0.3 & 11.8 $\pm$ 0.3 & 13.4 $\pm$ 0.3 & 17.0 $\pm$ 0.6 & 33.4 $\pm$ 1.5 & 3.8 $\pm$ 0.3 & 6.5 $\pm$ 0.6 & 13.2 $\pm$ 0.4 & 26.9 $\pm$ 1.0 \\
    2 & 11.6 $\pm$ 0.3 & 16.8 $\pm$ 0.2 & 11.6 $\pm$ 0.3 & 14.0 $\pm$ 0.7 & 17.5 $\pm$ 0.7 & 35.0 $\pm$ 1.7 & 3.9 $\pm$ 0.4 & 6.3 $\pm$ 0.7 & 13.6 $\pm$ 0.4 & 28.7 $\pm$ 1.0 \\
    3 & 11.2 $\pm$ 0.3 & 16.2 $\pm$ 0.2 & 11.2 $\pm$ 0.3 & 12.7 $\pm$ 0.6 & 17.4 $\pm$ 0.6 & 34.1 $\pm$ 1.0 & 4.3 $\pm$ 0.4 & 6.3 $\pm$ 0.3 & 13.2 $\pm$ 0.3 & 27.8 $\pm$ 0.7 \\
    4 & 11.1 $\pm$ 0.3 & 15.8 $\pm$ 0.3 & 11.1 $\pm$ 0.3 & 12.7 $\pm$ 0.8 & 18.1 $\pm$ 1.0 & 34.6 $\pm$ 1.8 & 4.7 $\pm$ 0.5 & 6.7 $\pm$ 0.7 & 13.4 $\pm$ 0.6 & 27.9 $\pm$ 1.2 \\
    5 & 11.0 $\pm$ 0.3 & 16.3 $\pm$ 0.2 & 11.0 $\pm$ 0.3 & 13.3 $\pm$ 0.5 & 18.6 $\pm$ 0.7 & 36.7 $\pm$ 1.4 & 5.0 $\pm$ 0.4 & 7.2 $\pm$ 0.7 & 13.6 $\pm$ 0.5 & 29.5 $\pm$ 0.8 \\
    6 & 10.6 $\pm$ 0.2 & 16.0 $\pm$ 0.1 & 10.6 $\pm$ 0.2 & 13.0 $\pm$ 0.8 & 18.5 $\pm$ 0.7 & 37.3 $\pm$ 1.2 & 5.3 $\pm$ 0.4 & 7.8 $\pm$ 0.5 & 13.2 $\pm$ 0.5 & 29.5 $\pm$ 0.8 \\
    7 & 10.5 $\pm$ 0.2 & 15.6 $\pm$ 0.3 & 10.5 $\pm$ 0.2 & 12.5 $\pm$ 0.6 & 18.6 $\pm$ 0.8 & 36.9 $\pm$ 1.5 & 5.4 $\pm$ 0.4 & 7.9 $\pm$ 0.7 & 13.1 $\pm$ 0.6 & 29.0 $\pm$ 1.0 \\
    8 & 10.4 $\pm$ 0.3 & 15.9 $\pm$ 0.3 & 10.4 $\pm$ 0.3 & 12.6 $\pm$ 0.3 & 18.2 $\pm$ 0.7 & 38.3 $\pm$ 1.9 & 5.4 $\pm$ 0.4 & 8.1 $\pm$ 0.7 & 12.8 $\pm$ 0.4 & 30.2 $\pm$ 1.4 \\
    9 & 10.1 $\pm$ 0.3 & 15.6 $\pm$ 0.6 & 10.1 $\pm$ 0.3 & 12.1 $\pm$ 1.2 & 18.8 $\pm$ 0.8 & 38.5 $\pm$ 1.9 & 5.9 $\pm$ 0.5 & 8.3 $\pm$ 0.8 & 12.9 $\pm$ 0.4 & 30.2 $\pm$ 1.3 \\
    10 & 10.0 $\pm$ 0.3 & 15.2 $\pm$ 0.4 & 10.0 $\pm$ 0.3 & 11.6 $\pm$ 0.7 & 18.9 $\pm$ 1.0 & 37.7 $\pm$ 1.6 & 6.1 $\pm$ 0.7 & 8.5 $\pm$ 0.5 & 12.8 $\pm$ 0.6 & 29.1 $\pm$ 1.2 \\
    11 & 9.8 $\pm$ 0.4 & 14.8 $\pm$ 0.3 & 9.8 $\pm$ 0.4 & 11.4 $\pm$ 0.3 & 18.7 $\pm$ 0.6 & 36.0 $\pm$ 2.0 & 6.1 $\pm$ 0.5 & 8.2 $\pm$ 0.7 & 12.6 $\pm$ 0.6 & 27.8 $\pm$ 1.4 \\
    12 & 10.0 $\pm$ 0.3 & 14.8 $\pm$ 0.6 & 10.0 $\pm$ 0.3 & 11.4 $\pm$ 0.6 & 18.2 $\pm$ 0.8 & 36.3 $\pm$ 2.7 & 5.7 $\pm$ 0.6 & 8.4 $\pm$ 1.0 & 12.5 $\pm$ 0.4 & 28.0 $\pm$ 1.8 \\
    13 & 9.9 $\pm$ 0.2 & 15.1 $\pm$ 0.4 & 9.9 $\pm$ 0.2 & 11.5 $\pm$ 0.3 & 18.3 $\pm$ 0.9 & 36.1 $\pm$ 1.6 & 5.7 $\pm$ 0.6 & 7.9 $\pm$ 0.5 & 12.6 $\pm$ 0.5 & 28.1 $\pm$ 1.1 \\
    14 & 10.0 $\pm$ 0.4 & 14.7 $\pm$ 0.4 & 10.0 $\pm$ 0.4 & 11.6 $\pm$ 0.4 & 18.8 $\pm$ 0.8 & 36.8 $\pm$ 1.9 & 6.1 $\pm$ 0.5 & 8.7 $\pm$ 0.7 & 12.7 $\pm$ 0.5 & 28.1 $\pm$ 1.2 \\
    15 & 10.3 $\pm$ 0.2 & 14.8 $\pm$ 0.4 & 10.3 $\pm$ 0.2 & 11.5 $\pm$ 0.4 & 19.4 $\pm$ 1.1 & 35.7 $\pm$ 1.5 & 6.2 $\pm$ 0.6 & 8.4 $\pm$ 0.7 & 13.2 $\pm$ 0.6 & 27.3 $\pm$ 0.8 \\
    16 & 10.1 $\pm$ 0.3 & 14.2 $\pm$ 0.3 & 10.1 $\pm$ 0.3 & 10.9 $\pm$ 0.1 & 20.0 $\pm$ 0.7 & 34.8 $\pm$ 1.0 & 6.6 $\pm$ 0.5 & 8.5 $\pm$ 0.5 & 13.3 $\pm$ 0.5 & 26.3 $\pm$ 0.6 \\
    17 & 10.0 $\pm$ 0.3 & 14.6 $\pm$ 0.3 & 10.0 $\pm$ 0.3 & 11.6 $\pm$ 0.3 & 19.6 $\pm$ 0.8 & 35.6 $\pm$ 1.6 & 6.6 $\pm$ 0.5 & 8.6 $\pm$ 0.6 & 13.0 $\pm$ 0.5 & 27.0 $\pm$ 1.1 \\
    18 & 9.8 $\pm$ 0.3 & 14.6 $\pm$ 0.3 & 9.8 $\pm$ 0.3 & 11.7 $\pm$ 0.2 & 19.7 $\pm$ 0.8 & 36.5 $\pm$ 2.0 & 6.8 $\pm$ 0.5 & 9.0 $\pm$ 0.9 & 12.9 $\pm$ 0.5 & 27.5 $\pm$ 1.1 \\
    19 & 9.9 $\pm$ 0.3 & 14.1 $\pm$ 0.3 & 9.9 $\pm$ 0.3 & 10.9 $\pm$ 0.4 & 20.0 $\pm$ 0.9 & 34.9 $\pm$ 2.0 & 6.9 $\pm$ 0.6 & 8.7 $\pm$ 0.8 & 13.1 $\pm$ 0.5 & 26.3 $\pm$ 1.2 \\
    20 & 9.7 $\pm$ 0.3 & 14.0 $\pm$ 0.3 & 9.7 $\pm$ 0.3 & 11.6 $\pm$ 0.4 & 19.6 $\pm$ 1.0 & 34.7 $\pm$ 1.6 & 6.6 $\pm$ 0.5 & 8.6 $\pm$ 0.7 & 12.9 $\pm$ 0.7 & 26.1 $\pm$ 0.9 \\
    21 & 9.7 $\pm$ 0.3 & 13.4 $\pm$ 0.3 & 9.7 $\pm$ 0.3 & 10.6 $\pm$ 0.3 & 19.0 $\pm$ 0.8 & 32.7 $\pm$ 1.8 & 6.4 $\pm$ 0.6 & 8.3 $\pm$ 0.9 & 12.5 $\pm$ 0.4 & 24.4 $\pm$ 1.0 \\
    22 & 9.6 $\pm$ 0.3 & 13.5 $\pm$ 0.5 & 9.6 $\pm$ 0.3 & 10.9 $\pm$ 0.5 & 18.7 $\pm$ 1.5 & 33.1 $\pm$ 2.0 & 6.4 $\pm$ 0.8 & 8.4 $\pm$ 0.8 & 12.3 $\pm$ 0.8 & 24.7 $\pm$ 1.4 \\
    23 & 9.2 $\pm$ 0.2 & 13.3 $\pm$ 0.4 & 9.2 $\pm$ 0.2 & 10.4 $\pm$ 0.4 & 18.7 $\pm$ 0.9 & 33.2 $\pm$ 1.6 & 6.6 $\pm$ 0.5 & 8.6 $\pm$ 0.7 & 12.1 $\pm$ 0.7 & 24.6 $\pm$ 0.9 \\
    24 & 9.2 $\pm$ 0.3 & 13.0 $\pm$ 0.3 & 9.2 $\pm$ 0.3 & 10.1 $\pm$ 0.3 & 18.6 $\pm$ 0.8 & 31.3 $\pm$ 1.1 & 6.6 $\pm$ 0.5 & 8.0 $\pm$ 0.5 & 12.0 $\pm$ 0.5 & 23.4 $\pm$ 0.6 \\
    25 & 9.1 $\pm$ 0.3 & 12.8 $\pm$ 0.4 & 9.1 $\pm$ 0.3 & 10.4 $\pm$ 0.4 & 17.9 $\pm$ 1.2 & 31.4 $\pm$ 1.2 & 6.2 $\pm$ 0.6 & 8.2 $\pm$ 0.4 & 11.7 $\pm$ 0.7 & 23.2 $\pm$ 0.9 \\
    26 & 9.1 $\pm$ 0.3 & 12.9 $\pm$ 0.2 & 9.1 $\pm$ 0.3 & 10.6 $\pm$ 0.3 & 18.1 $\pm$ 0.8 & 31.8 $\pm$ 2.2 & 6.3 $\pm$ 0.4 & 8.4 $\pm$ 1.0 & 11.7 $\pm$ 0.6 & 23.4 $\pm$ 1.2 \\
    27 & 9.0 $\pm$ 0.4 & 12.6 $\pm$ 0.5 & 9.0 $\pm$ 0.4 & 10.2 $\pm$ 0.5 & 18.1 $\pm$ 1.1 & 31.0 $\pm$ 2.2 & 6.4 $\pm$ 0.5 & 8.1 $\pm$ 0.9 & 11.7 $\pm$ 0.7 & 22.9 $\pm$ 1.3 \\
    28 & 9.0 $\pm$ 0.3 & 12.7 $\pm$ 0.4 & 9.0 $\pm$ 0.3 & 10.3 $\pm$ 0.6 & 18.0 $\pm$ 0.6 & 31.4 $\pm$ 2.0 & 6.4 $\pm$ 0.5 & 8.1 $\pm$ 0.9 & 11.6 $\pm$ 0.5 & 23.3 $\pm$ 1.1 \\
    29 & 9.1 $\pm$ 0.2 & 12.9 $\pm$ 0.4 & 9.1 $\pm$ 0.2 & 10.2 $\pm$ 0.5 & 18.0 $\pm$ 1.2 & 31.7 $\pm$ 1.4 & 6.4 $\pm$ 0.6 & 8.3 $\pm$ 0.5 & 11.6 $\pm$ 0.7 & 23.4 $\pm$ 0.9 \\
    30 & 8.9 $\pm$ 0.3 & 12.5 $\pm$ 0.3 & 8.9 $\pm$ 0.3 & 10.1 $\pm$ 0.1 & 17.8 $\pm$ 0.9 & 30.7 $\pm$ 1.4 & 6.4 $\pm$ 0.6 & 8.0 $\pm$ 0.6 & 11.4 $\pm$ 0.6 & 22.7 $\pm$ 0.8 \\
    31 & 8.7 $\pm$ 0.3 & 12.5 $\pm$ 0.4 & 8.7 $\pm$ 0.3 & 9.6 $\pm$ 0.5 & 18.4 $\pm$ 1.0 & 31.0 $\pm$ 1.6 & 6.9 $\pm$ 0.5 & 8.3 $\pm$ 0.7 & 11.5 $\pm$ 0.8 & 22.7 $\pm$ 1.1 \\
    32 & 7.6 $\pm$ 0.4 & 10.4 $\pm$ 0.4 & 7.6 $\pm$ 0.4 & 8.1 $\pm$ 0.6 & 17.3 $\pm$ 1.0 & 27.9 $\pm$ 2.8 & 7.2 $\pm$ 0.6 & 8.9 $\pm$ 1.4 & 10.1 $\pm$ 0.5 & 19.0 $\pm$ 1.5 \\
    \bottomrule
  \end{tabular}}
\end{table}

%% file: tables/table_S5.tex
\begin{table}[htb]
  \centering
  \caption{Per-layer within-run gain by word-frequency bin (pp) and the rare-to-frequent ratio, mean $\pm$ SD over seeds.}
  \label{tab:layer-bins}
  \footnotesize
  \setlength{\tabcolsep}{4pt}
  \resizebox{\textwidth}{!}{%
  \begin{tabular}{lcccccccc}
    \toprule
    Layer & \makecell[c]{rare\\noTF} & \makecell[c]{rare\\TF} & \makecell[c]{mid\\noTF} & \makecell[c]{mid\\TF} & \makecell[c]{frequent\\noTF} & \makecell[c]{frequent\\TF} & \makecell[c]{rare / frequent\\noTF} & \makecell[c]{rare / frequent\\TF} \\
    \midrule
    0 & 3.56 $\pm$ 0.47 & 3.29 $\pm$ 0.50 & 4.00 $\pm$ 0.17 & 3.48 $\pm$ 0.24 & 10.84 $\pm$ 0.37 & 10.44 $\pm$ 0.15 & 0.33 $\pm$ 0.04 & 0.32 $\pm$ 0.05 \\
    1 & 5.19 $\pm$ 0.49 & 8.44 $\pm$ 0.70 & 6.27 $\pm$ 0.57 & 9.99 $\pm$ 0.57 & 12.53 $\pm$ 0.34 & 16.75 $\pm$ 0.36 & 0.41 $\pm$ 0.03 & 0.50 $\pm$ 0.04 \\
    2 & 5.90 $\pm$ 0.54 & 10.00 $\pm$ 0.35 & 6.66 $\pm$ 0.29 & 12.00 $\pm$ 0.82 & 12.30 $\pm$ 0.39 & 17.53 $\pm$ 0.25 & 0.48 $\pm$ 0.05 & 0.57 $\pm$ 0.02 \\
    3 & 6.20 $\pm$ 0.39 & 10.55 $\pm$ 0.72 & 6.68 $\pm$ 0.27 & 12.65 $\pm$ 0.64 & 11.79 $\pm$ 0.36 & 16.71 $\pm$ 0.24 & 0.53 $\pm$ 0.05 & 0.63 $\pm$ 0.05 \\
    4 & 6.18 $\pm$ 0.75 & 11.18 $\pm$ 0.91 & 7.21 $\pm$ 0.57 & 12.70 $\pm$ 0.51 & 11.62 $\pm$ 0.30 & 16.31 $\pm$ 0.35 & 0.53 $\pm$ 0.07 & 0.69 $\pm$ 0.06 \\
    5 & 6.58 $\pm$ 0.63 & 12.25 $\pm$ 0.45 & 7.49 $\pm$ 0.58 & 14.04 $\pm$ 0.76 & 11.52 $\pm$ 0.26 & 16.65 $\pm$ 0.25 & 0.57 $\pm$ 0.05 & 0.74 $\pm$ 0.03 \\
    6 & 6.49 $\pm$ 0.52 & 12.46 $\pm$ 0.75 & 7.73 $\pm$ 0.27 & 14.14 $\pm$ 1.02 & 11.05 $\pm$ 0.25 & 16.33 $\pm$ 0.15 & 0.59 $\pm$ 0.05 & 0.76 $\pm$ 0.05 \\
    7 & 6.59 $\pm$ 0.51 & 12.16 $\pm$ 1.04 & 7.44 $\pm$ 0.33 & 13.81 $\pm$ 0.83 & 10.94 $\pm$ 0.28 & 15.87 $\pm$ 0.36 & 0.60 $\pm$ 0.06 & 0.77 $\pm$ 0.08 \\
    8 & 6.54 $\pm$ 0.36 & 12.58 $\pm$ 0.58 & 7.83 $\pm$ 0.33 & 14.94 $\pm$ 1.00 & 10.79 $\pm$ 0.36 & 16.10 $\pm$ 0.31 & 0.61 $\pm$ 0.04 & 0.78 $\pm$ 0.04 \\
    9 & 6.55 $\pm$ 0.39 & 12.16 $\pm$ 1.09 & 7.20 $\pm$ 0.25 & 14.23 $\pm$ 1.11 & 10.55 $\pm$ 0.34 & 15.87 $\pm$ 0.56 & 0.62 $\pm$ 0.05 & 0.77 $\pm$ 0.06 \\
    10 & 6.17 $\pm$ 0.25 & 11.62 $\pm$ 0.93 & 7.44 $\pm$ 0.38 & 13.92 $\pm$ 0.92 & 10.38 $\pm$ 0.34 & 15.43 $\pm$ 0.51 & 0.60 $\pm$ 0.04 & 0.75 $\pm$ 0.07 \\
    11 & 6.15 $\pm$ 0.56 & 11.41 $\pm$ 0.49 & 6.98 $\pm$ 0.27 & 13.78 $\pm$ 1.02 & 10.22 $\pm$ 0.48 & 15.10 $\pm$ 0.33 & 0.60 $\pm$ 0.06 & 0.76 $\pm$ 0.03 \\
    12 & 6.22 $\pm$ 0.82 & 10.98 $\pm$ 0.76 & 7.46 $\pm$ 0.35 & 13.66 $\pm$ 0.44 & 10.38 $\pm$ 0.28 & 15.07 $\pm$ 0.64 & 0.60 $\pm$ 0.08 & 0.73 $\pm$ 0.06 \\
    13 & 6.35 $\pm$ 0.62 & 11.73 $\pm$ 0.87 & 7.36 $\pm$ 0.45 & 14.12 $\pm$ 0.53 & 10.29 $\pm$ 0.28 & 15.31 $\pm$ 0.49 & 0.62 $\pm$ 0.07 & 0.77 $\pm$ 0.06 \\
    14 & 6.38 $\pm$ 0.52 & 11.72 $\pm$ 1.16 & 7.67 $\pm$ 0.56 & 13.99 $\pm$ 0.87 & 10.35 $\pm$ 0.39 & 14.95 $\pm$ 0.45 & 0.62 $\pm$ 0.06 & 0.78 $\pm$ 0.09 \\
    15 & 6.60 $\pm$ 0.36 & 11.56 $\pm$ 0.61 & 7.51 $\pm$ 0.46 & 13.85 $\pm$ 0.81 & 10.75 $\pm$ 0.25 & 14.99 $\pm$ 0.44 & 0.61 $\pm$ 0.03 & 0.77 $\pm$ 0.06 \\
    16 & 6.13 $\pm$ 0.43 & 10.87 $\pm$ 0.45 & 7.27 $\pm$ 0.22 & 13.06 $\pm$ 0.76 & 10.50 $\pm$ 0.38 & 14.42 $\pm$ 0.32 & 0.58 $\pm$ 0.04 & 0.75 $\pm$ 0.04 \\
    17 & 6.31 $\pm$ 0.90 & 11.21 $\pm$ 0.46 & 7.30 $\pm$ 0.59 & 13.56 $\pm$ 0.39 & 10.39 $\pm$ 0.36 & 14.88 $\pm$ 0.33 & 0.61 $\pm$ 0.09 & 0.75 $\pm$ 0.02 \\
    18 & 6.03 $\pm$ 0.27 & 11.36 $\pm$ 0.38 & 7.33 $\pm$ 0.52 & 13.39 $\pm$ 0.83 & 10.22 $\pm$ 0.31 & 14.82 $\pm$ 0.36 & 0.59 $\pm$ 0.02 & 0.77 $\pm$ 0.01 \\
    19 & 6.31 $\pm$ 0.60 & 10.74 $\pm$ 0.49 & 7.04 $\pm$ 0.61 & 12.72 $\pm$ 0.83 & 10.30 $\pm$ 0.24 & 14.38 $\pm$ 0.37 & 0.61 $\pm$ 0.05 & 0.75 $\pm$ 0.04 \\
    20 & 6.13 $\pm$ 0.30 & 10.29 $\pm$ 0.42 & 6.77 $\pm$ 0.57 & 12.78 $\pm$ 0.43 & 10.15 $\pm$ 0.31 & 14.29 $\pm$ 0.37 & 0.60 $\pm$ 0.02 & 0.72 $\pm$ 0.05 \\
    21 & 5.62 $\pm$ 0.38 & 10.93 $\pm$ 1.03 & 6.93 $\pm$ 0.37 & 12.02 $\pm$ 1.04 & 10.07 $\pm$ 0.31 & 13.65 $\pm$ 0.23 & 0.56 $\pm$ 0.02 & 0.80 $\pm$ 0.07 \\
    22 & 5.32 $\pm$ 0.45 & 10.35 $\pm$ 0.50 & 6.83 $\pm$ 0.29 & 12.23 $\pm$ 0.84 & 9.99 $\pm$ 0.27 & 13.80 $\pm$ 0.57 & 0.53 $\pm$ 0.03 & 0.75 $\pm$ 0.02 \\
    23 & 5.24 $\pm$ 0.39 & 10.23 $\pm$ 0.23 & 6.87 $\pm$ 0.54 & 12.09 $\pm$ 0.37 & 9.61 $\pm$ 0.24 & 13.57 $\pm$ 0.45 & 0.55 $\pm$ 0.04 & 0.75 $\pm$ 0.03 \\
    24 & 5.14 $\pm$ 0.32 & 9.63 $\pm$ 0.53 & 6.78 $\pm$ 0.61 & 11.49 $\pm$ 0.71 & 9.55 $\pm$ 0.30 & 13.32 $\pm$ 0.35 & 0.54 $\pm$ 0.05 & 0.72 $\pm$ 0.04 \\
    25 & 4.99 $\pm$ 0.41 & 9.34 $\pm$ 0.60 & 6.59 $\pm$ 0.36 & 11.87 $\pm$ 0.94 & 9.53 $\pm$ 0.31 & 13.04 $\pm$ 0.46 & 0.52 $\pm$ 0.05 & 0.72 $\pm$ 0.05 \\
    26 & 5.06 $\pm$ 0.23 & 9.18 $\pm$ 0.37 & 6.56 $\pm$ 0.27 & 11.80 $\pm$ 0.63 & 9.50 $\pm$ 0.29 & 13.22 $\pm$ 0.24 & 0.53 $\pm$ 0.02 & 0.69 $\pm$ 0.02 \\
    27 & 5.08 $\pm$ 0.31 & 9.80 $\pm$ 0.96 & 6.37 $\pm$ 0.47 & 11.65 $\pm$ 0.63 & 9.43 $\pm$ 0.37 & 12.85 $\pm$ 0.50 & 0.54 $\pm$ 0.02 & 0.76 $\pm$ 0.05 \\
    28 & 5.04 $\pm$ 0.37 & 9.18 $\pm$ 0.40 & 6.39 $\pm$ 0.44 & 11.54 $\pm$ 0.71 & 9.43 $\pm$ 0.28 & 12.99 $\pm$ 0.43 & 0.53 $\pm$ 0.04 & 0.71 $\pm$ 0.05 \\
    29 & 5.11 $\pm$ 0.37 & 9.81 $\pm$ 0.67 & 6.48 $\pm$ 0.23 & 11.62 $\pm$ 0.98 & 9.49 $\pm$ 0.21 & 13.11 $\pm$ 0.48 & 0.54 $\pm$ 0.03 & 0.75 $\pm$ 0.04 \\
    30 & 4.80 $\pm$ 0.41 & 9.72 $\pm$ 0.72 & 6.50 $\pm$ 0.55 & 11.73 $\pm$ 0.87 & 9.27 $\pm$ 0.28 & 12.70 $\pm$ 0.35 & 0.52 $\pm$ 0.04 & 0.77 $\pm$ 0.06 \\
    31 & 4.83 $\pm$ 0.26 & 9.62 $\pm$ 0.47 & 6.16 $\pm$ 0.56 & 10.79 $\pm$ 0.81 & 9.12 $\pm$ 0.31 & 12.78 $\pm$ 0.47 & 0.53 $\pm$ 0.02 & 0.75 $\pm$ 0.04 \\
    32 & 3.45 $\pm$ 0.25 & 6.93 $\pm$ 0.27 & 4.67 $\pm$ 0.57 & 8.86 $\pm$ 0.55 & 8.06 $\pm$ 0.37 & 10.69 $\pm$ 0.45 & 0.43 $\pm$ 0.02 & 0.65 $\pm$ 0.01 \\
    \bottomrule
  \end{tabular}}
\end{table}

%% file: tables/table_S6.tex
\begin{table}[htb]
  \centering
  \caption{Per-layer position probes of the transformer decoder: the within-run gain and its nested decomposition into position-only, preceding-EEG and current-trial terms (pp), which sum to the within-run gain, and the position-corrected gain (within-run minus position-only). The last column, dispersion, is a validity check on the position-only probe (report 1): the mean cosine distance of the position-only predictions from their own centroid (a cosine distance, not a percentage). A dispersion of zero would mean that the probe's predictions had collapsed onto a single vector when the EEG is masked and only the positional embedding is available, in which case a near-zero position-only gain would be uninformative. It is clearly above zero at every layer, so the position-only terms are valid estimates; at layer 0 the term is zero despite a positive dispersion, as expected when the target carries no context. Mean $\pm$ SD over seeds.}
  \label{tab:layer-probes}
  \footnotesize
  \setlength{\tabcolsep}{4pt}
  \begin{tabular}{lcccccc}
    \toprule
    Layer & within-run & position-only & preceding-EEG & current-trial & position-corr. & dispersion \\
    \midrule
    0 & 9.7 $\pm$ 0.1 & 0.0 $\pm$ 0.2 & 0.0 $\pm$ 0.1 & 9.6 $\pm$ 0.2 & 9.6 $\pm$ 0.2 & 0.156 $\pm$ 0.014 \\
    1 & 15.9 $\pm$ 0.3 & 2.5 $\pm$ 0.4 & 3.4 $\pm$ 0.3 & 10.0 $\pm$ 0.2 & 13.4 $\pm$ 0.3 & 0.121 $\pm$ 0.009 \\
    2 & 16.8 $\pm$ 0.2 & 2.9 $\pm$ 0.6 & 4.0 $\pm$ 0.4 & 10.0 $\pm$ 0.3 & 14.0 $\pm$ 0.7 & 0.117 $\pm$ 0.014 \\
    3 & 16.2 $\pm$ 0.2 & 3.5 $\pm$ 0.4 & 3.2 $\pm$ 0.4 & 9.5 $\pm$ 0.3 & 12.7 $\pm$ 0.6 & 0.118 $\pm$ 0.012 \\
    4 & 15.8 $\pm$ 0.3 & 3.2 $\pm$ 0.7 & 3.3 $\pm$ 0.6 & 9.4 $\pm$ 0.6 & 12.7 $\pm$ 0.8 & 0.101 $\pm$ 0.011 \\
    5 & 16.3 $\pm$ 0.2 & 3.0 $\pm$ 0.6 & 3.5 $\pm$ 0.6 & 9.8 $\pm$ 0.2 & 13.3 $\pm$ 0.5 & 0.096 $\pm$ 0.009 \\
    6 & 16.0 $\pm$ 0.1 & 3.0 $\pm$ 0.8 & 3.6 $\pm$ 0.7 & 9.4 $\pm$ 0.4 & 13.0 $\pm$ 0.8 & 0.100 $\pm$ 0.012 \\
    7 & 15.6 $\pm$ 0.3 & 3.0 $\pm$ 0.5 & 3.3 $\pm$ 0.4 & 9.3 $\pm$ 0.1 & 12.5 $\pm$ 0.6 & 0.087 $\pm$ 0.005 \\
    8 & 15.9 $\pm$ 0.3 & 3.3 $\pm$ 0.3 & 3.4 $\pm$ 0.3 & 9.1 $\pm$ 0.2 & 12.6 $\pm$ 0.3 & 0.091 $\pm$ 0.009 \\
    9 & 15.6 $\pm$ 0.6 & 3.5 $\pm$ 0.9 & 3.3 $\pm$ 0.8 & 8.8 $\pm$ 0.5 & 12.1 $\pm$ 1.2 & 0.095 $\pm$ 0.012 \\
    10 & 15.2 $\pm$ 0.4 & 3.6 $\pm$ 0.8 & 3.1 $\pm$ 0.5 & 8.5 $\pm$ 0.4 & 11.6 $\pm$ 0.7 & 0.082 $\pm$ 0.010 \\
    11 & 14.8 $\pm$ 0.3 & 3.5 $\pm$ 0.4 & 2.7 $\pm$ 0.3 & 8.7 $\pm$ 0.2 & 11.4 $\pm$ 0.3 & 0.097 $\pm$ 0.010 \\
    12 & 14.8 $\pm$ 0.6 & 3.4 $\pm$ 0.4 & 2.9 $\pm$ 0.3 & 8.5 $\pm$ 0.4 & 11.4 $\pm$ 0.6 & 0.087 $\pm$ 0.009 \\
    13 & 15.1 $\pm$ 0.4 & 3.6 $\pm$ 0.5 & 3.0 $\pm$ 0.2 & 8.5 $\pm$ 0.1 & 11.5 $\pm$ 0.3 & 0.087 $\pm$ 0.003 \\
    14 & 14.7 $\pm$ 0.4 & 3.1 $\pm$ 0.1 & 3.2 $\pm$ 0.3 & 8.4 $\pm$ 0.5 & 11.6 $\pm$ 0.4 & 0.086 $\pm$ 0.011 \\
    15 & 14.8 $\pm$ 0.4 & 3.3 $\pm$ 0.3 & 2.6 $\pm$ 0.3 & 9.0 $\pm$ 0.4 & 11.5 $\pm$ 0.4 & 0.090 $\pm$ 0.010 \\
    16 & 14.2 $\pm$ 0.3 & 3.2 $\pm$ 0.3 & 2.7 $\pm$ 0.4 & 8.3 $\pm$ 0.4 & 10.9 $\pm$ 0.1 & 0.090 $\pm$ 0.008 \\
    17 & 14.6 $\pm$ 0.3 & 3.1 $\pm$ 0.1 & 2.9 $\pm$ 0.3 & 8.7 $\pm$ 0.4 & 11.6 $\pm$ 0.3 & 0.091 $\pm$ 0.010 \\
    18 & 14.6 $\pm$ 0.3 & 2.8 $\pm$ 0.3 & 2.9 $\pm$ 0.3 & 8.8 $\pm$ 0.2 & 11.7 $\pm$ 0.2 & 0.090 $\pm$ 0.006 \\
    19 & 14.1 $\pm$ 0.3 & 3.2 $\pm$ 0.1 & 2.5 $\pm$ 0.1 & 8.4 $\pm$ 0.4 & 10.9 $\pm$ 0.4 & 0.091 $\pm$ 0.011 \\
    20 & 14.0 $\pm$ 0.3 & 2.4 $\pm$ 0.3 & 3.1 $\pm$ 0.4 & 8.5 $\pm$ 0.3 & 11.6 $\pm$ 0.4 & 0.077 $\pm$ 0.006 \\
    21 & 13.4 $\pm$ 0.3 & 2.9 $\pm$ 0.2 & 2.6 $\pm$ 0.3 & 8.0 $\pm$ 0.3 & 10.6 $\pm$ 0.3 & 0.102 $\pm$ 0.018 \\
    22 & 13.5 $\pm$ 0.5 & 2.6 $\pm$ 0.3 & 2.9 $\pm$ 0.3 & 8.0 $\pm$ 0.3 & 10.9 $\pm$ 0.5 & 0.096 $\pm$ 0.011 \\
    23 & 13.3 $\pm$ 0.4 & 2.9 $\pm$ 0.2 & 2.6 $\pm$ 0.3 & 7.9 $\pm$ 0.2 & 10.4 $\pm$ 0.4 & 0.090 $\pm$ 0.024 \\
    24 & 13.0 $\pm$ 0.3 & 3.0 $\pm$ 0.2 & 2.3 $\pm$ 0.2 & 7.7 $\pm$ 0.2 & 10.1 $\pm$ 0.3 & 0.095 $\pm$ 0.010 \\
    25 & 12.8 $\pm$ 0.4 & 2.4 $\pm$ 0.3 & 2.7 $\pm$ 0.3 & 7.7 $\pm$ 0.6 & 10.4 $\pm$ 0.4 & 0.092 $\pm$ 0.012 \\
    26 & 12.9 $\pm$ 0.2 & 2.4 $\pm$ 0.4 & 3.0 $\pm$ 0.4 & 7.6 $\pm$ 0.3 & 10.6 $\pm$ 0.3 & 0.090 $\pm$ 0.012 \\
    27 & 12.6 $\pm$ 0.5 & 2.4 $\pm$ 0.4 & 2.6 $\pm$ 0.3 & 7.6 $\pm$ 0.4 & 10.2 $\pm$ 0.5 & 0.092 $\pm$ 0.015 \\
    28 & 12.7 $\pm$ 0.4 & 2.4 $\pm$ 0.6 & 2.7 $\pm$ 0.4 & 7.6 $\pm$ 0.4 & 10.3 $\pm$ 0.6 & 0.093 $\pm$ 0.012 \\
    29 & 12.9 $\pm$ 0.4 & 2.6 $\pm$ 0.3 & 2.6 $\pm$ 0.6 & 7.6 $\pm$ 0.4 & 10.2 $\pm$ 0.5 & 0.093 $\pm$ 0.006 \\
    30 & 12.5 $\pm$ 0.3 & 2.4 $\pm$ 0.3 & 2.9 $\pm$ 0.4 & 7.1 $\pm$ 0.4 & 10.1 $\pm$ 0.1 & 0.096 $\pm$ 0.008 \\
    31 & 12.5 $\pm$ 0.4 & 2.9 $\pm$ 0.7 & 2.5 $\pm$ 0.6 & 7.1 $\pm$ 0.3 & 9.6 $\pm$ 0.5 & 0.101 $\pm$ 0.013 \\
    32 & 10.4 $\pm$ 0.4 & 2.3 $\pm$ 0.7 & 2.2 $\pm$ 0.7 & 5.9 $\pm$ 0.4 & 8.1 $\pm$ 0.6 & 0.104 $\pm$ 0.016 \\
    \bottomrule
  \end{tabular}
\end{table}

%% file: tables/table_S7.tex
\begin{table}[htb]
  \centering
  \caption{Per-layer lag-dependent run-mate swap for both decoders (pp, mean $\pm$ SD over seeds): the cross-run gain after replacing each trial's prediction with that of a random run-mate (report 1's context-tracking control), the same random swap applied to the target-geometry control (a pure word decoder with no information from the EEG, calibrated to the decoder's within-run gain), and the swap with the run-mate 4 or 256 trials away (earlier and later averaged). The table is the full-depth version of Table~\ref{tab:lag} and Figure~\ref{fig:lag}. The random-swap gain lies far above the geometry control at every layer above 0, so the context-tracking gain is not a by-product of word decoding and target geometry. Without a sequence model the gain at lag 256 is close to the gain at lag 4, so the component that the trials of a run share is run-level; with the sequence model the gain is large at lag 4 and collapses at lag 256, so its context is local. At layer 0, where the target is context-free, all columns are within 2\,pp of zero.}
  \label{tab:layer-lag}
  \footnotesize
  \setlength{\tabcolsep}{4pt}
  \resizebox{\textwidth}{!}{%
  \begin{tabular}{lcccccccc}
    \toprule
    Layer & \makecell[c]{random\\noTF} & \makecell[c]{geom. control\\noTF} & \makecell[c]{lag 4\\noTF} & \makecell[c]{lag 256\\noTF} & \makecell[c]{random\\TF} & \makecell[c]{geom. control\\TF} & \makecell[c]{lag 4\\TF} & \makecell[c]{lag 256\\TF} \\
    \midrule
    0 & 1.0 $\pm$ 0.1 & 0.1 $\pm$ 0.1 & 1.1 $\pm$ 0.1 & 1.0 $\pm$ 0.1 & 1.6 $\pm$ 0.0 & 0.1 $\pm$ 0.0 & 1.8 $\pm$ 0.1 & 1.7 $\pm$ 0.0 \\
    1 & 3.8 $\pm$ 0.3 & 0.4 $\pm$ 0.0 & 5.2 $\pm$ 0.3 & 3.5 $\pm$ 0.3 & 6.5 $\pm$ 0.6 & 1.4 $\pm$ 0.0 & 20.7 $\pm$ 1.3 & 0.7 $\pm$ 0.3 \\
    2 & 3.9 $\pm$ 0.4 & 0.4 $\pm$ 0.1 & 5.4 $\pm$ 0.4 & 3.6 $\pm$ 0.4 & 6.3 $\pm$ 0.7 & 1.3 $\pm$ 0.1 & 22.0 $\pm$ 1.6 & 0.1 $\pm$ 0.3 \\
    3 & 4.3 $\pm$ 0.4 & 0.4 $\pm$ 0.1 & 5.8 $\pm$ 0.5 & 3.9 $\pm$ 0.4 & 6.3 $\pm$ 0.3 & 1.3 $\pm$ 0.1 & 21.9 $\pm$ 0.9 & 0.2 $\pm$ 0.1 \\
    4 & 4.7 $\pm$ 0.5 & 0.5 $\pm$ 0.1 & 6.4 $\pm$ 0.6 & 4.4 $\pm$ 0.5 & 6.7 $\pm$ 0.7 & 1.3 $\pm$ 0.1 & 22.2 $\pm$ 1.2 & 0.5 $\pm$ 0.3 \\
    5 & 5.0 $\pm$ 0.4 & 0.6 $\pm$ 0.1 & 6.8 $\pm$ 0.4 & 4.6 $\pm$ 0.4 & 7.2 $\pm$ 0.7 & 1.4 $\pm$ 0.1 & 24.1 $\pm$ 1.2 & 0.6 $\pm$ 0.3 \\
    6 & 5.3 $\pm$ 0.4 & 0.6 $\pm$ 0.1 & 7.1 $\pm$ 0.4 & 4.9 $\pm$ 0.4 & 7.8 $\pm$ 0.5 & 1.4 $\pm$ 0.1 & 25.0 $\pm$ 1.0 & 1.2 $\pm$ 0.3 \\
    7 & 5.4 $\pm$ 0.4 & 0.6 $\pm$ 0.0 & 7.4 $\pm$ 0.4 & 5.0 $\pm$ 0.3 & 7.9 $\pm$ 0.6 & 1.4 $\pm$ 0.1 & 25.0 $\pm$ 1.2 & 1.2 $\pm$ 0.3 \\
    8 & 5.4 $\pm$ 0.4 & 0.6 $\pm$ 0.1 & 7.3 $\pm$ 0.3 & 4.9 $\pm$ 0.4 & 8.1 $\pm$ 0.7 & 1.6 $\pm$ 0.1 & 26.8 $\pm$ 1.4 & 0.8 $\pm$ 0.3 \\
    9 & 5.9 $\pm$ 0.5 & 0.6 $\pm$ 0.0 & 8.0 $\pm$ 0.6 & 5.4 $\pm$ 0.5 & 8.3 $\pm$ 0.8 & 1.6 $\pm$ 0.1 & 27.3 $\pm$ 1.6 & 1.0 $\pm$ 0.2 \\
    10 & 6.1 $\pm$ 0.7 & 0.6 $\pm$ 0.0 & 8.2 $\pm$ 0.7 & 5.7 $\pm$ 0.6 & 8.5 $\pm$ 0.6 & 1.5 $\pm$ 0.1 & 26.9 $\pm$ 1.3 & 1.4 $\pm$ 0.2 \\
    11 & 6.1 $\pm$ 0.5 & 0.6 $\pm$ 0.1 & 8.3 $\pm$ 0.5 & 5.6 $\pm$ 0.5 & 8.2 $\pm$ 0.7 & 1.5 $\pm$ 0.1 & 25.5 $\pm$ 1.7 & 1.5 $\pm$ 0.4 \\
    12 & 5.7 $\pm$ 0.6 & 0.7 $\pm$ 0.1 & 7.8 $\pm$ 0.5 & 5.3 $\pm$ 0.6 & 8.3 $\pm$ 1.0 & 1.5 $\pm$ 0.1 & 25.8 $\pm$ 2.3 & 1.8 $\pm$ 0.3 \\
    13 & 5.7 $\pm$ 0.6 & 0.6 $\pm$ 0.1 & 7.8 $\pm$ 0.6 & 5.3 $\pm$ 0.5 & 8.0 $\pm$ 0.5 & 1.5 $\pm$ 0.1 & 25.7 $\pm$ 1.2 & 1.2 $\pm$ 0.1 \\
    14 & 6.1 $\pm$ 0.5 & 0.7 $\pm$ 0.1 & 8.4 $\pm$ 0.5 & 5.7 $\pm$ 0.4 & 8.7 $\pm$ 0.7 & 1.5 $\pm$ 0.1 & 26.5 $\pm$ 1.6 & 1.8 $\pm$ 0.2 \\
    15 & 6.2 $\pm$ 0.6 & 0.6 $\pm$ 0.1 & 8.3 $\pm$ 0.6 & 5.8 $\pm$ 0.5 & 8.4 $\pm$ 0.7 & 1.4 $\pm$ 0.1 & 24.8 $\pm$ 1.4 & 2.2 $\pm$ 0.2 \\
    16 & 6.6 $\pm$ 0.5 & 0.6 $\pm$ 0.1 & 8.9 $\pm$ 0.4 & 6.2 $\pm$ 0.5 & 8.5 $\pm$ 0.5 & 1.4 $\pm$ 0.1 & 24.2 $\pm$ 1.0 & 2.5 $\pm$ 0.1 \\
    17 & 6.6 $\pm$ 0.5 & 0.6 $\pm$ 0.0 & 8.8 $\pm$ 0.5 & 6.1 $\pm$ 0.5 & 8.6 $\pm$ 0.6 & 1.4 $\pm$ 0.1 & 25.1 $\pm$ 1.2 & 2.4 $\pm$ 0.1 \\
    18 & 6.8 $\pm$ 0.5 & 0.7 $\pm$ 0.1 & 8.9 $\pm$ 0.5 & 6.2 $\pm$ 0.6 & 9.0 $\pm$ 0.9 & 1.4 $\pm$ 0.1 & 25.5 $\pm$ 1.7 & 2.7 $\pm$ 0.4 \\
    19 & 6.9 $\pm$ 0.6 & 0.7 $\pm$ 0.0 & 9.1 $\pm$ 0.6 & 6.4 $\pm$ 0.6 & 8.7 $\pm$ 0.8 & 1.5 $\pm$ 0.0 & 24.4 $\pm$ 1.6 & 2.4 $\pm$ 0.2 \\
    20 & 6.6 $\pm$ 0.5 & 0.7 $\pm$ 0.1 & 8.9 $\pm$ 0.5 & 6.3 $\pm$ 0.6 & 8.6 $\pm$ 0.7 & 1.4 $\pm$ 0.1 & 24.1 $\pm$ 1.5 & 2.7 $\pm$ 0.4 \\
    21 & 6.4 $\pm$ 0.6 & 0.6 $\pm$ 0.0 & 8.4 $\pm$ 0.5 & 6.0 $\pm$ 0.5 & 8.3 $\pm$ 0.9 & 1.3 $\pm$ 0.1 & 22.7 $\pm$ 1.7 & 2.8 $\pm$ 0.4 \\
    22 & 6.4 $\pm$ 0.8 & 0.7 $\pm$ 0.1 & 8.4 $\pm$ 1.0 & 6.1 $\pm$ 0.7 & 8.4 $\pm$ 0.8 & 1.3 $\pm$ 0.1 & 23.3 $\pm$ 1.6 & 2.8 $\pm$ 0.3 \\
    23 & 6.6 $\pm$ 0.5 & 0.6 $\pm$ 0.1 & 8.6 $\pm$ 0.5 & 6.2 $\pm$ 0.5 & 8.6 $\pm$ 0.7 & 1.3 $\pm$ 0.1 & 23.2 $\pm$ 1.4 & 3.0 $\pm$ 0.3 \\
    24 & 6.6 $\pm$ 0.5 & 0.6 $\pm$ 0.1 & 8.7 $\pm$ 0.5 & 6.2 $\pm$ 0.5 & 7.9 $\pm$ 0.5 & 1.2 $\pm$ 0.1 & 21.7 $\pm$ 0.9 & 2.8 $\pm$ 0.3 \\
    25 & 6.2 $\pm$ 0.6 & 0.6 $\pm$ 0.1 & 8.1 $\pm$ 0.8 & 5.8 $\pm$ 0.5 & 8.2 $\pm$ 0.4 & 1.2 $\pm$ 0.0 & 21.9 $\pm$ 0.9 & 3.2 $\pm$ 0.2 \\
    26 & 6.3 $\pm$ 0.5 & 0.6 $\pm$ 0.1 & 8.2 $\pm$ 0.5 & 5.9 $\pm$ 0.4 & 8.4 $\pm$ 1.0 & 1.2 $\pm$ 0.1 & 22.3 $\pm$ 1.7 & 3.2 $\pm$ 0.5 \\
    27 & 6.4 $\pm$ 0.5 & 0.6 $\pm$ 0.1 & 8.4 $\pm$ 0.6 & 6.0 $\pm$ 0.5 & 8.1 $\pm$ 0.9 & 1.2 $\pm$ 0.1 & 21.5 $\pm$ 1.8 & 3.0 $\pm$ 0.4 \\
    28 & 6.4 $\pm$ 0.5 & 0.6 $\pm$ 0.1 & 8.3 $\pm$ 0.5 & 6.0 $\pm$ 0.4 & 8.1 $\pm$ 0.9 & 1.2 $\pm$ 0.1 & 22.2 $\pm$ 1.7 & 2.8 $\pm$ 0.4 \\
    29 & 6.4 $\pm$ 0.6 & 0.7 $\pm$ 0.1 & 8.3 $\pm$ 0.7 & 6.0 $\pm$ 0.5 & 8.3 $\pm$ 0.5 & 1.2 $\pm$ 0.1 & 22.4 $\pm$ 1.0 & 2.9 $\pm$ 0.2 \\
    30 & 6.4 $\pm$ 0.6 & 0.7 $\pm$ 0.0 & 8.4 $\pm$ 0.5 & 6.0 $\pm$ 0.5 & 8.1 $\pm$ 0.6 & 1.2 $\pm$ 0.1 & 21.9 $\pm$ 1.1 & 2.9 $\pm$ 0.2 \\
    31 & 6.9 $\pm$ 0.5 & 0.7 $\pm$ 0.1 & 9.0 $\pm$ 0.6 & 6.4 $\pm$ 0.5 & 8.3 $\pm$ 0.7 & 1.4 $\pm$ 0.1 & 22.0 $\pm$ 1.4 & 3.1 $\pm$ 0.5 \\
    32 & 7.2 $\pm$ 0.6 & 0.6 $\pm$ 0.0 & 9.2 $\pm$ 0.7 & 6.8 $\pm$ 0.6 & 8.9 $\pm$ 1.4 & 1.2 $\pm$ 0.1 & 20.8 $\pm$ 2.3 & 4.6 $\pm$ 0.8 \\
    \bottomrule
  \end{tabular}}
\end{table}

%% file: tables/table_S8.tex
\begin{table}[htb]
  \centering
  \caption{Per-layer retained fraction of a validation run's cross-run gain when its trials take the predictions of a narrative neighbour, a validation run 2--8 places away in the text (Section~\ref{sec:session}; definition in Section~\ref{sec:s0}), mean $\pm$ SD over seeds. Columns: the decoder (noTF, TF); how the neighbour's trials are assigned (\emph{trials matched}: trial $m$ of the run takes trial $m$ of the neighbour run, which preserves any component that depends on position within the run; \emph{trials scrambled}: a random trial of the neighbour run); and whether the neighbour run was recorded in the same session as the run (\emph{same session}) or in a different one (\emph{different session}). The table is the full-depth version of Table~\ref{tab:session} and Figure~\ref{fig:session}a--d. A recording-session signature would make the different-session fraction collapse towards zero; instead it stays at roughly three quarters of the same-session fraction at every layer above 0, so most of what neighbouring runs share follows the text. Scrambling the neighbour's trials removes about a fifth of the fraction without a sequence model and about two thirds of it with the transformer, whose positional embedding carries the within-run position component measured by the probes of Table~\ref{tab:layer-probes}.}
  \label{tab:layer-session}
  \footnotesize
  \setlength{\tabcolsep}{4pt}
  \resizebox{\textwidth}{!}{%
  \begin{tabular}{lcccccccc}
    \toprule
    Layer & \makecell[c]{noTF\\trials matched\\same session} & \makecell[c]{noTF\\trials matched\\different session} & \makecell[c]{noTF\\trials scrambled\\same session} & \makecell[c]{noTF\\trials scrambled\\different session} & \makecell[c]{TF\\trials matched\\same session} & \makecell[c]{TF\\trials matched\\different session} & \makecell[c]{TF\\trials scrambled\\same session} & \makecell[c]{TF\\trials scrambled\\different session} \\
    \midrule
    0 & 0.08 $\pm$ 0.02 & 0.08 $\pm$ 0.03 & 0.07 $\pm$ 0.02 & 0.08 $\pm$ 0.02 & 0.13 $\pm$ 0.02 & 0.11 $\pm$ 0.01 & 0.12 $\pm$ 0.02 & 0.11 $\pm$ 0.01 \\
    1 & 0.25 $\pm$ 0.04 & 0.21 $\pm$ 0.03 & 0.19 $\pm$ 0.04 & 0.16 $\pm$ 0.03 & 0.38 $\pm$ 0.04 & 0.32 $\pm$ 0.03 & 0.10 $\pm$ 0.02 & 0.08 $\pm$ 0.02 \\
    2 & 0.25 $\pm$ 0.03 & 0.21 $\pm$ 0.02 & 0.19 $\pm$ 0.04 & 0.16 $\pm$ 0.03 & 0.41 $\pm$ 0.04 & 0.36 $\pm$ 0.03 & 0.10 $\pm$ 0.02 & 0.08 $\pm$ 0.02 \\
    3 & 0.26 $\pm$ 0.04 & 0.22 $\pm$ 0.03 & 0.20 $\pm$ 0.04 & 0.17 $\pm$ 0.03 & 0.42 $\pm$ 0.04 & 0.36 $\pm$ 0.04 & 0.11 $\pm$ 0.02 & 0.08 $\pm$ 0.02 \\
    4 & 0.28 $\pm$ 0.04 & 0.22 $\pm$ 0.04 & 0.22 $\pm$ 0.04 & 0.18 $\pm$ 0.04 & 0.44 $\pm$ 0.05 & 0.36 $\pm$ 0.07 & 0.12 $\pm$ 0.02 & 0.09 $\pm$ 0.03 \\
    5 & 0.29 $\pm$ 0.04 & 0.22 $\pm$ 0.03 & 0.23 $\pm$ 0.04 & 0.17 $\pm$ 0.03 & 0.44 $\pm$ 0.04 & 0.35 $\pm$ 0.04 & 0.12 $\pm$ 0.02 & 0.08 $\pm$ 0.02 \\
    6 & 0.30 $\pm$ 0.04 & 0.23 $\pm$ 0.03 & 0.24 $\pm$ 0.04 & 0.18 $\pm$ 0.04 & 0.43 $\pm$ 0.03 & 0.35 $\pm$ 0.04 & 0.13 $\pm$ 0.02 & 0.09 $\pm$ 0.02 \\
    7 & 0.31 $\pm$ 0.04 & 0.23 $\pm$ 0.03 & 0.24 $\pm$ 0.04 & 0.18 $\pm$ 0.03 & 0.43 $\pm$ 0.05 & 0.35 $\pm$ 0.03 & 0.13 $\pm$ 0.02 & 0.09 $\pm$ 0.01 \\
    8 & 0.31 $\pm$ 0.04 & 0.24 $\pm$ 0.03 & 0.24 $\pm$ 0.04 & 0.19 $\pm$ 0.03 & 0.43 $\pm$ 0.05 & 0.35 $\pm$ 0.04 & 0.12 $\pm$ 0.02 & 0.08 $\pm$ 0.02 \\
    9 & 0.32 $\pm$ 0.04 & 0.23 $\pm$ 0.03 & 0.24 $\pm$ 0.04 & 0.18 $\pm$ 0.04 & 0.45 $\pm$ 0.05 & 0.36 $\pm$ 0.05 & 0.13 $\pm$ 0.02 & 0.08 $\pm$ 0.02 \\
    10 & 0.33 $\pm$ 0.04 & 0.24 $\pm$ 0.04 & 0.26 $\pm$ 0.05 & 0.18 $\pm$ 0.04 & 0.46 $\pm$ 0.06 & 0.35 $\pm$ 0.04 & 0.14 $\pm$ 0.03 & 0.08 $\pm$ 0.01 \\
    11 & 0.33 $\pm$ 0.04 & 0.24 $\pm$ 0.04 & 0.26 $\pm$ 0.05 & 0.19 $\pm$ 0.04 & 0.45 $\pm$ 0.04 & 0.36 $\pm$ 0.04 & 0.13 $\pm$ 0.02 & 0.09 $\pm$ 0.02 \\
    12 & 0.32 $\pm$ 0.04 & 0.24 $\pm$ 0.05 & 0.25 $\pm$ 0.05 & 0.19 $\pm$ 0.04 & 0.45 $\pm$ 0.05 & 0.37 $\pm$ 0.04 & 0.13 $\pm$ 0.02 & 0.10 $\pm$ 0.02 \\
    13 & 0.31 $\pm$ 0.04 & 0.23 $\pm$ 0.03 & 0.25 $\pm$ 0.05 & 0.19 $\pm$ 0.03 & 0.46 $\pm$ 0.05 & 0.35 $\pm$ 0.04 & 0.13 $\pm$ 0.02 & 0.08 $\pm$ 0.02 \\
    14 & 0.33 $\pm$ 0.04 & 0.24 $\pm$ 0.03 & 0.26 $\pm$ 0.04 & 0.18 $\pm$ 0.04 & 0.46 $\pm$ 0.06 & 0.34 $\pm$ 0.04 & 0.14 $\pm$ 0.03 & 0.08 $\pm$ 0.02 \\
    15 & 0.32 $\pm$ 0.04 & 0.23 $\pm$ 0.03 & 0.25 $\pm$ 0.04 & 0.18 $\pm$ 0.03 & 0.45 $\pm$ 0.06 & 0.35 $\pm$ 0.04 & 0.14 $\pm$ 0.03 & 0.09 $\pm$ 0.01 \\
    16 & 0.33 $\pm$ 0.04 & 0.23 $\pm$ 0.04 & 0.26 $\pm$ 0.05 & 0.18 $\pm$ 0.04 & 0.47 $\pm$ 0.04 & 0.35 $\pm$ 0.03 & 0.16 $\pm$ 0.02 & 0.09 $\pm$ 0.02 \\
    17 & 0.32 $\pm$ 0.04 & 0.23 $\pm$ 0.04 & 0.25 $\pm$ 0.04 & 0.18 $\pm$ 0.04 & 0.47 $\pm$ 0.04 & 0.37 $\pm$ 0.05 & 0.15 $\pm$ 0.02 & 0.10 $\pm$ 0.02 \\
    18 & 0.33 $\pm$ 0.04 & 0.24 $\pm$ 0.04 & 0.26 $\pm$ 0.04 & 0.19 $\pm$ 0.04 & 0.47 $\pm$ 0.04 & 0.35 $\pm$ 0.05 & 0.16 $\pm$ 0.02 & 0.09 $\pm$ 0.02 \\
    19 & 0.33 $\pm$ 0.05 & 0.24 $\pm$ 0.04 & 0.26 $\pm$ 0.05 & 0.19 $\pm$ 0.04 & 0.49 $\pm$ 0.06 & 0.35 $\pm$ 0.04 & 0.16 $\pm$ 0.03 & 0.10 $\pm$ 0.02 \\
    20 & 0.33 $\pm$ 0.04 & 0.24 $\pm$ 0.04 & 0.26 $\pm$ 0.05 & 0.19 $\pm$ 0.04 & 0.48 $\pm$ 0.05 & 0.36 $\pm$ 0.07 & 0.16 $\pm$ 0.03 & 0.10 $\pm$ 0.03 \\
    21 & 0.33 $\pm$ 0.04 & 0.24 $\pm$ 0.04 & 0.26 $\pm$ 0.05 & 0.19 $\pm$ 0.04 & 0.48 $\pm$ 0.04 & 0.37 $\pm$ 0.05 & 0.16 $\pm$ 0.03 & 0.10 $\pm$ 0.03 \\
    22 & 0.33 $\pm$ 0.05 & 0.24 $\pm$ 0.05 & 0.26 $\pm$ 0.05 & 0.19 $\pm$ 0.04 & 0.49 $\pm$ 0.06 & 0.37 $\pm$ 0.03 & 0.17 $\pm$ 0.04 & 0.10 $\pm$ 0.02 \\
    23 & 0.34 $\pm$ 0.04 & 0.25 $\pm$ 0.04 & 0.27 $\pm$ 0.05 & 0.20 $\pm$ 0.04 & 0.49 $\pm$ 0.03 & 0.37 $\pm$ 0.06 & 0.17 $\pm$ 0.03 & 0.11 $\pm$ 0.03 \\
    24 & 0.34 $\pm$ 0.04 & 0.25 $\pm$ 0.04 & 0.27 $\pm$ 0.05 & 0.20 $\pm$ 0.04 & 0.49 $\pm$ 0.05 & 0.39 $\pm$ 0.05 & 0.17 $\pm$ 0.03 & 0.11 $\pm$ 0.02 \\
    25 & 0.34 $\pm$ 0.05 & 0.26 $\pm$ 0.04 & 0.27 $\pm$ 0.06 & 0.21 $\pm$ 0.03 & 0.49 $\pm$ 0.05 & 0.38 $\pm$ 0.06 & 0.17 $\pm$ 0.03 & 0.11 $\pm$ 0.03 \\
    26 & 0.34 $\pm$ 0.05 & 0.25 $\pm$ 0.05 & 0.27 $\pm$ 0.05 & 0.20 $\pm$ 0.05 & 0.51 $\pm$ 0.05 & 0.38 $\pm$ 0.04 & 0.18 $\pm$ 0.04 & 0.12 $\pm$ 0.02 \\
    27 & 0.34 $\pm$ 0.05 & 0.26 $\pm$ 0.03 & 0.28 $\pm$ 0.05 & 0.21 $\pm$ 0.03 & 0.51 $\pm$ 0.04 & 0.39 $\pm$ 0.07 & 0.18 $\pm$ 0.03 & 0.11 $\pm$ 0.03 \\
    28 & 0.34 $\pm$ 0.04 & 0.26 $\pm$ 0.04 & 0.27 $\pm$ 0.05 & 0.21 $\pm$ 0.04 & 0.52 $\pm$ 0.03 & 0.39 $\pm$ 0.05 & 0.18 $\pm$ 0.02 & 0.11 $\pm$ 0.03 \\
    29 & 0.34 $\pm$ 0.04 & 0.25 $\pm$ 0.04 & 0.27 $\pm$ 0.04 & 0.20 $\pm$ 0.04 & 0.50 $\pm$ 0.03 & 0.38 $\pm$ 0.05 & 0.18 $\pm$ 0.02 & 0.10 $\pm$ 0.02 \\
    30 & 0.34 $\pm$ 0.04 & 0.26 $\pm$ 0.04 & 0.27 $\pm$ 0.05 & 0.21 $\pm$ 0.04 & 0.52 $\pm$ 0.06 & 0.39 $\pm$ 0.06 & 0.18 $\pm$ 0.03 & 0.11 $\pm$ 0.03 \\
    31 & 0.35 $\pm$ 0.04 & 0.26 $\pm$ 0.04 & 0.28 $\pm$ 0.04 & 0.21 $\pm$ 0.04 & 0.54 $\pm$ 0.06 & 0.40 $\pm$ 0.05 & 0.19 $\pm$ 0.03 & 0.12 $\pm$ 0.03 \\
    32 & 0.38 $\pm$ 0.04 & 0.28 $\pm$ 0.05 & 0.32 $\pm$ 0.05 & 0.23 $\pm$ 0.04 & 0.54 $\pm$ 0.05 & 0.38 $\pm$ 0.07 & 0.22 $\pm$ 0.03 & 0.12 $\pm$ 0.03 \\
    \bottomrule
  \end{tabular}}
\end{table}

%% file: tables/table_S9.tex
\begin{table}[htb]
  \centering
  \caption{Per-layer training and evaluation diagnostics for both decoders, mean $\pm$ SD over seeds: the epoch of the selected checkpoint (budget 100 epochs), the validation loss at that epoch (not comparable across layers, because the contrastive loss depends on the geometry of the projected targets), the final learned temperature (logit scale; clamp 100), and the headroom-normalised top-10 accuracy in mixed and in within-run pools, that is, the gain over the permutation baseline divided by the largest gain a perfect decoder could reach (1 minus the baseline), so that 0 is chance and 1 is perfect retrieval. The table is the numerical version of Supplementary Figure~\ref{fig:s2}. The selected epochs lie below the 100-epoch ceiling at every layer and the temperatures far below the clamp, so no fit was cut short by the budget or ran away in temperature. The normalised accuracies restate the gains of Figures~\ref{fig:word} and~\ref{fig:context} on one scale: in mixed pools the decoders capture 12--13\% of the headroom at layer 0 and at most 23\% (noTF) and 43\% (TF) deeper in the model, and within run 9--20\%, with the within-run values peaking at layers 1--2 and the mixed-pool values at mid depth.}
  \label{tab:diagnostics}
  \footnotesize
  \setlength{\tabcolsep}{4pt}
  \resizebox{\textwidth}{!}{%
  \begin{tabular}{lcccccccccc}
    \toprule
    Layer & \makecell[c]{selected epoch\\noTF} & \makecell[c]{selected epoch\\TF} & \makecell[c]{validation loss\\noTF} & \makecell[c]{validation loss\\TF} & \makecell[c]{temperature\\noTF} & \makecell[c]{temperature\\TF} & \makecell[c]{norm. acc. (mixed)\\noTF} & \makecell[c]{norm. acc. (mixed)\\TF} & \makecell[c]{norm. acc. (within-run)\\noTF} & \makecell[c]{norm. acc. (within-run)\\TF} \\
    \midrule
    0 & 83 $\pm$ 8 & 78 $\pm$ 13 & 8.52 $\pm$ 0.01 & 8.55 $\pm$ 0.01 & 16.09 $\pm$ 0.01 & 16.69 $\pm$ 0.02 & 0.120 $\pm$ 0.003 & 0.128 $\pm$ 0.001 & 0.121 $\pm$ 0.004 & 0.115 $\pm$ 0.001 \\
    1 & 63 $\pm$ 16 & 91 $\pm$ 8 & 8.34 $\pm$ 0.03 & 7.33 $\pm$ 0.04 & 15.94 $\pm$ 0.03 & 17.05 $\pm$ 0.02 & 0.196 $\pm$ 0.006 & 0.377 $\pm$ 0.017 & 0.142 $\pm$ 0.004 & 0.191 $\pm$ 0.004 \\
    2 & 58 $\pm$ 8 & 86 $\pm$ 6 & 8.34 $\pm$ 0.05 & 7.23 $\pm$ 0.03 & 16.01 $\pm$ 0.02 & 17.08 $\pm$ 0.01 & 0.201 $\pm$ 0.008 & 0.396 $\pm$ 0.019 & 0.141 $\pm$ 0.005 & 0.203 $\pm$ 0.003 \\
    3 & 72 $\pm$ 12 & 83 $\pm$ 9 & 8.38 $\pm$ 0.03 & 7.30 $\pm$ 0.06 & 16.02 $\pm$ 0.03 & 17.04 $\pm$ 0.01 & 0.200 $\pm$ 0.007 & 0.386 $\pm$ 0.012 & 0.135 $\pm$ 0.004 & 0.194 $\pm$ 0.003 \\
    4 & 74 $\pm$ 16 & 81 $\pm$ 9 & 8.38 $\pm$ 0.03 & 7.34 $\pm$ 0.09 & 15.99 $\pm$ 0.03 & 17.00 $\pm$ 0.01 & 0.208 $\pm$ 0.011 & 0.391 $\pm$ 0.020 & 0.134 $\pm$ 0.004 & 0.191 $\pm$ 0.004 \\
    5 & 61 $\pm$ 7 & 92 $\pm$ 6 & 8.35 $\pm$ 0.05 & 7.27 $\pm$ 0.05 & 15.93 $\pm$ 0.02 & 16.98 $\pm$ 0.03 & 0.213 $\pm$ 0.008 & 0.414 $\pm$ 0.015 & 0.133 $\pm$ 0.003 & 0.196 $\pm$ 0.003 \\
    6 & 73 $\pm$ 23 & 81 $\pm$ 11 & 8.37 $\pm$ 0.07 & 7.29 $\pm$ 0.04 & 15.93 $\pm$ 0.04 & 16.98 $\pm$ 0.03 & 0.212 $\pm$ 0.007 & 0.420 $\pm$ 0.013 & 0.129 $\pm$ 0.002 & 0.192 $\pm$ 0.001 \\
    7 & 73 $\pm$ 9 & 82 $\pm$ 9 & 8.39 $\pm$ 0.04 & 7.31 $\pm$ 0.04 & 15.92 $\pm$ 0.03 & 16.94 $\pm$ 0.03 & 0.212 $\pm$ 0.009 & 0.416 $\pm$ 0.016 & 0.127 $\pm$ 0.003 & 0.187 $\pm$ 0.003 \\
    8 & 67 $\pm$ 22 & 86 $\pm$ 5 & 8.40 $\pm$ 0.06 & 7.23 $\pm$ 0.10 & 15.93 $\pm$ 0.04 & 16.98 $\pm$ 0.02 & 0.208 $\pm$ 0.008 & 0.432 $\pm$ 0.022 & 0.126 $\pm$ 0.004 & 0.191 $\pm$ 0.004 \\
    9 & 69 $\pm$ 14 & 78 $\pm$ 17 & 8.38 $\pm$ 0.05 & 7.24 $\pm$ 0.08 & 15.92 $\pm$ 0.02 & 16.94 $\pm$ 0.02 & 0.215 $\pm$ 0.009 & 0.433 $\pm$ 0.022 & 0.123 $\pm$ 0.004 & 0.188 $\pm$ 0.007 \\
    10 & 78 $\pm$ 20 & 85 $\pm$ 9 & 8.38 $\pm$ 0.05 & 7.30 $\pm$ 0.07 & 15.92 $\pm$ 0.02 & 16.93 $\pm$ 0.02 & 0.216 $\pm$ 0.011 & 0.424 $\pm$ 0.018 & 0.121 $\pm$ 0.004 & 0.182 $\pm$ 0.006 \\
    11 & 83 $\pm$ 12 & 81 $\pm$ 10 & 8.41 $\pm$ 0.05 & 7.36 $\pm$ 0.07 & 15.93 $\pm$ 0.04 & 16.93 $\pm$ 0.03 & 0.213 $\pm$ 0.007 & 0.405 $\pm$ 0.023 & 0.119 $\pm$ 0.005 & 0.179 $\pm$ 0.004 \\
    12 & 65 $\pm$ 17 & 78 $\pm$ 10 & 8.41 $\pm$ 0.04 & 7.37 $\pm$ 0.11 & 15.94 $\pm$ 0.02 & 16.94 $\pm$ 0.01 & 0.208 $\pm$ 0.009 & 0.409 $\pm$ 0.030 & 0.121 $\pm$ 0.003 & 0.178 $\pm$ 0.007 \\
    13 & 58 $\pm$ 7 & 88 $\pm$ 9 & 8.40 $\pm$ 0.05 & 7.34 $\pm$ 0.07 & 15.95 $\pm$ 0.02 & 16.94 $\pm$ 0.02 & 0.209 $\pm$ 0.010 & 0.406 $\pm$ 0.018 & 0.120 $\pm$ 0.003 & 0.181 $\pm$ 0.005 \\
    14 & 75 $\pm$ 9 & 87 $\pm$ 12 & 8.41 $\pm$ 0.05 & 7.34 $\pm$ 0.05 & 15.90 $\pm$ 0.03 & 16.91 $\pm$ 0.02 & 0.214 $\pm$ 0.009 & 0.414 $\pm$ 0.021 & 0.121 $\pm$ 0.005 & 0.177 $\pm$ 0.005 \\
    15 & 78 $\pm$ 16 & 82 $\pm$ 13 & 8.38 $\pm$ 0.04 & 7.38 $\pm$ 0.06 & 15.95 $\pm$ 0.02 & 16.91 $\pm$ 0.02 & 0.221 $\pm$ 0.012 & 0.402 $\pm$ 0.017 & 0.125 $\pm$ 0.003 & 0.177 $\pm$ 0.004 \\
    16 & 85 $\pm$ 9 & 91 $\pm$ 5 & 8.37 $\pm$ 0.04 & 7.46 $\pm$ 0.04 & 15.93 $\pm$ 0.02 & 16.86 $\pm$ 0.04 & 0.227 $\pm$ 0.008 & 0.392 $\pm$ 0.010 & 0.122 $\pm$ 0.004 & 0.170 $\pm$ 0.003 \\
    17 & 83 $\pm$ 14 & 92 $\pm$ 5 & 8.38 $\pm$ 0.04 & 7.42 $\pm$ 0.05 & 15.92 $\pm$ 0.04 & 16.88 $\pm$ 0.02 & 0.223 $\pm$ 0.010 & 0.401 $\pm$ 0.019 & 0.121 $\pm$ 0.004 & 0.176 $\pm$ 0.004 \\
    18 & 80 $\pm$ 16 & 88 $\pm$ 7 & 8.37 $\pm$ 0.05 & 7.42 $\pm$ 0.04 & 15.85 $\pm$ 0.01 & 16.84 $\pm$ 0.02 & 0.225 $\pm$ 0.009 & 0.411 $\pm$ 0.022 & 0.119 $\pm$ 0.004 & 0.175 $\pm$ 0.004 \\
    19 & 85 $\pm$ 7 & 89 $\pm$ 12 & 8.36 $\pm$ 0.05 & 7.45 $\pm$ 0.06 & 15.76 $\pm$ 0.02 & 16.77 $\pm$ 0.03 & 0.228 $\pm$ 0.010 & 0.393 $\pm$ 0.022 & 0.120 $\pm$ 0.003 & 0.169 $\pm$ 0.004 \\
    20 & 80 $\pm$ 10 & 89 $\pm$ 14 & 8.37 $\pm$ 0.05 & 7.47 $\pm$ 0.05 & 15.73 $\pm$ 0.02 & 16.80 $\pm$ 0.01 & 0.223 $\pm$ 0.011 & 0.391 $\pm$ 0.018 & 0.118 $\pm$ 0.004 & 0.168 $\pm$ 0.004 \\
    21 & 79 $\pm$ 18 & 84 $\pm$ 12 & 8.38 $\pm$ 0.04 & 7.55 $\pm$ 0.05 & 15.75 $\pm$ 0.03 & 16.78 $\pm$ 0.05 & 0.217 $\pm$ 0.009 & 0.369 $\pm$ 0.020 & 0.117 $\pm$ 0.003 & 0.162 $\pm$ 0.003 \\
    22 & 74 $\pm$ 26 & 89 $\pm$ 9 & 8.38 $\pm$ 0.02 & 7.54 $\pm$ 0.08 & 15.73 $\pm$ 0.03 & 16.77 $\pm$ 0.03 & 0.213 $\pm$ 0.017 & 0.374 $\pm$ 0.022 & 0.116 $\pm$ 0.003 & 0.163 $\pm$ 0.006 \\
    23 & 74 $\pm$ 12 & 92 $\pm$ 11 & 8.40 $\pm$ 0.05 & 7.56 $\pm$ 0.06 & 15.75 $\pm$ 0.04 & 16.77 $\pm$ 0.04 & 0.214 $\pm$ 0.010 & 0.375 $\pm$ 0.018 & 0.112 $\pm$ 0.003 & 0.161 $\pm$ 0.005 \\
    24 & 78 $\pm$ 13 & 86 $\pm$ 8 & 8.42 $\pm$ 0.05 & 7.60 $\pm$ 0.05 & 15.76 $\pm$ 0.02 & 16.75 $\pm$ 0.01 & 0.213 $\pm$ 0.009 & 0.354 $\pm$ 0.012 & 0.112 $\pm$ 0.003 & 0.157 $\pm$ 0.004 \\
    25 & 66 $\pm$ 23 & 84 $\pm$ 6 & 8.41 $\pm$ 0.03 & 7.61 $\pm$ 0.06 & 15.77 $\pm$ 0.03 & 16.79 $\pm$ 0.04 & 0.205 $\pm$ 0.013 & 0.355 $\pm$ 0.013 & 0.111 $\pm$ 0.003 & 0.154 $\pm$ 0.005 \\
    26 & 64 $\pm$ 15 & 83 $\pm$ 12 & 8.41 $\pm$ 0.05 & 7.62 $\pm$ 0.09 & 15.77 $\pm$ 0.01 & 16.76 $\pm$ 0.02 & 0.207 $\pm$ 0.010 & 0.359 $\pm$ 0.024 & 0.111 $\pm$ 0.003 & 0.156 $\pm$ 0.003 \\
    27 & 71 $\pm$ 18 & 88 $\pm$ 11 & 8.42 $\pm$ 0.05 & 7.65 $\pm$ 0.07 & 15.77 $\pm$ 0.02 & 16.75 $\pm$ 0.03 & 0.207 $\pm$ 0.012 & 0.351 $\pm$ 0.024 & 0.110 $\pm$ 0.004 & 0.152 $\pm$ 0.006 \\
    28 & 75 $\pm$ 19 & 85 $\pm$ 5 & 8.43 $\pm$ 0.07 & 7.62 $\pm$ 0.07 & 15.76 $\pm$ 0.03 & 16.79 $\pm$ 0.01 & 0.206 $\pm$ 0.007 & 0.356 $\pm$ 0.022 & 0.110 $\pm$ 0.003 & 0.154 $\pm$ 0.005 \\
    29 & 61 $\pm$ 22 & 83 $\pm$ 9 & 8.41 $\pm$ 0.04 & 7.61 $\pm$ 0.06 & 15.77 $\pm$ 0.02 & 16.78 $\pm$ 0.02 & 0.207 $\pm$ 0.013 & 0.358 $\pm$ 0.016 & 0.111 $\pm$ 0.003 & 0.155 $\pm$ 0.005 \\
    30 & 68 $\pm$ 23 & 83 $\pm$ 6 & 8.43 $\pm$ 0.04 & 7.67 $\pm$ 0.06 & 15.75 $\pm$ 0.03 & 16.73 $\pm$ 0.04 & 0.204 $\pm$ 0.010 & 0.347 $\pm$ 0.016 & 0.108 $\pm$ 0.003 & 0.151 $\pm$ 0.004 \\
    31 & 68 $\pm$ 10 & 82 $\pm$ 11 & 8.43 $\pm$ 0.06 & 7.66 $\pm$ 0.07 & 15.65 $\pm$ 0.05 & 16.66 $\pm$ 0.03 & 0.211 $\pm$ 0.011 & 0.350 $\pm$ 0.017 & 0.106 $\pm$ 0.004 & 0.151 $\pm$ 0.005 \\
    32 & 75 $\pm$ 11 & 81 $\pm$ 11 & 8.48 $\pm$ 0.04 & 7.86 $\pm$ 0.09 & 15.50 $\pm$ 0.03 & 16.43 $\pm$ 0.04 & 0.199 $\pm$ 0.011 & 0.316 $\pm$ 0.031 & 0.094 $\pm$ 0.004 & 0.126 $\pm$ 0.005 \\
    \bottomrule
  \end{tabular}}
\end{table}

%% file: tables/table_S10.tex
\begin{table}[htb]
  \centering
  \caption{Restricted candidate set at selected layers. The validation predictions of the main text, scored on trials whose word belongs to a fixed list of 50 words (Supplement Section~\ref{sec:s-small-pool}). Balanced pools hold one trial of every list word, so a pool is a 50-way identification task with uniform chance (top-1 / top-10: 50 content words: 2.0\% / 20.0\%; 50 most frequent words: 2.0\% / 20.0\%; the empirical permutation baseline reproduces these). Two controls are scored on the same pools: every prediction replaced by that of the run-mate one trial earlier (the local discourse state and position without the current word) or of a random run-mate (report 1's context-tracking control). Within-run: report 1's within-run gain restricted to the list words, in pp. Accuracies in percent; mean $\pm$ SD over seeds.}
  \label{tab:small-pool}
  \footnotesize
  \setlength{\tabcolsep}{4pt}
  \resizebox{\textwidth}{!}{%
  \begin{tabular}{lllcccccc}
    \toprule
    List & Decoder & Layer & Top-1 & \makecell[c]{Top-1,\\lag-1 swap} & \makecell[c]{Top-1,\\random swap} & Top-10 & \makecell[c]{Top-10,\\lag-1 swap} & \makecell[c]{Within-run\\gain (pp)} \\
    \midrule
    50 content words & noTF & 0 & 5.1 $\pm$ 0.7 & 2.7 $\pm$ 0.2 & 2.9 $\pm$ 0.1 & 33.8 $\pm$ 1.2 & 23.8 $\pm$ 1.5 & 8.0 $\pm$ 0.6 \\
     &  & 1 & 12.8 $\pm$ 0.5 & 7.2 $\pm$ 0.5 & 5.6 $\pm$ 0.3 & 59.5 $\pm$ 0.9 & 45.5 $\pm$ 1.4 & 13.2 $\pm$ 0.7 \\
     &  & 2 & 13.0 $\pm$ 0.9 & 7.2 $\pm$ 0.6 & 5.7 $\pm$ 0.4 & 60.2 $\pm$ 1.8 & 46.6 $\pm$ 1.5 & 13.3 $\pm$ 0.9 \\
     &  & 4 & 14.0 $\pm$ 1.1 & 8.5 $\pm$ 0.4 & 6.4 $\pm$ 0.5 & 61.7 $\pm$ 2.2 & 50.0 $\pm$ 1.5 & 12.1 $\pm$ 0.8 \\
     &  & 8 & 13.6 $\pm$ 1.0 & 9.6 $\pm$ 0.5 & 6.9 $\pm$ 0.6 & 62.1 $\pm$ 2.3 & 53.5 $\pm$ 2.6 & 11.1 $\pm$ 0.4 \\
     &  & 16 & 15.6 $\pm$ 0.5 & 11.3 $\pm$ 0.4 & 8.3 $\pm$ 0.6 & 64.5 $\pm$ 0.8 & 56.8 $\pm$ 1.9 & 10.9 $\pm$ 0.8 \\
     &  & 20 & 15.4 $\pm$ 1.2 & 11.2 $\pm$ 0.8 & 8.1 $\pm$ 0.5 & 64.6 $\pm$ 1.9 & 56.7 $\pm$ 1.8 & 10.8 $\pm$ 0.4 \\
     &  & 32 & 13.9 $\pm$ 0.9 & 11.4 $\pm$ 0.9 & 8.8 $\pm$ 0.5 & 61.1 $\pm$ 1.6 & 56.9 $\pm$ 1.7 & 7.9 $\pm$ 0.4 \\
     & TF & 0 & 4.8 $\pm$ 0.3 & 3.2 $\pm$ 0.4 & 3.3 $\pm$ 0.2 & 34.0 $\pm$ 1.3 & 26.1 $\pm$ 0.9 & 7.1 $\pm$ 1.2 \\
     &  & 1 & 29.6 $\pm$ 1.9 & 21.9 $\pm$ 1.5 & 8.0 $\pm$ 0.7 & 86.5 $\pm$ 1.2 & 80.7 $\pm$ 1.2 & 26.2 $\pm$ 1.0 \\
     &  & 2 & 30.5 $\pm$ 1.8 & 23.2 $\pm$ 1.7 & 7.7 $\pm$ 0.8 & 88.0 $\pm$ 0.8 & 82.9 $\pm$ 0.6 & 26.5 $\pm$ 1.1 \\
     &  & 4 & 30.3 $\pm$ 2.3 & 24.0 $\pm$ 1.6 & 8.0 $\pm$ 0.7 & 87.5 $\pm$ 2.0 & 83.4 $\pm$ 1.8 & 25.7 $\pm$ 0.9 \\
     &  & 8 & 33.0 $\pm$ 2.0 & 27.9 $\pm$ 1.3 & 9.1 $\pm$ 0.6 & 89.2 $\pm$ 2.0 & 86.7 $\pm$ 2.0 & 25.4 $\pm$ 1.0 \\
     &  & 16 & 30.0 $\pm$ 0.7 & 25.2 $\pm$ 0.8 & 9.7 $\pm$ 0.4 & 85.9 $\pm$ 1.4 & 82.6 $\pm$ 1.0 & 22.3 $\pm$ 1.1 \\
     &  & 20 & 29.1 $\pm$ 0.7 & 24.5 $\pm$ 0.9 & 9.3 $\pm$ 0.4 & 86.8 $\pm$ 0.6 & 83.7 $\pm$ 0.6 & 21.9 $\pm$ 1.0 \\
     &  & 32 & 24.1 $\pm$ 2.4 & 21.6 $\pm$ 1.6 & 9.8 $\pm$ 1.3 & 80.5 $\pm$ 2.4 & 78.3 $\pm$ 2.6 & 17.6 $\pm$ 0.6 \\
    50 most frequent words & noTF & 0 & 5.3 $\pm$ 0.1 & 2.4 $\pm$ 0.1 & 2.3 $\pm$ 0.0 & 34.6 $\pm$ 0.8 & 21.8 $\pm$ 0.3 & 9.2 $\pm$ 0.4 \\
     &  & 1 & 13.5 $\pm$ 0.5 & 7.7 $\pm$ 0.3 & 5.7 $\pm$ 0.3 & 61.0 $\pm$ 1.9 & 47.6 $\pm$ 1.7 & 11.0 $\pm$ 0.4 \\
     &  & 2 & 13.5 $\pm$ 0.6 & 7.8 $\pm$ 0.4 & 5.8 $\pm$ 0.4 & 60.8 $\pm$ 2.1 & 48.3 $\pm$ 1.7 & 10.7 $\pm$ 0.4 \\
     &  & 4 & 13.5 $\pm$ 0.9 & 8.7 $\pm$ 0.5 & 6.3 $\pm$ 0.5 & 60.9 $\pm$ 2.3 & 50.7 $\pm$ 2.2 & 9.5 $\pm$ 0.3 \\
     &  & 8 & 13.7 $\pm$ 0.5 & 9.7 $\pm$ 0.4 & 7.0 $\pm$ 0.5 & 61.3 $\pm$ 2.3 & 53.8 $\pm$ 2.0 & 8.4 $\pm$ 0.5 \\
     &  & 16 & 15.3 $\pm$ 0.7 & 11.2 $\pm$ 0.4 & 8.4 $\pm$ 0.5 & 63.7 $\pm$ 1.9 & 56.6 $\pm$ 1.8 & 8.2 $\pm$ 0.3 \\
     &  & 20 & 15.0 $\pm$ 0.9 & 11.0 $\pm$ 0.5 & 8.4 $\pm$ 0.6 & 63.8 $\pm$ 2.1 & 57.3 $\pm$ 1.8 & 7.9 $\pm$ 0.4 \\
     &  & 32 & 13.6 $\pm$ 0.8 & 11.2 $\pm$ 0.7 & 8.7 $\pm$ 0.6 & 60.3 $\pm$ 1.8 & 56.0 $\pm$ 1.6 & 6.3 $\pm$ 0.3 \\
     & TF & 0 & 5.3 $\pm$ 0.1 & 2.6 $\pm$ 0.1 & 2.6 $\pm$ 0.1 & 34.3 $\pm$ 0.4 & 23.1 $\pm$ 0.2 & 8.9 $\pm$ 0.4 \\
     &  & 1 & 31.6 $\pm$ 1.8 & 24.6 $\pm$ 1.4 & 8.5 $\pm$ 0.6 & 87.5 $\pm$ 1.9 & 83.1 $\pm$ 2.3 & 17.8 $\pm$ 0.3 \\
     &  & 2 & 32.0 $\pm$ 1.9 & 24.9 $\pm$ 1.8 & 8.0 $\pm$ 0.8 & 88.9 $\pm$ 0.9 & 84.7 $\pm$ 1.2 & 18.3 $\pm$ 0.2 \\
     &  & 4 & 30.1 $\pm$ 1.5 & 24.5 $\pm$ 1.2 & 8.2 $\pm$ 0.6 & 87.0 $\pm$ 1.4 & 83.3 $\pm$ 1.5 & 16.3 $\pm$ 0.3 \\
     &  & 8 & 33.2 $\pm$ 1.9 & 28.8 $\pm$ 1.6 & 9.5 $\pm$ 0.7 & 88.8 $\pm$ 2.4 & 86.3 $\pm$ 2.6 & 16.1 $\pm$ 0.2 \\
     &  & 16 & 29.7 $\pm$ 1.0 & 25.5 $\pm$ 0.8 & 10.1 $\pm$ 0.4 & 85.0 $\pm$ 1.6 & 82.1 $\pm$ 1.6 & 13.9 $\pm$ 0.3 \\
     &  & 20 & 29.5 $\pm$ 1.8 & 25.7 $\pm$ 1.5 & 10.1 $\pm$ 0.7 & 86.0 $\pm$ 1.1 & 83.4 $\pm$ 1.3 & 13.8 $\pm$ 0.4 \\
     &  & 32 & 23.3 $\pm$ 2.3 & 21.2 $\pm$ 2.0 & 10.0 $\pm$ 1.2 & 78.5 $\pm$ 2.5 & 76.8 $\pm$ 2.5 & 10.1 $\pm$ 0.7 \\
    \bottomrule
  \end{tabular}}
\end{table}

%% file: tables/table_S11.tex
\begin{table}[htb]
  \centering
  \caption{The two candidate word lists of the restricted-candidate-set analysis, with the number of presentations of each word in the full corpus of the participant (409 runs, 248,826 presentations, 15,352 word types; target events excluded). Content words: the 50 most frequent words that are alphabetic, not in the English stop-word list of scikit-learn 1.9.0, and not proper names (capitalised in more than 90\% of presentations); they account for 5.3\% of all presentations. Most frequent words: the 50 most frequent words of any kind (44.0\% of presentations). Words are the punctuation-stripped, lower-cased forms used for word grouping throughout.}
  \label{tab:word-lists}
  \footnotesize
  \setlength{\tabcolsep}{4pt}
  \begin{tabular}{llll}
    \toprule
    50 content words, 1--25 & 50 content words, 26--50 & 50 most frequent words, 1--25 & 50 most frequent words, 26--50 \\
    \midrule
    1. said (1,206) & 26. left (218) & 1. the (14,112) & 26. on (1,387) \\
    2. man (590) & 27. say (213) & 2. and (6,503) & 27. said (1,206) \\
    3. like (414) & 28. thought (211) & 3. of (5,896) & 28. this (1,147) \\
    4. did (404) & 29. tell (211) & 4. to (5,887) & 29. from (1,128) \\
    5. time (394) & 30. night (204) & 5. a (5,756) & 30. be (1,114) \\
    6. know (357) & 31. make (202) & 6. he (4,064) & 31. my (1,103) \\
    7. eyes (332) & 32. lady (200) & 7. that (3,744) & 32. they (1,077) \\
    8. come (331) & 33. look (198) & 8. i (3,627) & 33. we (1,027) \\
    9. just (323) & 34. yes (198) & 9. in (3,525) & 34. there (1,002) \\
    10. face (312) & 35. saw (197) & 10. was (3,523) & 35. no (956) \\
    11. room (295) & 36. knew (185) & 11. it (3,213) & 36. were (941) \\
    12. way (288) & 37. men (183) & 12. his (3,061) & 37. all (916) \\
    13. hand (287) & 38. heard (181) & 13. you (2,638) & 38. me (892) \\
    14. long (285) & 39. house (178) & 14. had (2,270) & 39. would (877) \\
    15. looked (276) & 40. end (175) & 15. is (2,149) & 40. one (857) \\
    16. little (276) & 41. day (173) & 16. her (1,996) & 41. been (819) \\
    17. told (270) & 42. let (173) & 17. with (1,910) & 42. are (814) \\
    18. away (268) & 43. boy (172) & 18. she (1,699) & 43. so (781) \\
    19. old (261) & 44. head (171) & 19. as (1,678) & 44. them (773) \\
    20. lord (252) & 45. sir (170) & 20. for (1,636) & 45. what (771) \\
    21. came (244) & 46. young (167) & 21. at (1,627) & 46. an (751) \\
    22. think (233) & 47. took (167) & 22. have (1,509) & 47. which (739) \\
    23. door (231) & 48. king (165) & 23. not (1,483) & 48. by (730) \\
    24. great (230) & 49. right (163) & 24. him (1,435) & 49. will (724) \\
    25. good (221) & 50. case (162) & 25. but (1,410) & 50. out (705) \\
    \bottomrule
  \end{tabular}
\end{table}

%% file: tables/table_S12.tex
\begin{table}[htb]
  \centering
  \caption{Number of trainable parameters by component, for the decoder without and with the sequence model (19 channels; the architecture of every fit in this report, identical at every target layer). Parameter tensors are counted at their full size, including the affine parameters of batch and layer normalisation; the temperature is the single learned log-scale scalar of the contrastive loss.}
  \label{tab:params}
  \footnotesize
  \setlength{\tabcolsep}{4pt}
  \begin{tabular}{llrr}
    \toprule
    Component & Detail & \makecell[c]{No sequence\\model} & \makecell[c]{With\\transformer} \\
    \midrule
    EEG encoder & global pathway, four convolutional blocks & 551{,}328 & 551{,}328 \\
     & global pathway, attention pooling & 385 & 385 \\
     & local pathway, three convolutional blocks & 181{,}536 & 181{,}536 \\
     & head, two fully-connected layers and output layer & 1{,}182{,}720 & 1{,}182{,}720 \\
    \texttt{[MASK]} token & 256-dimensional & 256 & 256 \\
    Input layer normalisation &  & 512 & 512 \\
    Sequence model & four transformer encoder layers & -- & 3{,}159{,}040 \\
     & positional embedding, $1{,}400 \times 256$ & -- & 358{,}400 \\
     & final layer normalisation & -- & 512 \\
    Output projection & $256 \times 256$, bias-free & 65{,}536 & 65{,}536 \\
    Target projection & $4{,}096 \times 256$, bias-free & 1{,}048{,}576 & 1{,}048{,}576 \\
    Temperature & learned log-scale scalar & 1 & 1 \\
    \midrule
    Total &  & 3{,}030{,}850 & 6{,}548{,}802 \\
    \bottomrule
  \end{tabular}
\end{table}